\documentclass[10pt,twocolumn,letterpaper]{article}

\usepackage[pagenumbers]{cvpr} 
\usepackage[accsupp]{axessibility}
 
\usepackage{graphicx}
\usepackage{amsmath}
\usepackage{amssymb}
\usepackage{soul}
\usepackage{bm}
\usepackage{booktabs}
\usepackage{float}
\usepackage{multirow}
\usepackage{diagbox}
\usepackage{wrapfig}
\usepackage{amssymb}
\usepackage{pifont}
\usepackage{nopageno}
\usepackage{balance}
\newcommand{\cmark}{\ding{51}}%
\newcommand{\xmark}{\ding{55}}%

\usepackage[pagebackref=true,breaklinks=true,colorlinks,bookmarks=false]{hyperref}

\usepackage[capitalize]{cleveref}
\crefname{section}{Sec.}{Secs.}
\Crefname{section}{Section}{Sections}
\Crefname{table}{Table}{Tables}
\crefname{table}{Tab.}{Tabs.}

\def\cvprPaperID{213} 
\def\confName{CVPR}
\def\confYear{2022}

\begin{document}
	\title{WarpingGAN: Warping Multiple Uniform Priors for Adversarial 3D Point Cloud Generation}
	\author{
Yinzhi Tang$^1$\footnotemark[1] \quad Yue Qian$^1$\footnotemark[1] \quad Qijian Zhang$^1$\quad Yiming Zeng$^1$\quad Junhui Hou$^1$\quad Xuefei Zhe$^2$\\
$^1$
City University of Hong Kong~~
$^2$
Tencent AI lab~~ \\
{\tt\small \{yztang4-c, yueqian4-c, qijizhang3-c, ym.zeng\}@my.cityu.edu.hk, jh.hou@cityu.edu.hk}
}

\twocolumn[{%
	\renewcommand\twocolumn[1][]{#1}%
	\maketitle
	\vspace{-1.1cm}
	\begin{center}
		\centering
		\setlength{\abovecaptionskip}{0.1cm}
		\setlength{\belowcaptionskip}{-0.15cm}
		\captionsetup{type=figure}
		\includegraphics[width=6.0in]{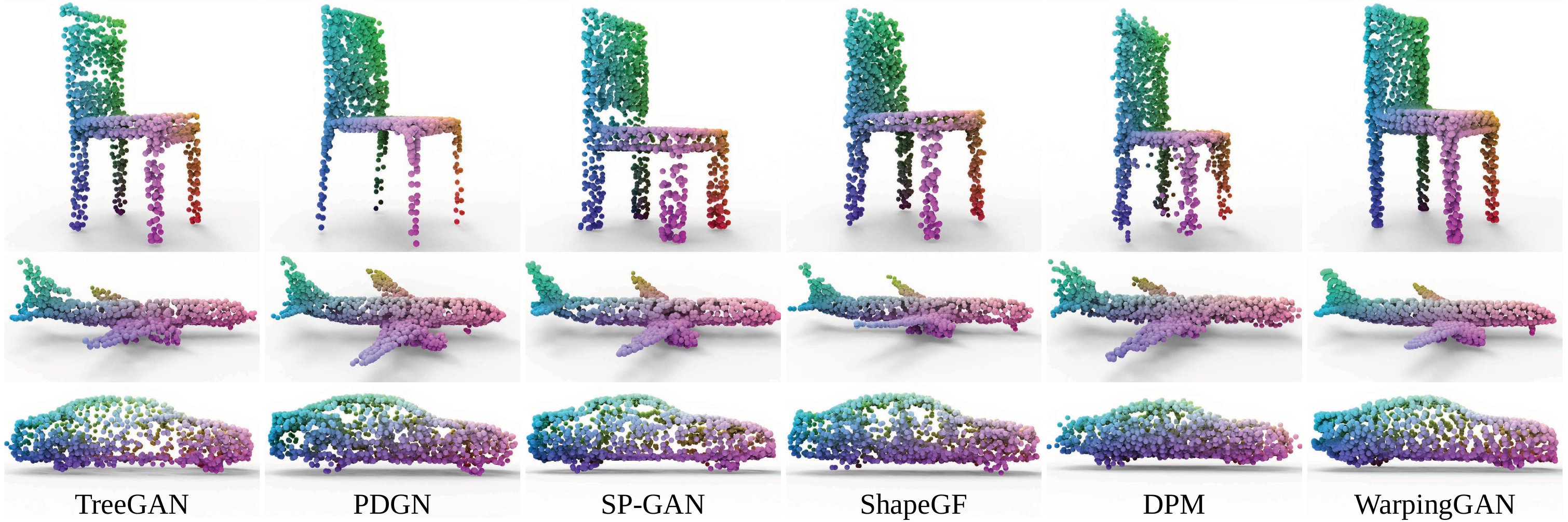}
		\captionof{figure}{Visual comparisons of generated shapes with state-of-the-art 3D point cloud generation methods. TreeGAN~\cite{shu20193d}, PDGN~\cite{hui2020progressive} and SP-GAN~\cite{li2021spgan} are GAN-based, while ShapeGF~\cite{cai2020learning} and DPM~\cite{luo2021diffusion} are probabilistic-based. 
		}
	\end{center}%
}]
\renewcommand{\thefootnote}{\fnsymbol{footnote}}
\footnotetext{This work was supported by the HK RGC Grant CityU 11202320 and 11218121. Corresponding author: J. Hou.}
\footnotetext[1]{Equal Contributions}

\begin{abstract}
	
   We propose WarpingGAN, an effective and efficient 3D point cloud generation network. Unlike existing methods that generate point clouds by directly learning the mapping functions between latent codes and 3D shapes, WarpingGAN learns a unified local-warping function to warp multiple identical pre-defined priors (i.e., sets of points uniformly distributed on regular 3D grids) into 3D shapes driven by local structure-aware semantics. In addition, we also ingeniously utilize the principle of the discriminator and tailor a stitching loss to eliminate the gaps between different partitions of a generated shape corresponding to different priors for boosting quality. Owing to the novel generating mechanism, WarpingGAN, a single lightweight network after one-time training, is capable of efficiently generating uniformly distributed 3D point clouds with various resolutions. Extensive experimental results demonstrate the superiority of our WarpingGAN over state-of-the-art methods in terms of quantitative metrics, visual quality, and efficiency. The source code is publicly available at \href{https://github.com/yztang4/WarpingGAN.git}{https://github.com/yztang4/WarpingGAN.git}.

\end{abstract}
\vspace{-2em}

\section{Introduction}

3D point clouds have been employed in various applications, such as  computer-aided design ~\cite{izadinia2020scene,koch2019abc}, augmented/virtual reality~\cite{liu2020arshadowgan}, animation~\cite{weng2019photo,huang2020arch}, and immersive telepresence~\cite{suo2021neuralhumanfvv}. However, obtaining 3D point cloud data is still costly and time-consuming in realistic scenarios, especially the shapes with complex geometry and  topology. Besides, the acquired point clouds with 3D sensing devices are usually incomplete and sparse due to occlusions, distances, and surface materials. The great success of generative adversarial network (GAN)-based 2D image generation \cite{goodfellow2014generative,brock2018large,zhu2017unpaired} makes synthesizing realistic-looking 3D point clouds promising, i.e., generating point clouds whose statistical distribution is similar to real point clouds. However, the essentially different data modality as well as the unique characteristics of 3D point clouds, i.e., the irregular structure and unorderness, makes it non-trivial to extend GAN-based methods for generating 2D images to 3D point cloud generation.

\begin{figure}[t]
\centering
\subfloat[\scriptsize TreeGAN]{
\includegraphics[width=0.7in]{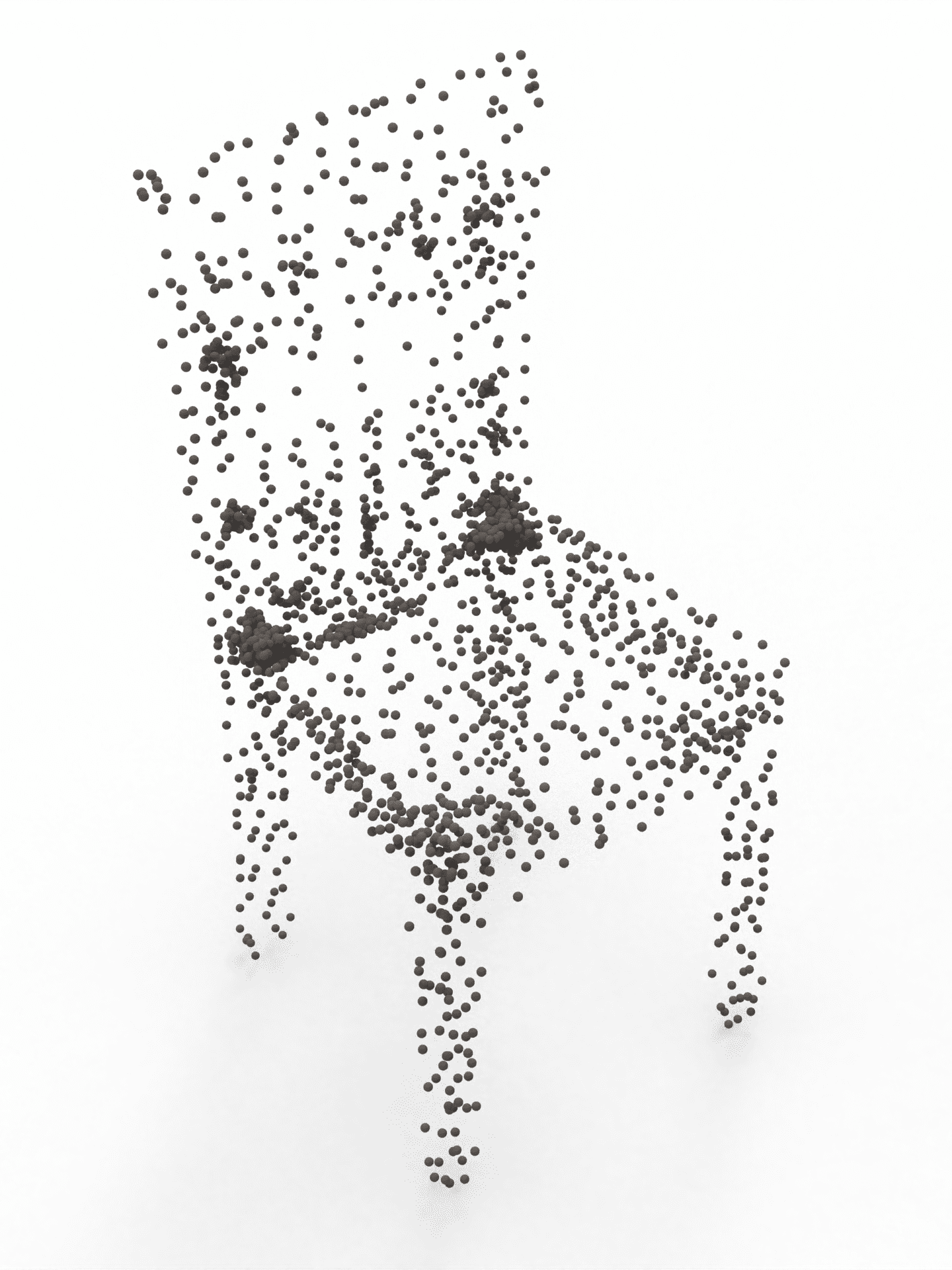}}
\label{uniform:treegan} 
\subfloat[\scriptsize PDGN]{
\includegraphics[width=0.7in]{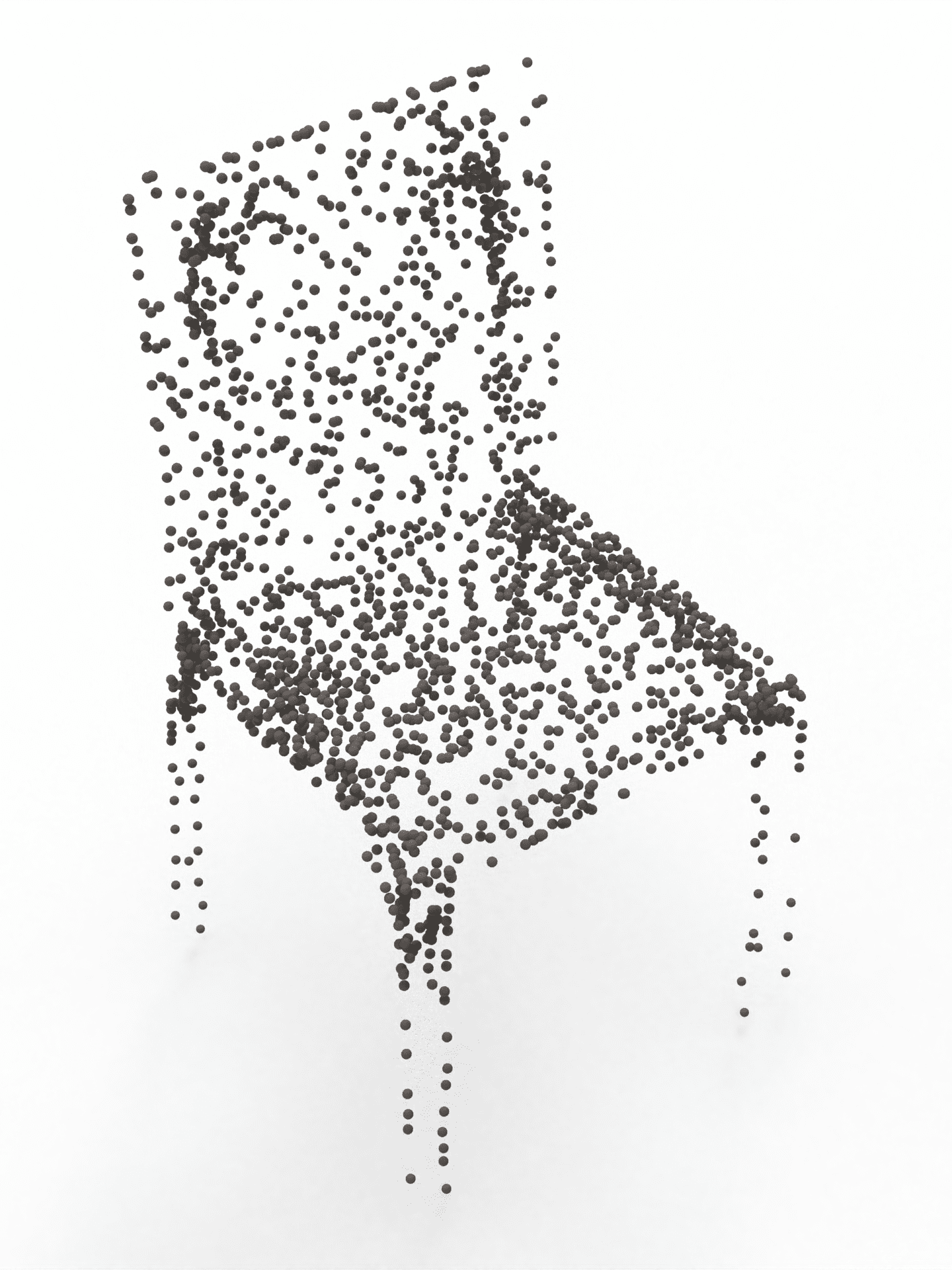}}
\label{uniform:pdgn} 
\subfloat[\scriptsize SP-GAN]{
\includegraphics[width=0.7in]{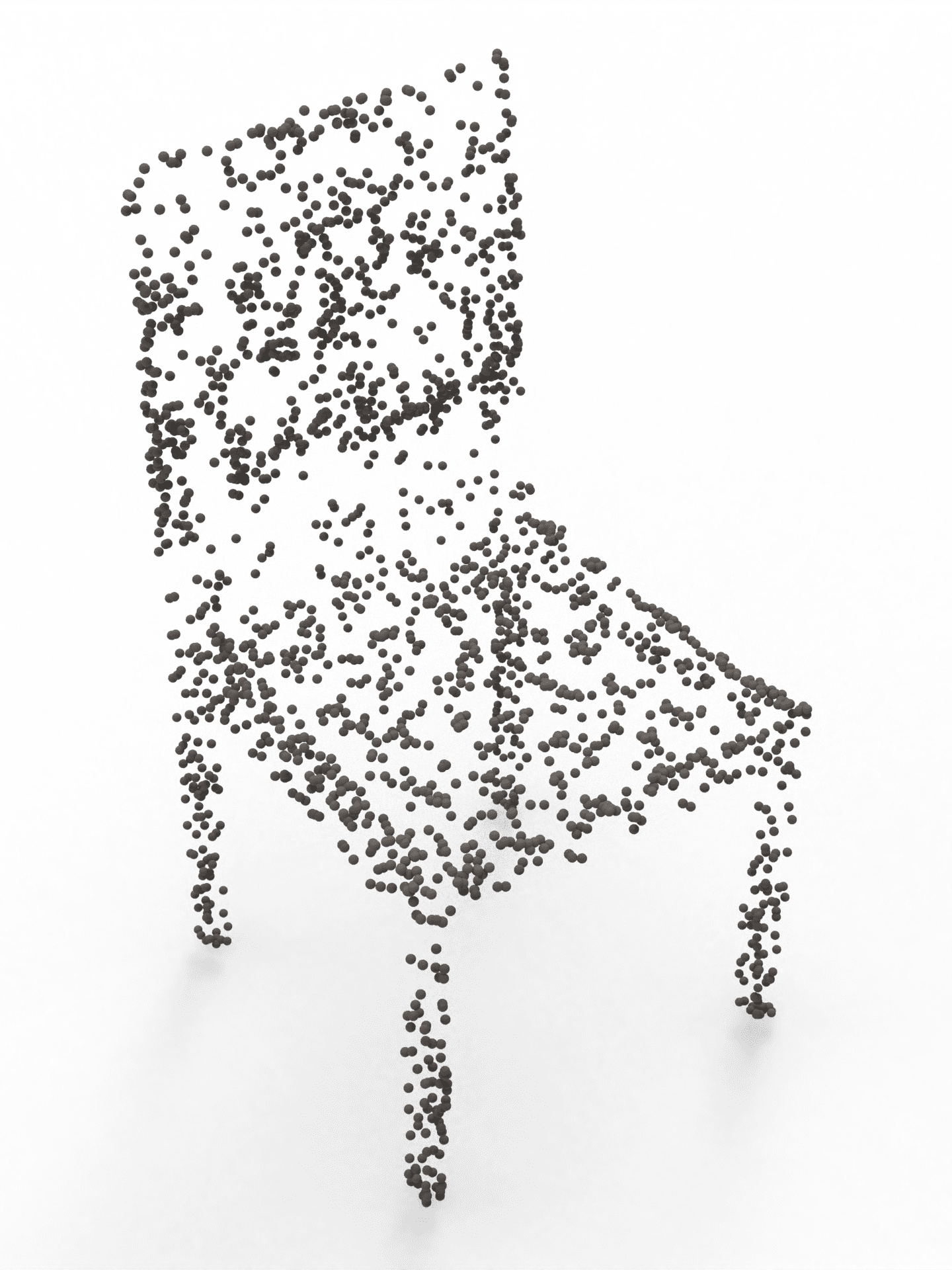}}
\label{uniform:spgan} 
\subfloat[\scriptsize WarpingGAN]{
\includegraphics[width=0.7in]{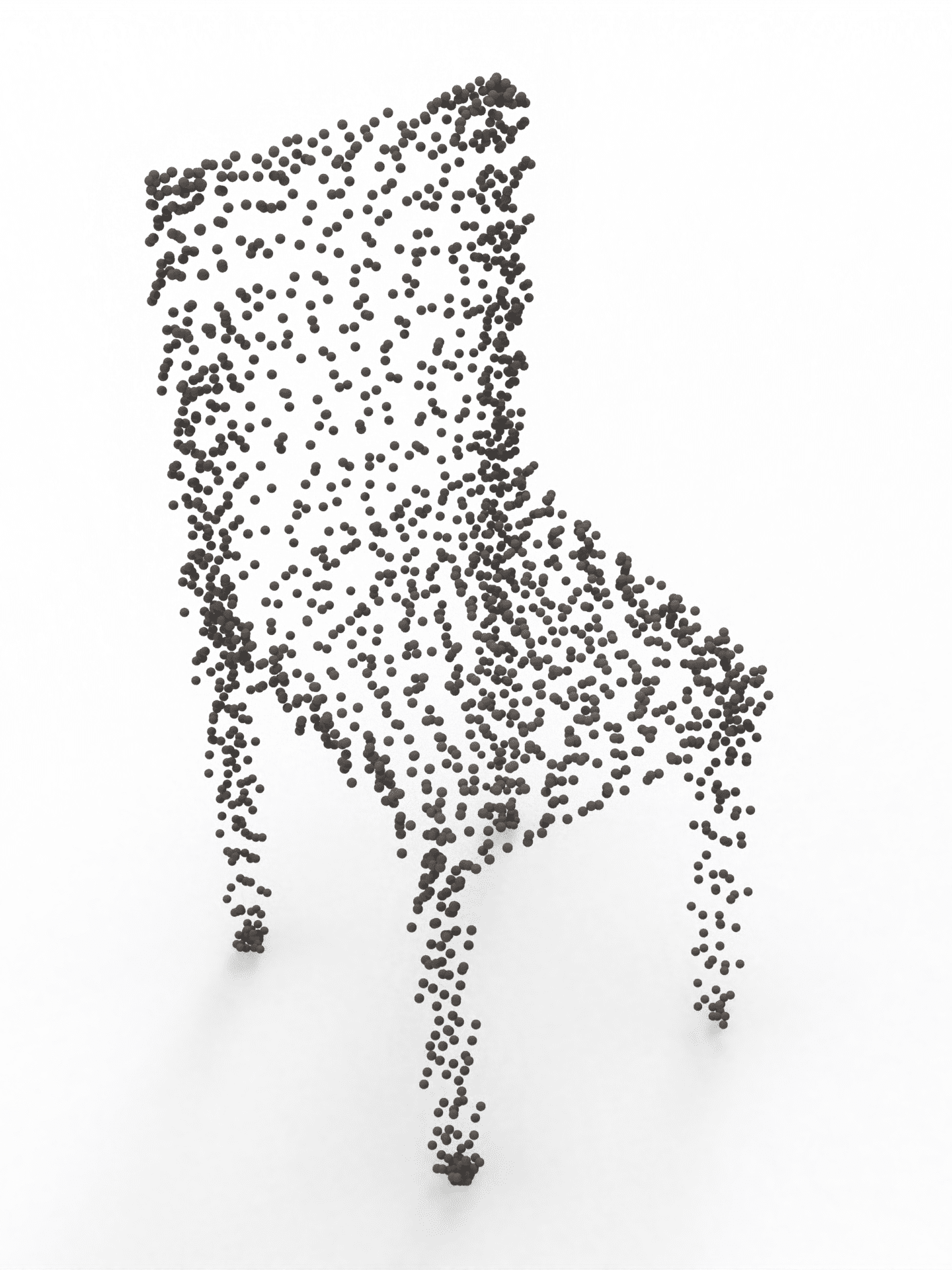}}
\label{uniform:warpinggan} 
\vspace{-0.2cm}
\caption{Visual comparisons of the point clouds generated by different GAN-based methods.}
\label{nonuniform} 
\vspace{-0.5cm}
\end{figure}

Recently, several works on 3D point cloud generation have been proposed \cite{shu20193d,li2021spgan,hui2020progressive,achlioptas2018learning,yang2019pointflow,valsesia2018learning,wen2021learning,kim2020softflow}. For example, GAN-based methods \cite{achlioptas2018learning,valsesia2018learning,shu20193d,hui2020progressive,wen2021learning} usually use multi-layer perceptrons (MLPs) as generators to directly learn mapping functions between latent codes and  3D point clouds, which require a large number of parameters to fit. Moreover, as the adversarial learning mechanism cannot impose strong constraints on global shapes and local geometric details, these approaches tend to generate non-uniformly distributed point clouds, as illustrated in Fig. \ref{nonuniform}. Yang \textit{et al}.\cite{yang2019pointflow} and Luo \textit{et al.} \cite{luo2021diffusion} considered 3D point cloud generation as probabilistic problems, which first sample points from a Gaussian space and then move them to the target position by learning the distribution transformation. However, these methods  generate blurry point clouds without clear global shapes and local details, since they tend to estimate the average distribution of training data. In addition to the limited quality, existing methods also suffer from low efficiency because time-consuming $k$ nearest neighbor ($k$NN) search is adopted in PDGN \cite{hui2020progressive} and SP-GAN \cite{li2021spgan}, a progressive generation process is utilized in TreeGAN \cite{shu20193d} and PDGN \cite{hui2020progressive}, and a two-stage training strategy is required in ShapeGF \cite{cai2020learning}, which also prohibits the end-to-end optimization.

To address the above-mentioned issues, we propose WarpingGAN, which introduces a novel mechanism for GAN-based 3D point cloud generation. By taking advantage of multiple 3D uniform priors, i.e., sets of points uniformly located on a unit 3D cube,  WarpingGAN formulates the generation process as the learning of a function that warps multiple 3D priors into different local regions of a 3D shape under the guidance of local structure-aware semantics, which is fundamentally different from existing methods that directly learn the process of producing a fixed number of points from the latent code. Meanwhile, we tailor a stitching loss, which minimizes the local difference between generated and real point clouds, to shrink the gaps between different partitions. Such a new mechanism makes WarpingGAN featured with compactness and high efficiency. Also, it enables WarpingGAN to generate point clouds with various numbers of points after one-time training. Besides, the uniformity of the 3D priors can implicitly regularize WarpingGAN to some extent to promote the generation of uniformly distributed point clouds, as shown in Fig. \ref{nonuniform}. 

In summary, we make the following contributions: 
\vspace{-0.3cm}
\begin{itemize}
	\setlength{\itemsep}{0pt}
	\setlength{\parsep}{0pt}
	\setlength{\parskip}{0pt}
	\item we investigate the GAN-based 3D point cloud generation from the new perspective of \textit{unified local-warping}, leading to WarpingGAN featured with  lightweight, high efficiency, and flexible output; and
    \item by taking advantage of the inherent design of the discriminator without introducing additional complex operations, we propose a stitching loss tailored to WarpingGAN to boost the generator; and 
    \item we conduct extensive experiments and analysis to demonstrate the superiority of WarpingGAN over state-of-the-art methods.
\end{itemize}


\section{Related Work}
\label{relatedwork}
  
Existing 3D point cloud generation methods can be roughly classified into two categories, i.e., GAN-based methods and Probabilistic-based. 

\textbf{GAN-based approaches.} 
As the first work, Achlioptas {\it et al.} \cite{achlioptas2018learning} proposed rGAN, whose generator consists of several fully connected layers. However, both the generator and discriminator of rGAN cannot well utilize the local information and tends to generate defective parts. Valsesia {\it et al.} \cite{valsesia2018learning} utilized a graph convolution network to learn local dependencies between a point and its neighbors by designing a localized operation. Shu \textit{et al}. proposed \cite{shu20193d} TreeGAN, where a tree structure is introduced to preserve ancestor information instead of neighbor information to generate new points. Hui {\it et al.} \cite{hui2020progressive} proposed a progressive learning strategy to generate multi-resolution point clouds, and a learning-based bilateral interpolation is utilized to exploit local geometric structure. The above methods employ MLPs on the global feature to directly generate point clouds, which can be hard to optimize and inflexible in terms of the number of points.  To simplify the learning process, TreeGAN and PDGN adopt the inefficient progressive architecture in the generator.

Recently, Li {\it et al.}~\cite{li2021spgan} proposed SP-GAN for point cloud generation and manipulation. They introduce a pre-defined sphere to perform deformation. By contrast, our method adopts multiple uniform 3D priors to warp each shape partition, enabling higher-quality point clouds. Furthermore, unlike SP-GAN, our architecture does not employ the time-consuming $k$NN operation in the generator. The experiment findings show that our proposed WarpingGAN is more effective and efficient than SP-GAN.

It is worth noting that Wang {\it et al.}~\cite{wang2020rethinking} experimentally found that the current GAN-based frameworks can only adopt PointNet as the discriminator. Other more advanced point cloud frameworks such as PointNet++~\cite{qi2017pointnetplusplus} and DGCNN~\cite{wang2019dynamic} are unable to be optimized as discriminators. However, PointNet learns point-wise features and uses the max-pooling symmetric function to select critical points to determine the global shape feature. Thus, only a small portion of critical points guide the global shape, and local geometric information is lost. Therefore, a well-designed generator is crucial for the GAN-based method.

\textbf{Probabilistic-based approaches} regard point clouds as samples from a distribution, and then move the sampled points to the target positions during the generative phase. PointFlow  \cite{yang2019pointflow} uses the continuous normalizing flow framework to transform the parameters of the distributions of shapes and the distribution of points given a shape. ShapeGF  \cite{cai2020learning} generates point clouds by learning the gradient field of its log-density and moves points gradually in the gradient direction. DPM \cite{luo2021diffusion} simulates the generation process as non-equilibrium thermodynamics by converting the noise distribution into the shape distribution with a Markov chain. Some of them \cite{yang2019pointflow,luo2021diffusion} can accomplish flexible generation because they treat each point independently. However, due to the absence of association between the points, they cannot properly settle the non-uniform and noisy problems for generated shapes. Moreover, many probabilistic-based methods \cite{cai2020learning} adopt a two-stage training process, which requires additional training for auto-encoders. It's worth noting that ShapeGF uses the GAN structure as well, not only the auto-encoder. 

\textbf{Point cloud auto-encoders}
aim to reconstruct the input point cloud with a narrow bottleneck layer. For example, FoldingNet \cite{yang2018foldingnet} introduces a folding-style operation to reconstruct 3D shapes by learning a mapping function from a 2D grid to a 3D point cloud. AtlasNet \cite{groueix2018papier} utilizes a series of 2D grids via multiple independent MLPs to reconstruct surfaces patch-wisely. Bednarik {\it et al.} \cite{bednarik2020shape} addressed patch collapse and overlap problems of AtlasNet by computing the differential properties of reconstructed surface with first and second derivatives. 

\textit{\textbf{Remark}}. We argue that such folding-style decoders are potentially suitable for the GAN-based point cloud generation task, which have been ignored in previous frameworks. In this work, we take a step forward in this direction by investigating a more powerful point cloud generator. Note that directly adopting the decoders of FoldingNet and AtlasNet as the generator of a GAN-base point cloud generation framework fail to generate satisfied point clouds. See the analysis in Section \ref{sec:formulation} and experimental demonstration in Section \ref{sec:ablation}.


\begin{figure*}[t]
	\centering
	\setlength{\abovecaptionskip}{-0.1cm}
	\setlength{\belowcaptionskip}{-0.6cm}
	\includegraphics[width=5.5 in]{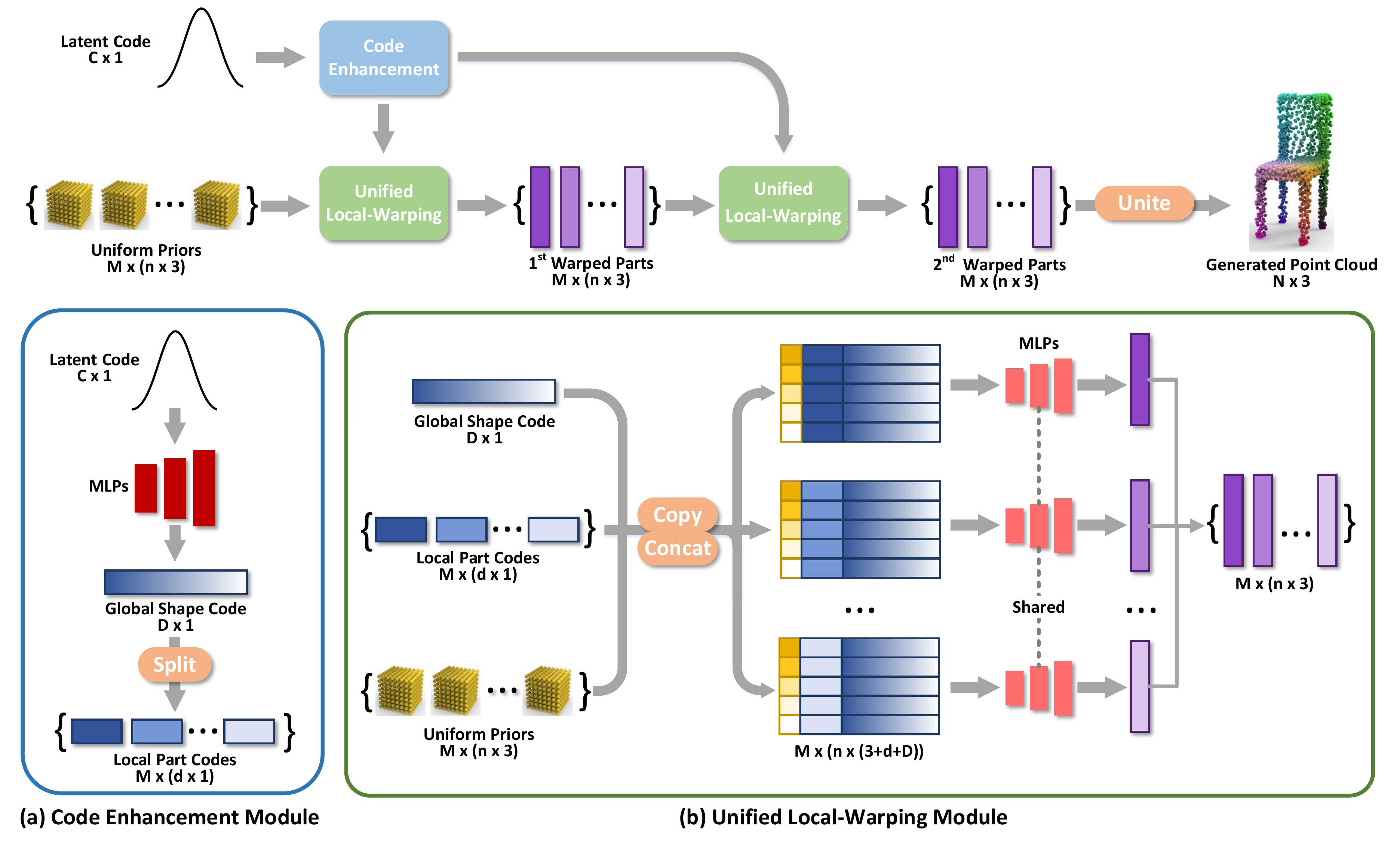}
	\caption{Illustration of the architecture of the \textbf{\textsl{generator}} of WarpingGAN, which consists of a code enhancement module  and a unified local-warping module. During training, it takes a latent code and  $M$ pre-defined uniform 3D priors as input. The unified local-warping module is performed twice to generate a 3D shape. }
	\label{flow}
\end{figure*}

\section{Proposed Method}

\subsection{Problem Analysis and Formulation}\label{sec:formulation}
Given a latent code $\mathbf{z}\in\mathbb{R}^C$  following a Gaussian distribution, the generator denoted as $\mathcal{F}(\cdot)$ attempts to produce a point cloud $\mathbf{P}\in\mathbb{R}^{N\times 3}$ which shares the same statistical distribution as a real dataset $\{\mathbf{\widetilde{P}}\}$. Most of existing GAN-based methods directly learn the mapping function between $\mathbf{z}$ and $\mathbf{P}$, i.e., $\mathcal{F}(\mathbf{z};{\bm \Theta})=\mathbf{P}$ with ${\bm \Theta}$ being the network parameters to be learned. However, due to the weak supervision ability of the discriminator, it is difficult to learn $\mathcal{F}(\cdot)$, especially for the dataset with complex shapes, thus limiting the quality of generated point clouds.  More specifically, Wang \textit{et al}.~\cite{wang2020rethinking} analyzed that as the only feasible discriminator, the PointNet architecture \cite{qi2017pointnet}, which applies the max-pooling operator to sample the feature space, can only perceive the distribution of a small number of critical points rather than the whole point cloud.
 
Motivated by the the folding-style design mentioned earlier, we consider warping a uniform 3D prior (i.e., a set of 3D points uniformly located in a regular 3D grid) denoted as $\mathbf{U}\in[0,1]^{N\times 3}$ into a 3D shape. Accordingly,  we reformulate the generation process point-wisely as 
$\mathcal{F}(\mathbf{u}_i, \mathbf{z}; {\bm \Theta})=\mathbf{p}_i~\forall i\in[1, N]$,  where $\mathbf{u}_i$ and $\mathbf{p}_i$ are  the $i$-th point of $\mathbf{U}$ and $\mathbf{P}$, respectively.
We expect that the pre-defined uniformity of the 3D prior could regularize the distribution of all of the generated points to mitigate the limitation of the max-pooling-based discriminator mentioned earlier. However, a single prior may fail to generate complex shapes well due to the essential difference between the topology of the prior and the 3D objects. Thus, we plan to use multiple uniform 3D priors denoted as $\left\{\mathbf{U}^j\in[0,1]^{n\times 3}\right\}_{j=1}^M$ and warp each of them to capture the local region (i.e., $\mathbf{P}^j\in\mathbb{R}^{n\times 3}$) of a generated point cloud $\mathbf{P}=\bigcup_{j=1}^M \mathbf{P}^{j}$, where $N=M\times n$. To achieve this, like AtlasNet~\cite{groueix2018papier}, one intuitive way is to learn a warping process for each pair of $\mathbf{U}^j$ and $\mathbf{P}^j$ independently, i.e., $\mathbf{p}_i^j=\mathcal{F}(\mathbf{u}_i^j, \mathbf{z}; {\bm \Theta}^{j})$, where $\mathbf{u}_i^j$ and $\mathbf{p}_i^j$ are the $i$-th points of $\mathbf{U}^j$ and $\mathbf{P}^j$, respectively. However, this manner significantly increases network parameters $\{ \bm \Theta^{j} \}_{j=1}^M$, making the network difficult to train. 

To achieve the generation in an efficient and effective manner,  we finally formulate it as a unified local-warping process. That is, we use $M$ different global-correlated local codes $\{\mathbf{z}^j\}_{j=1}^M$ (see Section \ref{subsec:generator} for details), each of which is expected to embed the semantic of a typical local region. Thus, a local code can drive the warping of a prior to the corresponding structure via an \textit{identical} MLPs. Accordingly, the process is generally written as 
 \begin{equation}
 \setlength{\abovedisplayskip}{2pt}
\setlength{\belowdisplayskip}{3pt}
\mathbf{p}^{j}_i=\mathcal{F}(\mathbf{u}^{j}_i,\mathbf{z}^{j};{\bm \Theta}), \forall j\in[1, M]~{\rm and}~\forall i\in [1, n].
\label{eq:form}
\end{equation} 
Compared with the AtlasNet-based idea, our warping process only requires unified parameters ${\bm \Theta}$ to be optimized, which is more compact and easier. Moreover, although we expect to use multiple priors rather than a single one improve generation quality, the weak supervision ability of the discriminator may not realize our objective, i.e., the supervision is insufficient to drive the local regions warped from different priors to tightly fit to each other, resulting in gaps between them. To handle this issue, we further tailor a simple yet effective stitching loss to supervise the training of the generator. 

\subsection{Warping-based Generator}
\label{subsec:generator}

Fig. \ref{flow} illustrates the overall architecture of the generator of our WarpingGAN, which consists of two modules, i.e., code enhancement and unified local warping. Specifically, taking a latent code $\mathbf{z}$ as input, the code enhancement module first enhances its representation ability to 3D shapes by changing its distribution. Conditioned on the enhanced code, the unified local-warping module is then successively performed twice to warp multiple pre-defined uniform 3D priors to different local regions, which are finally assembled into a 3D shape. Owing to the warping-style mechanism, WarpingGAN is featured with highly compact and efficient. Besides, WarpingGAN is able to generate point clouds with various numbers of points by changing the size of $\mathbf{U}^{j}$ (i.e., the value of $n$), after one-time training.

\textbf{Code enhancement.} 
 As aforementioned, 
 WarpingGAN aims to learn a warping function under the guidance of the latent code $\mathbf{z}$. However, $\mathbf{z}$ 
 is randomly drawn from a Gaussian distribution and thus lacks the implicit semantic information to represent the shape faithfully. 
 To fill this knowledge gap, we propose a code enhancement module shown in Fig. \ref{flow}(a), composed of five fully-connected layers to transform $\mathbf{z}$ to $\widetilde{\mathbf{z}}\in\mathbb{R}^D$ of a higher dimension ($D>C$). In Section~\ref{sec:ablation}, we illustrate that after the data-driven training process,  such a module can transform the 
 Gaussian distribution to a distribution that is comparable to that of the features extracted from real datasets that encode shape semantics.

\textbf{Unified local-warping.} 
Guided by the enhanced latent code $\widetilde{\mathbf{z}}$, the unified local-warping module in Fig. \ref{flow}(b) can generate from multiple 3D priors various local regions that are further assembled into a point cloud with complex topology. Specifically, we first partition $\widetilde{\mathbf{z}}$ into  $M$ local codes of an equal length denoted as $\{\widetilde{\mathbf{z}}^j\in\mathbb{R}^{D/M}\}_{j=1}^M$ and then concatenate each of the local codes with the global shape information, leading to  $\mathbf{z}^j=[\widetilde{\mathbf{z}}^{j}~\widetilde{\mathbf{z}}]\in\mathbb{R}^{D(M+1)/M}$. Finally, we concatenate $\mathbf{z}^j$ with each coordinate of $\mathbf{U}^{j}$, which is then fed into an MLPs to regress the points of the $j$-th local region in point-wise. Moreover, we perform such a warping process twice in a row, therefore, such a unified local-warping process is written as
\setlength{\abovedisplayskip}{0pt}
\setlength{\belowdisplayskip}{3pt}
\begin{align}
  \mathbf{p}^{j}_i =\mathcal{F}\left(\mathcal{F}(\mathbf{u}^{j}_i,\mathbf{z}^j;\bm \Theta_{1}),\mathbf{z}^j;\bm \Theta_{2}\right),& \\~~~~~~~~~~~~~~\forall j\in[1, M] &~{\rm and}~\forall i\in [1, n], \nonumber
  \label{eq:block12}
\end{align}
where ${\bm \Theta}_{1}$ and ${\bm \Theta}_{2}$ are the network parameters of the two consecutive warping processes.  
Note that it is crucial to concatenate $\widetilde{\mathbf{z}}$ in the local codes since $\widetilde{\mathbf{z}}$ provides the essential global shape information to coordinate different local codes (see the demonstration in Section \ref{sec:ablation}).

\begin{figure}[h]
\centering
\setlength{\abovecaptionskip}{-0.1cm}
\setlength{\belowcaptionskip}{-0.3cm}
\includegraphics[width=3.3in]{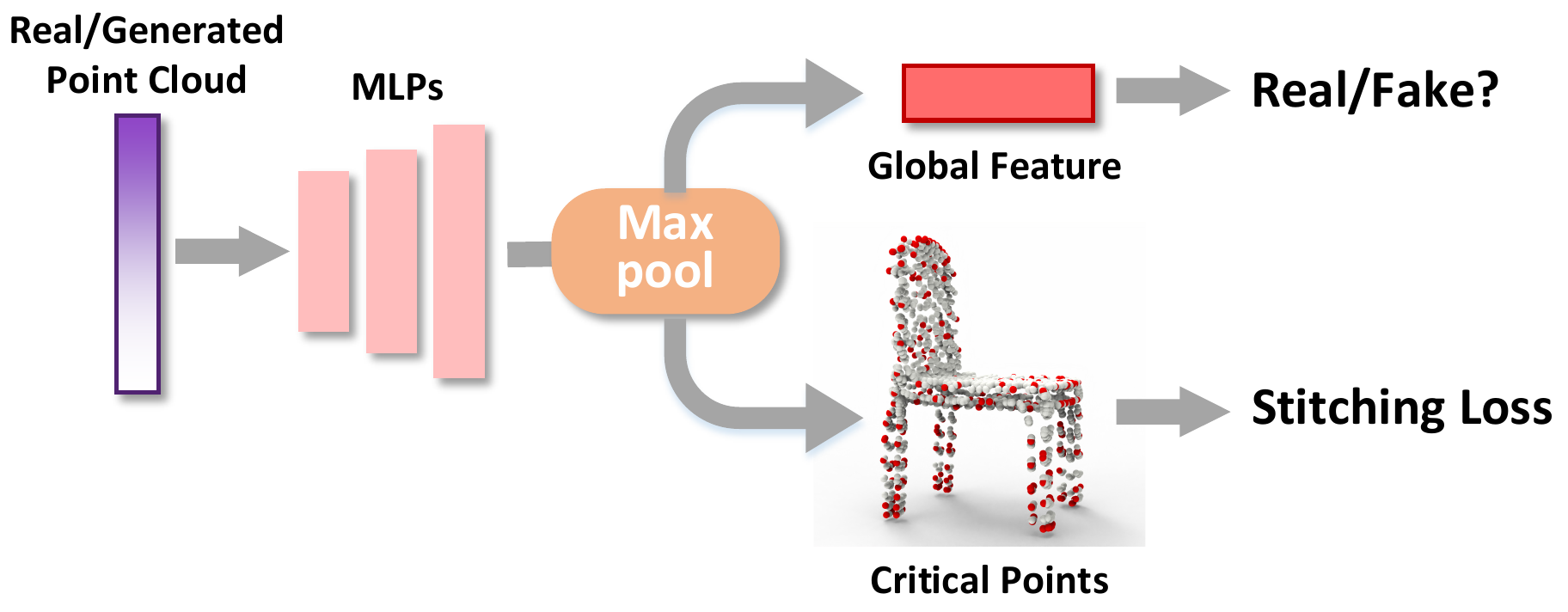}
\caption{The architecture of the discriminator of WarpingGAN. It takes real point clouds or  generated point clouds as input and  produces the confidence value and the critical points (i.e., \textcolor{red}{red} points) that are used by the stitching loss.}
\label{discriminator}
\end{figure}
\subsection{Training Objectives}

\textbf{Discriminator.}
Following previous works \cite{shu20193d,hui2020progressive}, we adopt the PointNet as the discriminator $\mathcal{D}(\cdot)$ for training the generator. As shown in Fig. \ref{discriminator}, it takes $\{\widetilde{\mathbf{P}}\}$ as input to perform binary classification. More specifically, it first learns point-wise features via MLPs and then adopts max-pooling to obtain a global shape feature that is utilized to determine the confidence value.  Meanwhile, owing to the max-pooling operation, we can retrieve a small number of \textit{critical points} from the input cloud, which depict the skeleton of the input shape \cite{qi2017pointnet}.

\textbf{Stitching loss.}
As analyzed in Section~\ref{sec:formulation}, the inherent design of the discriminator results in a limited supervision ability, which is inadequate to force the generated local regions to fit each other tightly. To tackle this issue, by taking advantage of critical points without introducing additional complex operations, we propose a stitching loss, which minimizes the local difference between $\mathbf{P}$ and $\widetilde{\mathbf{P}}$ .

Let $\{\mathbf{q}_i\}_{i=1}^Q\subset \mathbf{P}$ and $\{\mathbf{\widetilde{q}}_i\}_{i=1}^{\widetilde{Q}} \subset \mathbf{\widetilde{P}}$ denote the critical points retrieved from the input point clouds based on the max-pooling operation of the discriminator. 
We first find out the $K$ nearest neighbors $\mathcal{N}(\mathbf{q}_i)=\{\mathbf{p}_i^{k}\}_{k=1}^K\subset \mathbf{P}$ for each of $\{\mathbf{q}_i\}_{i=1}^Q$
and compute the pairwise distance $d_i^{k}=\|\mathbf{q}_i-\mathbf{p}_i^{k}\|_2$. We also conduct the same operation for $\{\mathbf{\widetilde{q}}_i\}_{i=1}^{\widetilde{Q}}$.  
With this information, we define the stitching loss as 
\begin{equation}
\setlength{\abovedisplayskip}{-0pt}
\setlength{\belowdisplayskip}{3pt}
\mathcal{L}_{s} = \left(\frac{1}{Q}\sum_{i=1}^{Q}\text{var}(\mathbf{q}_i, \mathcal{N}(\mathbf{q}_i))-\frac{1}{\widetilde{Q}}\sum_{i=1}^{\widetilde{Q}}\text{var}(\mathbf{\widetilde{q}}_i, \mathcal{N}(\mathbf{\widetilde{q}}_i))\right)^2,
\label{eq:Stitch}
\end{equation}
where $\text{var}(\mathbf{q}_i, \mathcal{N}(\mathbf{q}_i))=\sum_{k=1}^{K}\frac{(d_i^{k}-\bar{d_i})^2}{K}$ and $\bar{d_i}$ is the mean value of $\{d_i^{k}\}_{i=1}^Q$. 

Note that the stitching loss requires the $k$NN operation over the very small number of critical points only during training, and thus, the high efficiency of the proposed method is still retained during both training and testing. See Section \ref{subsec:comparison}.

\begin{figure*}[t]
\centering
\setlength{\abovecaptionskip}{0cm}
\setlength{\belowcaptionskip}{-0.4cm}
\includegraphics[width=5.8in]{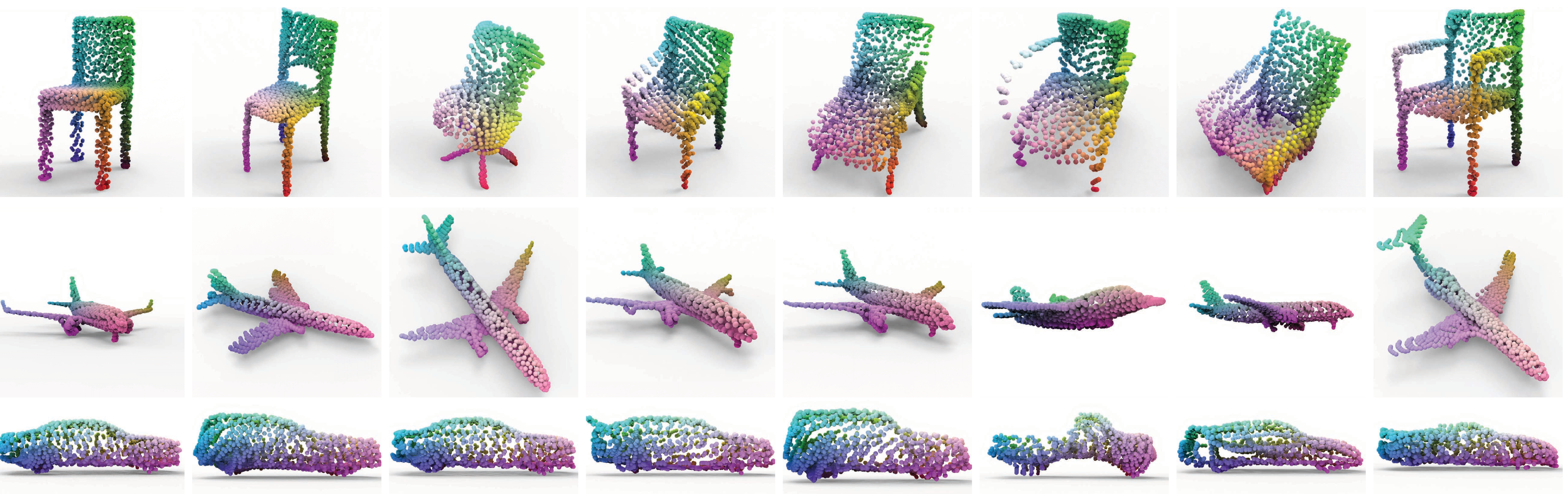}
\caption{Visual illustration of the generated \textit{Chair}, \textit{Airplane} and \textit{Car} shapes by our WarpingGAN. These shapes have fine global structures and present a variety of geometric typology. See \textit{Supplementary Material} for more visual results.}
\label{visualresults}
\end{figure*}
\begin{figure*}[t]
\centering
\setlength{\abovecaptionskip}{0cm}
\setlength{\belowcaptionskip}{-0.4cm}
\includegraphics[width=5.5in]{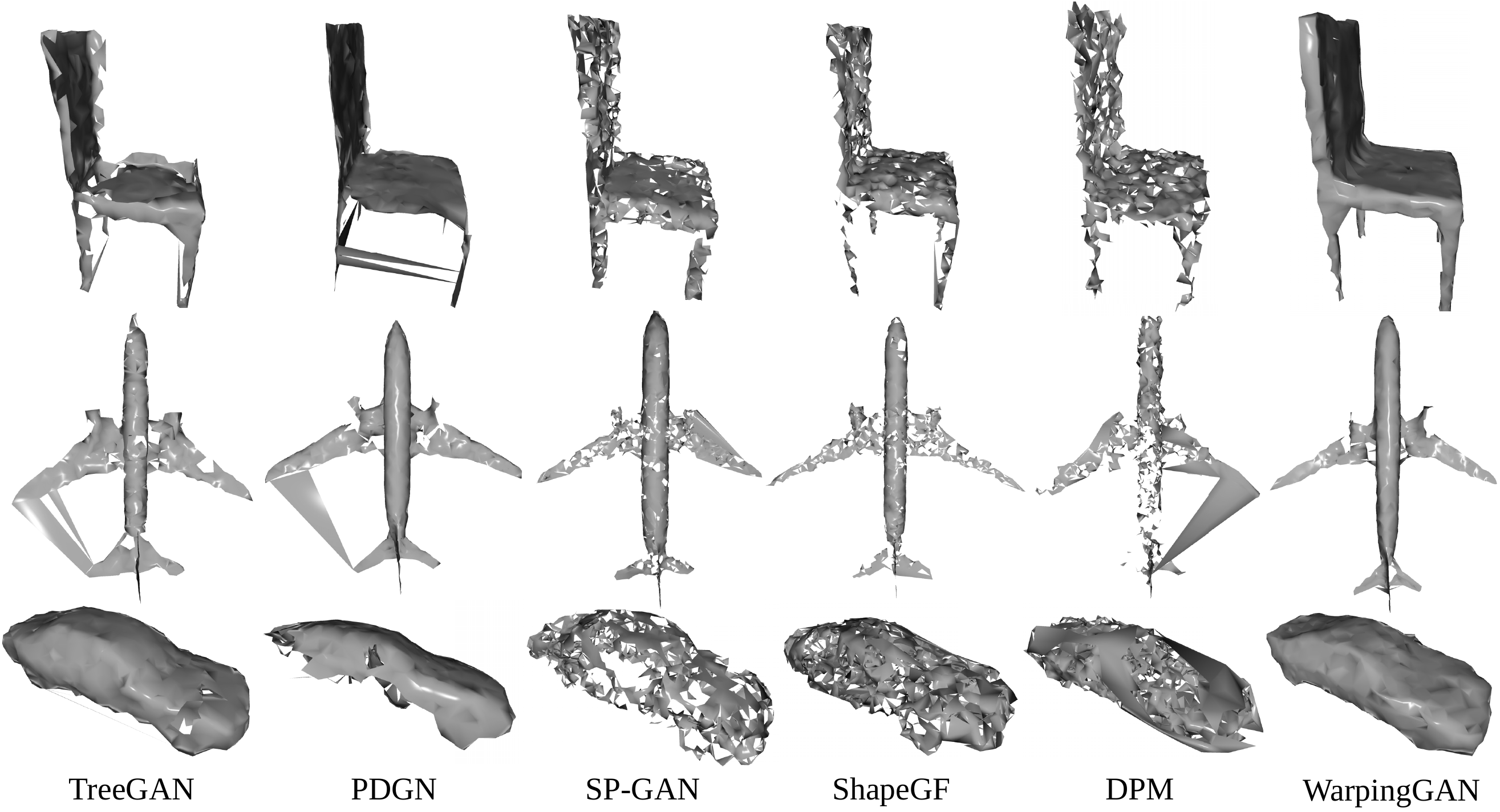}
\caption{Visual comparison of the reconstructed 3D surfaces from the point clouds (shown in \ref{sotacompare}) generated by different methods via the ball pivoting algorithm. 
}
\label{mesh}
\end{figure*}
\textbf{Joint optimization.} To train the proposed WarpingGAN, we adopt the improved WGAN loss \cite{gulrajani2017improved}, which consists of the loss $\mathcal{L}_{g}(\cdot)$ for the generator and $\mathcal{L}_{d}(\cdot)$ with the Lipschitz constraint for the discriminator. More precisely,
$\mathcal{L}_{g}(\cdot)$
is written as
\begin{equation}
\setlength{\abovedisplayskip}{-0pt}
\setlength{\belowdisplayskip}{3pt}
\mathcal{L}_{g} = -\mathbb{E}_{\mathbf{P} \sim \mathbb{P}_{\mathbf{P}} }[\mathcal{D}(\mathbf{P})] + \lambda_s \mathcal{L}_s,
\end{equation}
where
$\mathbb{P}_{\mathbf{P}}$ is the distribution of generated shape $\mathbf{P}$ and $\mathcal{L}_s$ is the proposed stitching loss which is balanced by the weight $\lambda_s>0$. 
$\mathcal{L}_{d}(\cdot)$ is formulated as
\begin{equation}
\setlength{\abovedisplayskip}{-0pt}
\setlength{\belowdisplayskip}{3pt}
\begin{aligned}
\mathcal{L}_{d} &= \mathbb{E}_{\mathbf{P} \sim \mathbb{P}_{\mathbf{P}} }[\mathcal{D}(\mathbf{P})]-\mathbb{E}_{\mathbf{P} \sim \mathbb{P}_{\mathbf{\widetilde{P}}} }[\mathcal{D}(\mathbf{\widetilde{P}})]\\
&+\lambda_{gp} \mathbb{E}_{\widehat{\mathbf{p}}\sim\mathbb{P}_{\widehat{\mathbf{P}}}}[(\|\triangledown_{\widehat{\mathbf{p}}}\mathcal{D}(\widehat{\mathbf{p}})\|_2-1)^2],
\end{aligned}
\end{equation}
where $\mathbb{P}_{\widetilde{\mathbf{P}}}$ is the distribution of the real point cloud $\widetilde{\mathbf{P}}$, and $\widehat{\mathbf{p}}$ is uniformly sampled by interpolating pairs of shapes sampled from $\mathbb{P}_{\mathbf{P}}$ and $\mathbb{P}_{\widetilde{\mathbf{P}}}$ to satisfy the 1-Lipschitz constraint \cite{gulrajani2017improved}, and  $\lambda_{gp}>0$ the weight to balance the  gradient penalty term.

\section{Experiments}
\subsection{Experiment Settings} 

\textbf{Dataset.}
Following the settings in \cite{shu20193d}, we selected three categories of ShapeNet, i.e.,  \textit{Chair}, \textit{Airplane} and \textit{Car shapes}, to train and evaluate WarpingGAN. Each point cloud contains $2048$ points. 

\parskip=0pt
\textbf{Implementation details.} WarpingGAN samples latent codes of dimension $C=128$ following a Gaussian distribution as input to generate point clouds each with $N=2048$ points. We set the number of priors $M$ to $16$ for all shapes, $K=40$ for computing the stitching loss, and $\lambda_s=0.05$ and $\lambda_{gp}=10$ during the training phase. We adopted LeakyReLU with a negative slope equal to 0.2 in each layer. We utilized Adam with the learning rate $r=0.0001$, $\beta_1=0$ and $\beta_2=0.99$ as the optimizer to optimize both the generator and discriminator, and set the batch size to 32. We implemented the whole network with PyTorch and trained it on Nvidia RTX 2080ti GPU with Intel(R) Xeon(R) CPU.

\begin{table}[t]
  \centering
  \caption{Quantitative comparison of WarpingGAN with five state-of-the-art methods over two categories. 
  The listed MMD and COV values were obtained by multiplying the original values with $10^{3}$ and $10^{2}$, respectively. 
  $\uparrow$ (resp. $\downarrow$) means the higher (resp. lower), the better.}
  \vspace{-0.2cm}
  \resizebox{\linewidth}{!}{
    \begin{tabular}{l|ccc|ccc}
    \hline
    \hline
    \multicolumn{1}{c|}{\multirow{2}[0]{*}{\diagbox{Method}{Metric}}} & \multicolumn{3}{c|}{Chair} & \multicolumn{3}{c}{Airplane} \\
          & \multicolumn{1}{c}{MMD$\downarrow$} & \multicolumn{1}{c}{COV$\uparrow$} & \multicolumn{1}{c|}{Uniform$\downarrow$} & \multicolumn{1}{c}{MMD$\downarrow$} & \multicolumn{1}{c}{COV$\uparrow$} & \multicolumn{1}{c}{Uniform$\downarrow$} \\
          \hline
    TreeGAN \cite{shu20193d} & 9.6 & 45.00 & 0.88  & 3.8  & 42.50 & 0.45 \\
    PDGN \cite{hui2020progressive} & 9.3 & 51.25 & 0.85  & 3.4  & 41.25 & 0.21 \\
    SP-GAN \cite{li2021spgan} & 11.5  & 41.25  & 0.34  & 3.5  & 46.25  & 0.05 \\
    ShapeGF \cite{cai2020learning} & 9.6 & 50.00 & 0.64  & 3.5  & 47.50 & 0.09 \\
    DPM \cite{luo2021diffusion}  & 9.4 & 37.50 & 1.45  & 3.4 & 33.75 & 0.35 \\
    WarpingGAN & \textbf{8.7}   & \textbf{53.75}    & \textbf{0.29}  & \textbf{3.3}     & \textbf{48.75}    & \textbf{0.02} \\
    \hline
    \hline
    \end{tabular}%
    }
  \label{sotacompare}%
  \vspace{-0.3cm}
\end{table}%

\subsection{Comparison with State-of-the-Art Methods} \label{subsec:comparison}

We compared the proposed WarpingGAN with five state-of-the-art point cloud generation frameworks, including three GAN-based methods, i.e., TreeGAN~\cite{shu20193d}, PDGN~\cite{hui2020progressive} and SP-GAN~\cite{li2021spgan}, and two probablistic-based methods, i.e.,  ShapeGF~\cite{cai2020learning} and DPM~\cite{luo2021diffusion}. 

\textbf{Quantitative comparison.} Following the settings of SP-GAN~\cite{li2021spgan}, we utilized Minimal Matching Distance (MMD) and Coverage (COV) to quantitatively evaluate the quality of generated point clouds by different methods. 

Besides, we also adopted the uniformity loss\footnote{We measured the normalized point clouds with various percentages of points, i.e., $p\in\{0.002, 0.004, 0.006, 0.008, 0.012, 0.015\}$. We reported the average uniformity over all $p$.} in \cite{li2019pu} to quantitatively measure the uniformity of generated point clouds. 
 
As listed in Table \ref{sotacompare}, our WarpingGAN outperforms all the other methods in terms of all metrics. Specifically, the lower MMD values imply that WarpingGAN can generate shapes with high fidelity to point clouds in the real dataset, the high COV values demonstrate that the generated shapes of WarpingGAN match well with the real shapes in terms of fraction, and the lower Uniform values indicate our WarpingGAN can generate point clouds with more uniformly distributed points. Moreover, we want to point out that metrics MMD and COV do not necessarily and reliably correlate to the quality of generated data, which has also been discussed in \cite{yang2019pointflow,Zhou_2021_ICCV}. Thus, we refer readers to examine visual quality of generated point clouds provided as follows and in \textit{Supplementary Material}.

\textbf{Visual comparison.} We visualized the generated 3D point clouds by different methods. Besides, we also reconstructed 3D meshes from them via the ball pivoting\footnote{For a fair comparison, the same hyper-parameters were applied to the generated point clouds by different methods.} algorithm \cite{bernardini1999ball}. 
As illustrated in Fig. \ref{sotacompare}, WarpingGAN generates shapes with finer global shapes and local details than the other methods. 
Particularly, the points of the generated data by our WarpingGAN are uniformly distributed, thus avoiding ``holes" which appear in the results by compared methods. Besides,  WarpingGAN does not generate outlier points. 
From Fig. \ref{mesh}, it can be observed that the reconstructed 3D meshes from the point clouds generated by our WarpingGAN have much better quality than those by other methods, which also 
validates the higher quality of the point clouds generated by our WarpingGAN. 
See \textit{Supplementary Material} for more visual results.

\textbf{Efficiency comparison.} We also compared the training time, inference time and parameter sizes of different methods. We only reported the inference time of ShapeGF and DPM. As listed in Table \ref{param}, during interface, WarpingGAN achieves the most compact model size and  $4$ $\sim$ $300$ faster than state-of-the-art methods, owing to the exclusion of the time-consuming $k$NN and progressive design. 
Although the stitching loss of WarpingGAN requires calculating $k$NN for subsets with a small size during training, which slightly increases the training time,  
the training process of WarpingGAN is still much more efficient than PDGN and SP-GAN.

\begin{table}[t]
	\centering
	\caption{Comparison of the training time, inference time and parameter size of different methods. The training time refers to the average time of one iteration.}
	\vspace{-0.2cm}
	\resizebox{.422\textwidth}{!}{
		\begin{tabular}{l|ccc}
			\hline\hline
			Method & \multicolumn{1}{c}{Training (s)} & \multicolumn{1}{c}{Inference (s)} & Params (M) \\
			\hline
			TreeGAN \cite{shu20193d} & \textbf{0.04}  & 0.014 & 40.69 \\
			PDGN \cite{hui2020progressive} & 0.63  & 0.077 & 12.71 \\
			SP-GAN \cite{li2021spgan} & 0.29  & 0.031 & 0.59 \\
			ShapeGF \cite{cai2020learning} & \multicolumn{1}{c}{-} & 2.660  & 4.40 \\
			DPM \cite{luo2021diffusion}  & \multicolumn{1}{c}{-} & 0.641 & 1.58 \\
			WarpingGAN & 0.08  & \textbf{0.008} & \textbf{0.58} \\
			\hline\hline
		\end{tabular}%
	}
    \vspace{-0.5cm}
	\label{param}%
\end{table}%

\begin{figure}[t]
\centering
\setlength{\abovecaptionskip}{0.1cm}
\setlength{\belowcaptionskip}{-0.6cm}
\includegraphics[width=2.5in]{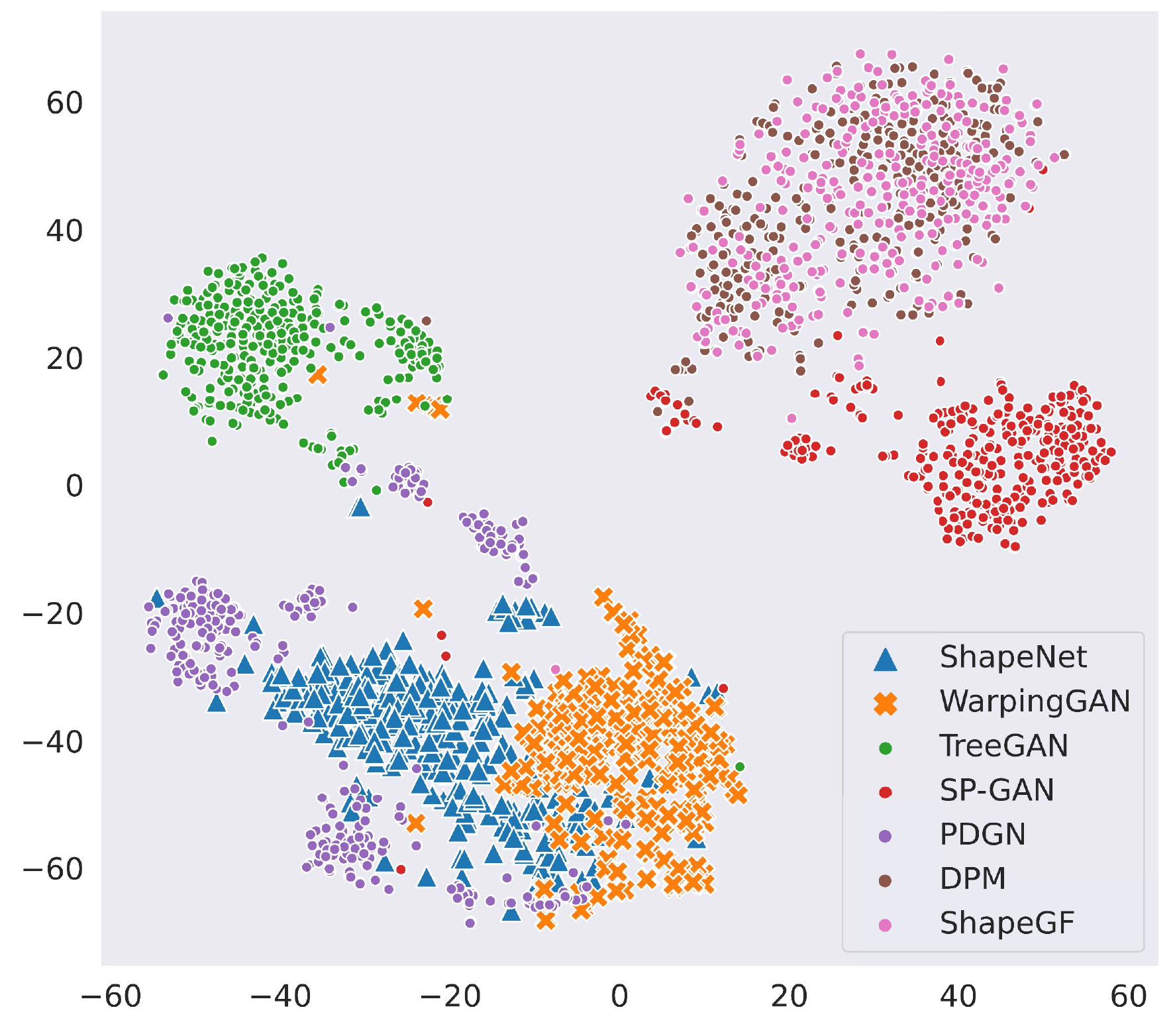}
\caption{Visual illustration of the t-SNE feature clustering of the real point clouds of ShapeNet and generated point clouds by different methods.}
\label{tsne}
\end{figure}

\textbf{Feature comparison.}
We also employed t-SNE \cite{van2008visualizing} to visually compare the \textit{Airplane} shapes generated by different methods in the feature space. Specifically, we adopted the pre-trained DGCNN with ModelNet40 to extract features of generated point clouds by different methods and the corresponding \textit{Airplane} category of ShapeNet. As shown in Fig. \ref{tsne},  compared with other methods, the feature distribution of WarpingGAN is \textit{more compact} and \textit{closer} to that of the real point cloud set, indicating that the generated point clouds by our WarpingGAN are more realistic.  

\begin{figure}[h]
\centering
\subfloat[1024 points]{
\includegraphics[width=0.8in]{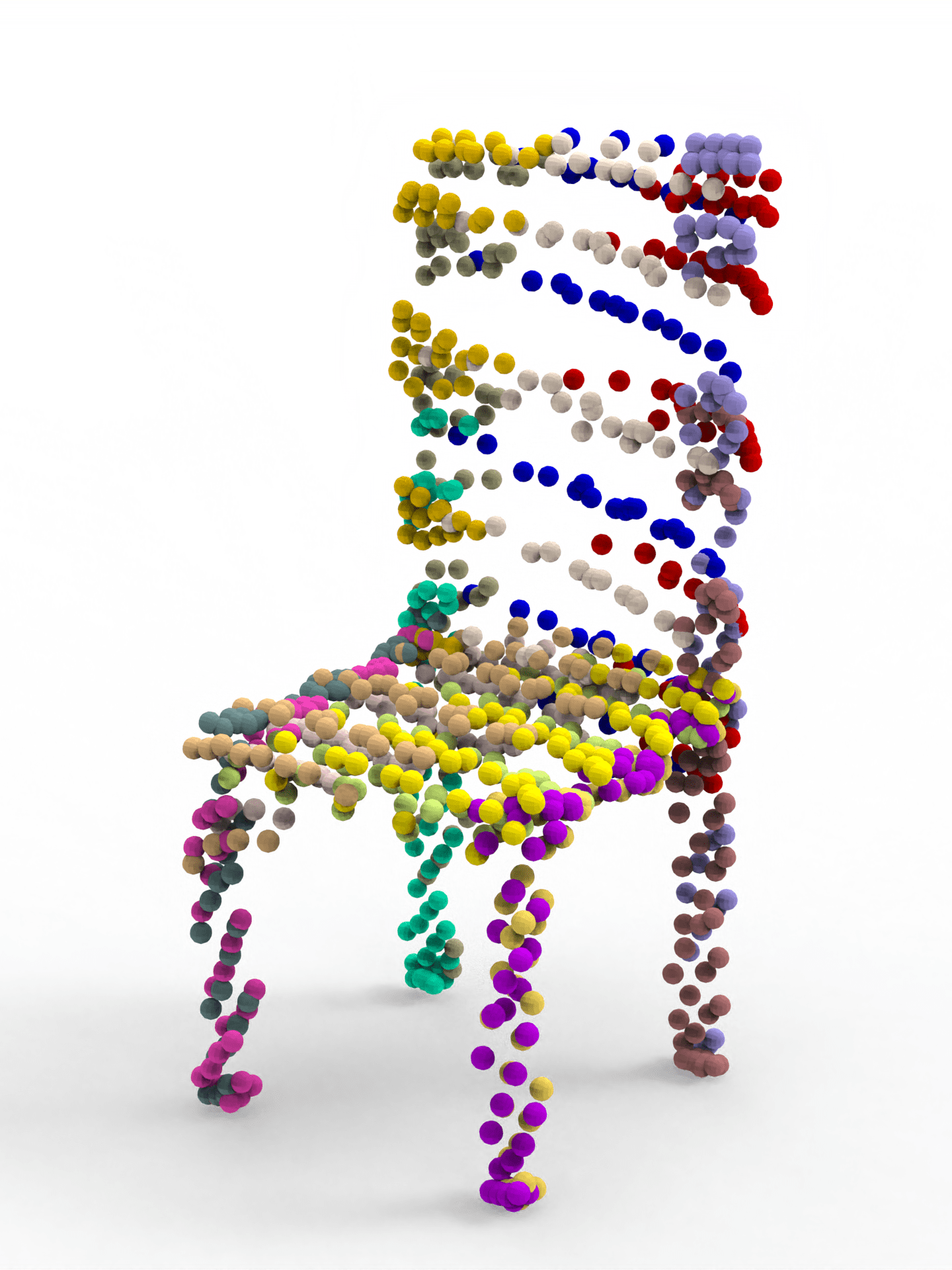}}
\label{flexible:1024} 
\subfloat[2048 points]{
\includegraphics[width=0.8in]{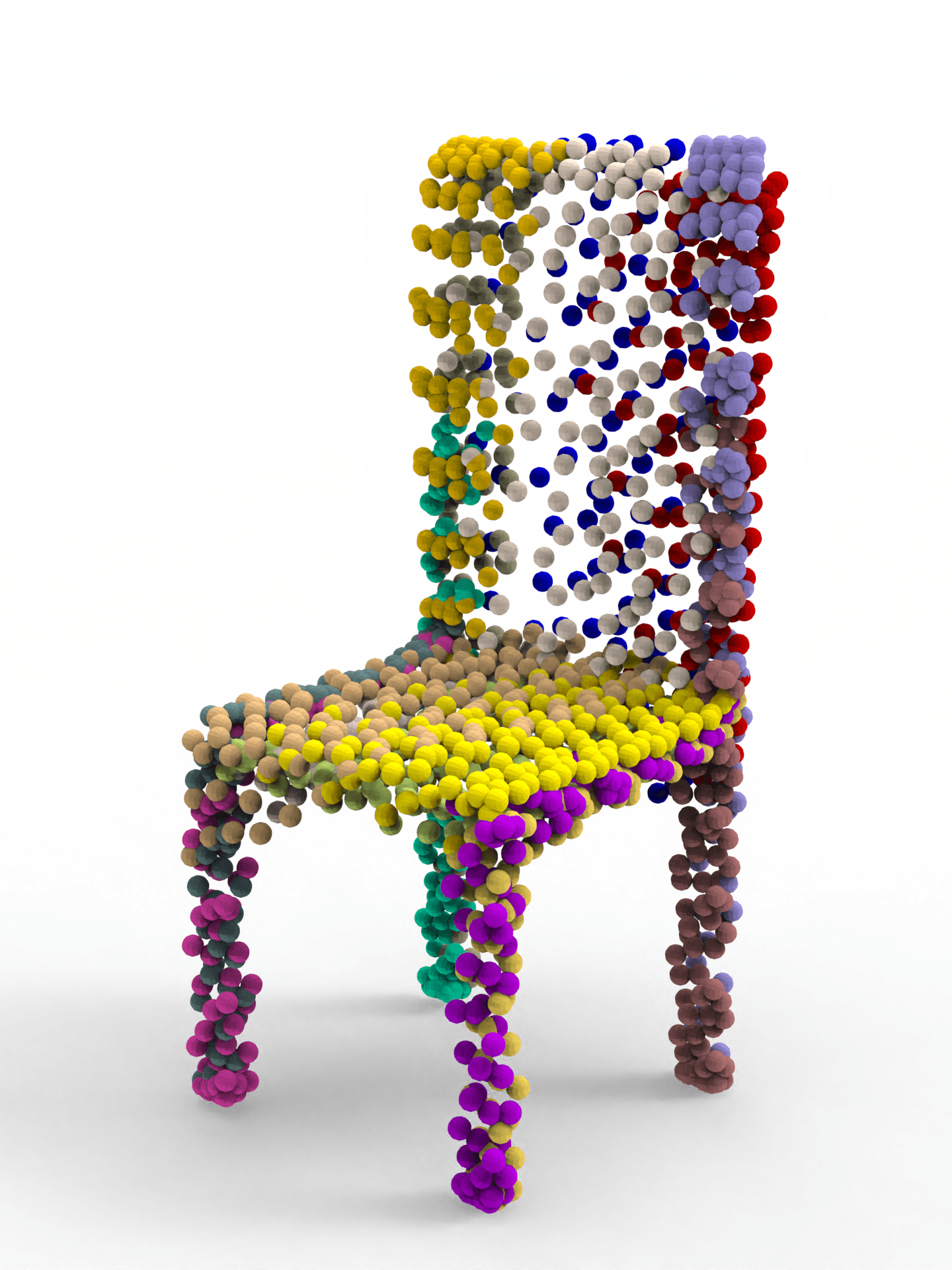}}
\label{flexible:2048} 
\subfloat[4096 points]{
\includegraphics[width=0.8in]{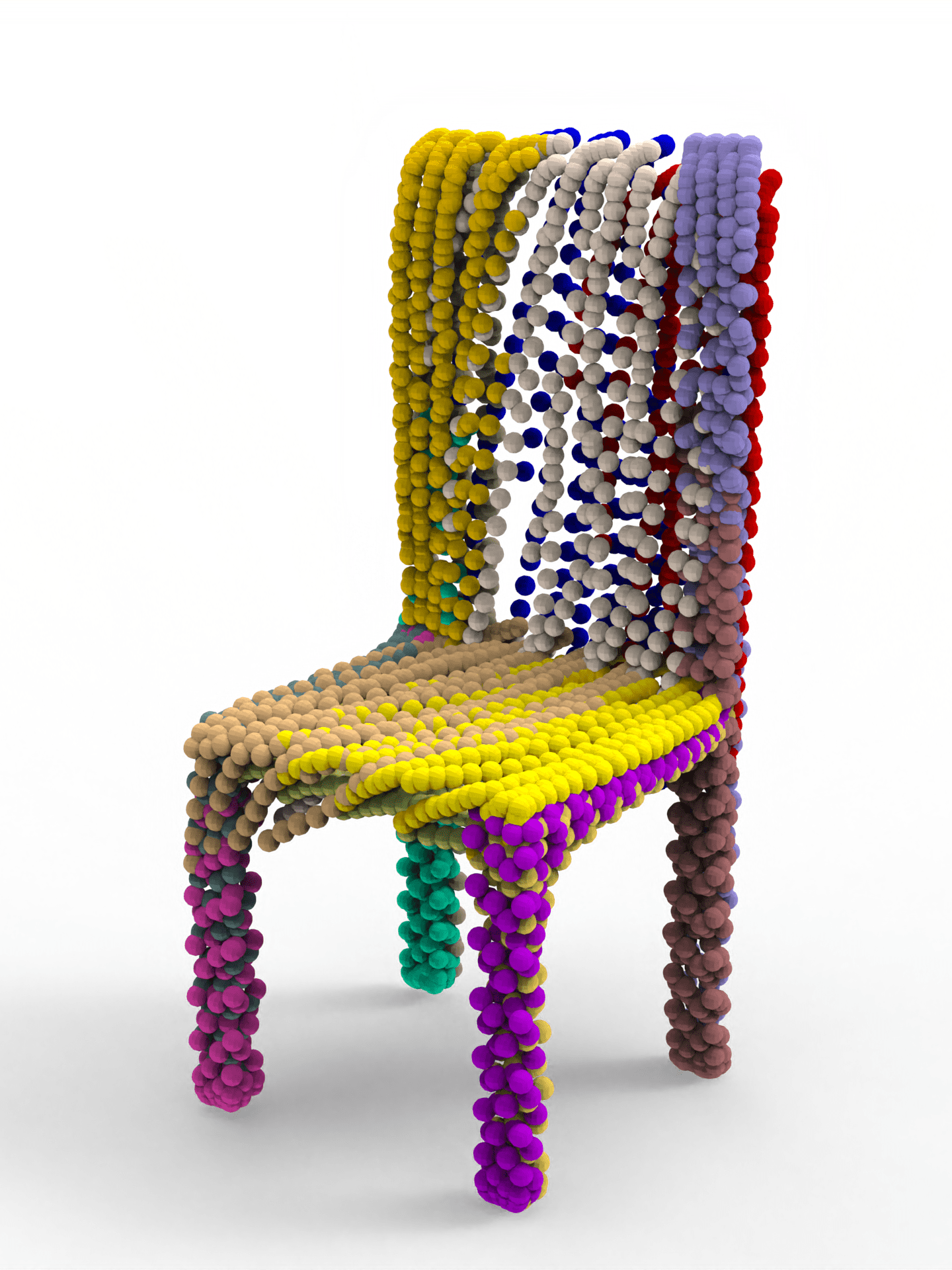}}
\label{flexible:4096} 

\vspace{-0.3cm}
\caption{
Visual illustration of the flexibility of our WarpingGAN in generating point clouds with various number of points after one-time training.
Points with the same color correspond to the same prior. 
}
\label{flexiblepoints} 
\vspace{-0.12cm}
\end{figure}

\textbf{Flexibility illustration.} 
To demonstrate the flexibility of our WarpingGAN, we fixed the WarpingGAN trained with $M=16$ and $N=2,048$ and modified the size of the 3D priors in order to generate point clouds with $N=1,024, 2048$ and $4,096$ from an identical latent code. As shown in Fig. \ref{flexiblepoints}, the three generated point clouds correspond to the same shape,  the warping manners of corresponded priors in different point clouds are the same, and the point cloud quality potentially improves with the number of points increasing. However, the compared methods that directly generate points from the latent code cannot achieve such a flexibility.

\begin{table}[t]
\centering
\caption{Ablation studies conducted on the \textit{Chair} category.
Exps. \#1 - \#8 are the results of our WarpingGAN with various settings. The result of our final model is highlighted in \textcolor{red}{red}.
Exps. \#9 and \#10 are the results of  FoldingNet-based and AtlasNet-based GANs, respectively.
``\textbf{-}" means the stitching loss is not applicable,
``U" and ``NU" mean uniform and non-uniform priors, respectively.}
	\vspace{-0.2em}
\resizebox{\linewidth}{!}{
\begin{tabular}{c|cccc|ccc}
\hline\hline
\multicolumn{1}{c|}{{Exp.} } & \multicolumn{1}{c}{{Code} } & \multicolumn{1}{c}{{Prior}}  & \multicolumn{1}{c}{ Prior} & \multicolumn{1}{c|}{Stitching} &  \multicolumn{1}{c}{\multirow{2}[0]{*}{MMD$\downarrow$}} & \multicolumn{1}{c}{\multirow{2}[0]{*}{COV$\uparrow$}} &\multicolumn{1}{c}{\multirow{2}[0]{*}{Uniform$\downarrow$}} \\\multicolumn{1}{c|}{{Num.} }
& \multicolumn{1}{c}{Enhance.} & \multicolumn{1}{c}{Type}  & \multicolumn{1}{c}{Num.} &  \multicolumn{1}{c|}{Loss}  &       \\
\hline
1     & \xmark     & 3D+U    & 1     & \textbf{-}     & 13.4     & 25.00  &1.25\\
2     & \cmark     & 3D+U    & 1     & \textbf{-}     & 11.0   & 43.75  &1.47\\
3     & \cmark     & 3D+U    & 16    & \xmark    & 9.9  & 45.00  &0.43\\
4     & \cmark     & 2D+U    & 16    & \cmark    & 10.0   &46.25  &0.32\\
5 & \cmark     & 3D+NU    & 16    & \cmark    & 10.2   & 50.00  &0.49\\
\hline
6     & \cmark     & 3D+U    & 4     & \cmark   & 9.6   & 51.25  &0.78\\

\textcolor{red}{7}    & \color{red}{\cmark}     & \textcolor{red}{3D+U}    & \textcolor{red}{16}    & \color{red}{\cmark}   & \textcolor{red}{8.7}   & \textcolor{red}{53.75}  &\textcolor{red}{0.29} \\
8     & \cmark     & 3D+U    & 64    & \cmark    & 8.3   & 46.25  &0.34\\
\hline
9     & \cmark     & 2D+U    & 1     & \textbf{-}     & 14.4   & 27.50 &1.57\\
10     & \cmark     & 2D+U    & 16    & \xmark    & 10.5   & 36.25 &0.90\\
\hline\hline
\end{tabular}%
}
\label{ablationquantitative}%
\end{table}%

\subsection{Ablation Study}\label{sec:ablation}

\begin{figure}[h]
\centering
\setlength{\abovecaptionskip}{0.1cm}
\subfloat[$wo/w$ code enhancement]{
\includegraphics[width=1.45in]{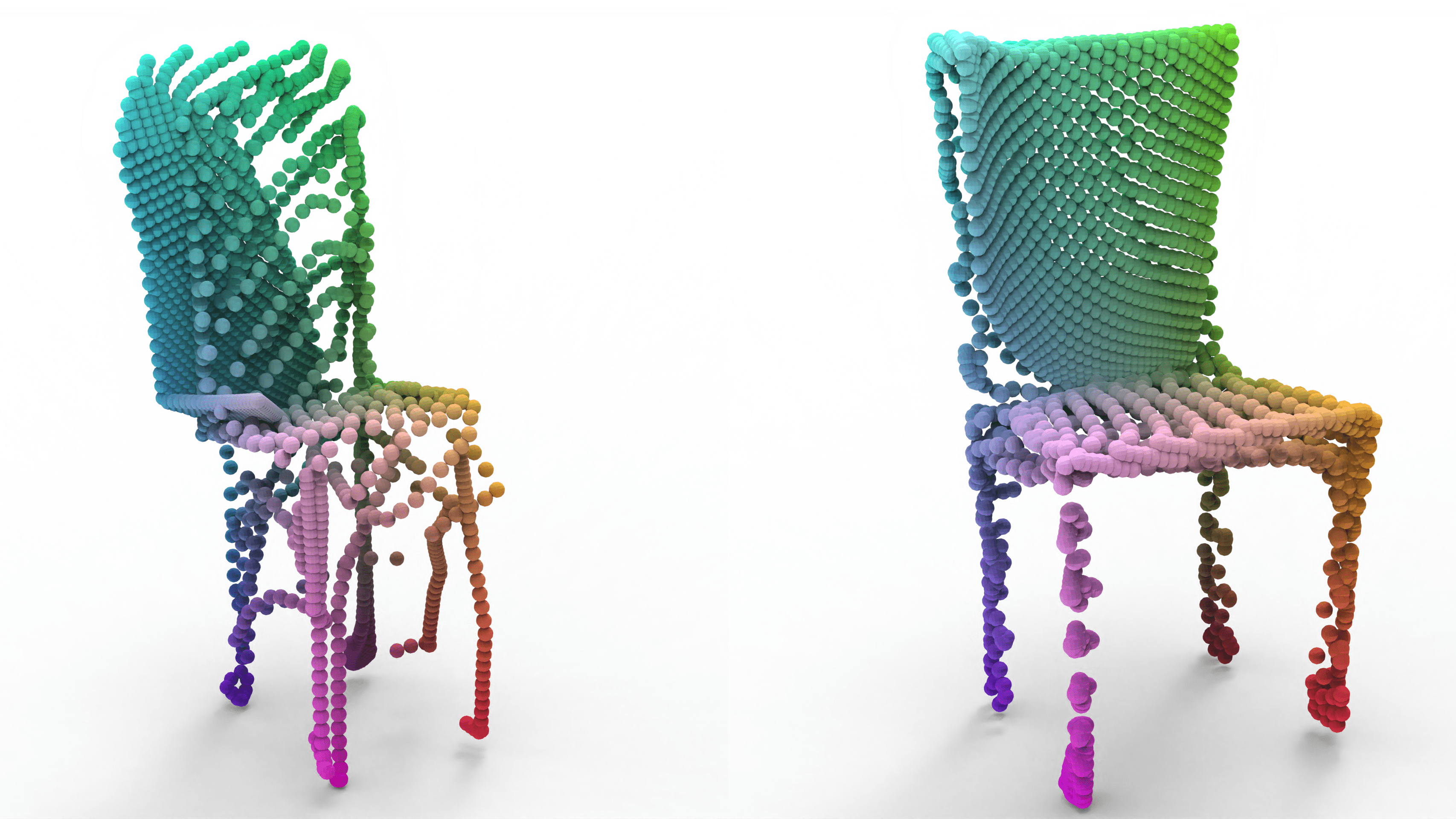}}
\label{a1:wow} 
\subfloat[$wo/w$ global shape code]{
\includegraphics[width=1.45in]{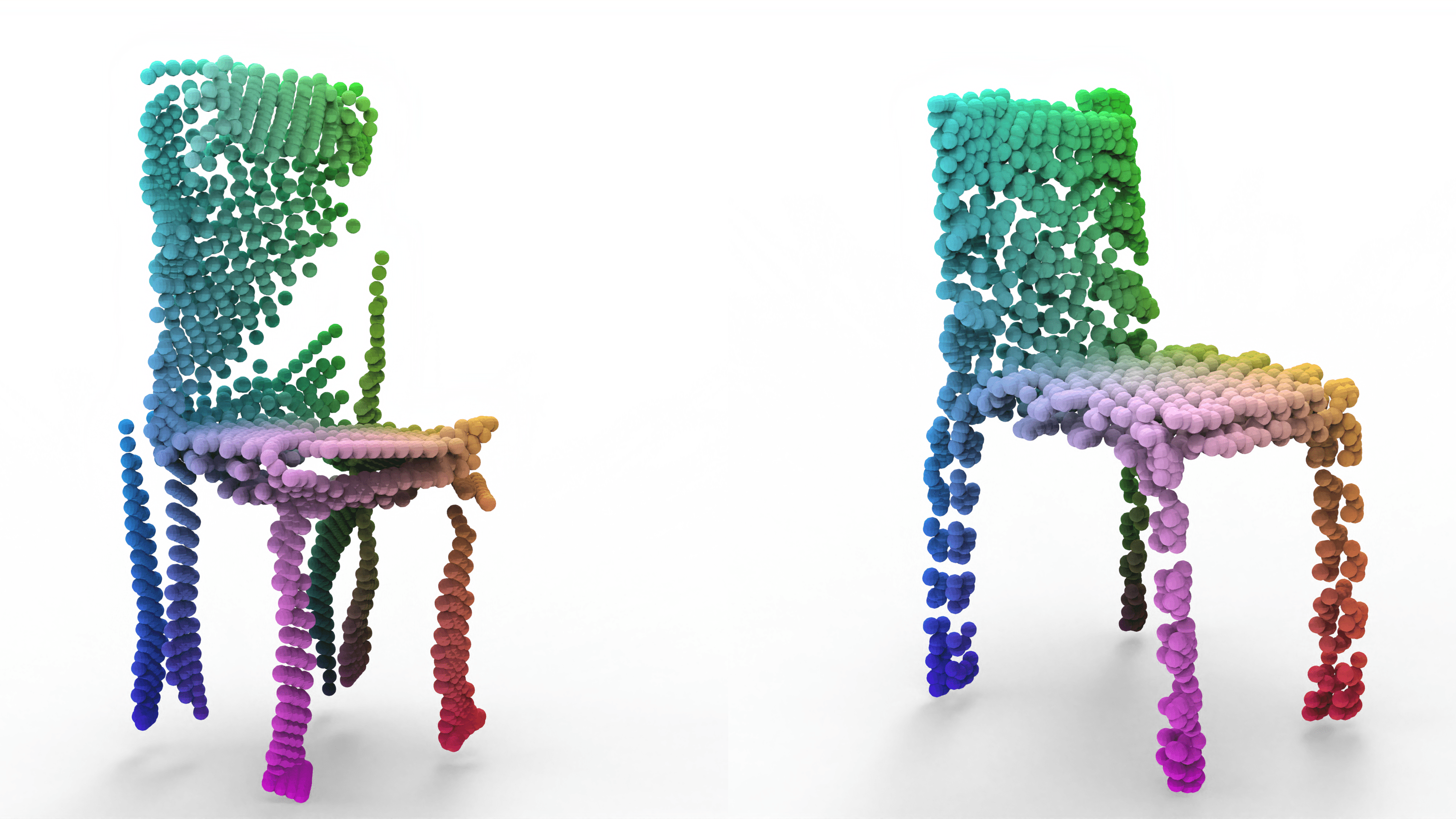}}
\label{a1:wog} 
\caption{Visual illustration of the effectiveness  of (a) the code enhancement module and (b) the global shape code,  where $wo$ and $w$ denote ``without'' and ``with'', respectively.}
\label{ceglobal} 
\vspace{-0.1cm}
\end{figure}
\begin{figure}[h]
\centering
\subfloat[latent code]{
\includegraphics[width=0.85in]{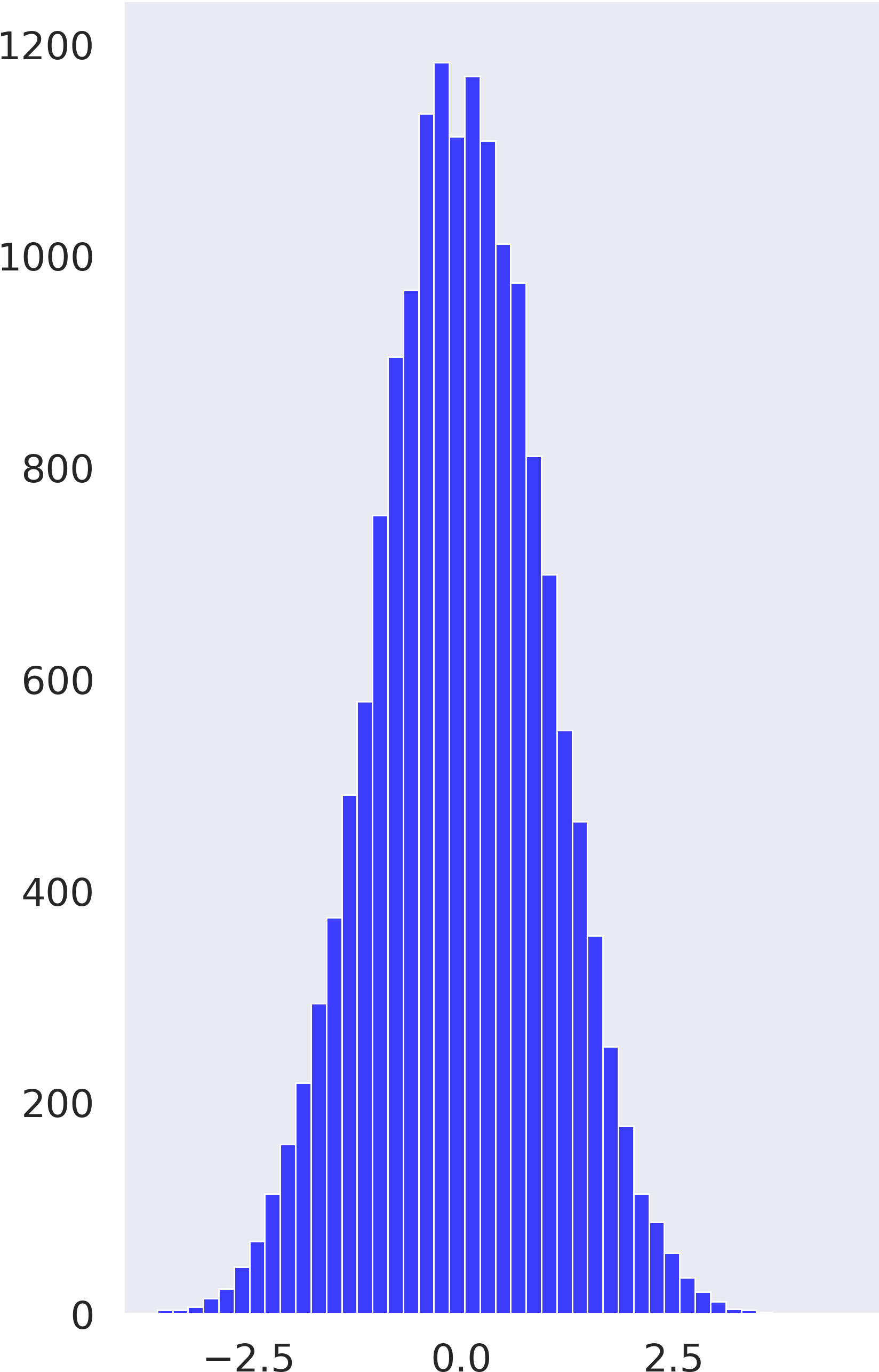}}
\label{dd:l} 
\subfloat[FoldingNet $\theta$]{
\includegraphics[width=0.85in]{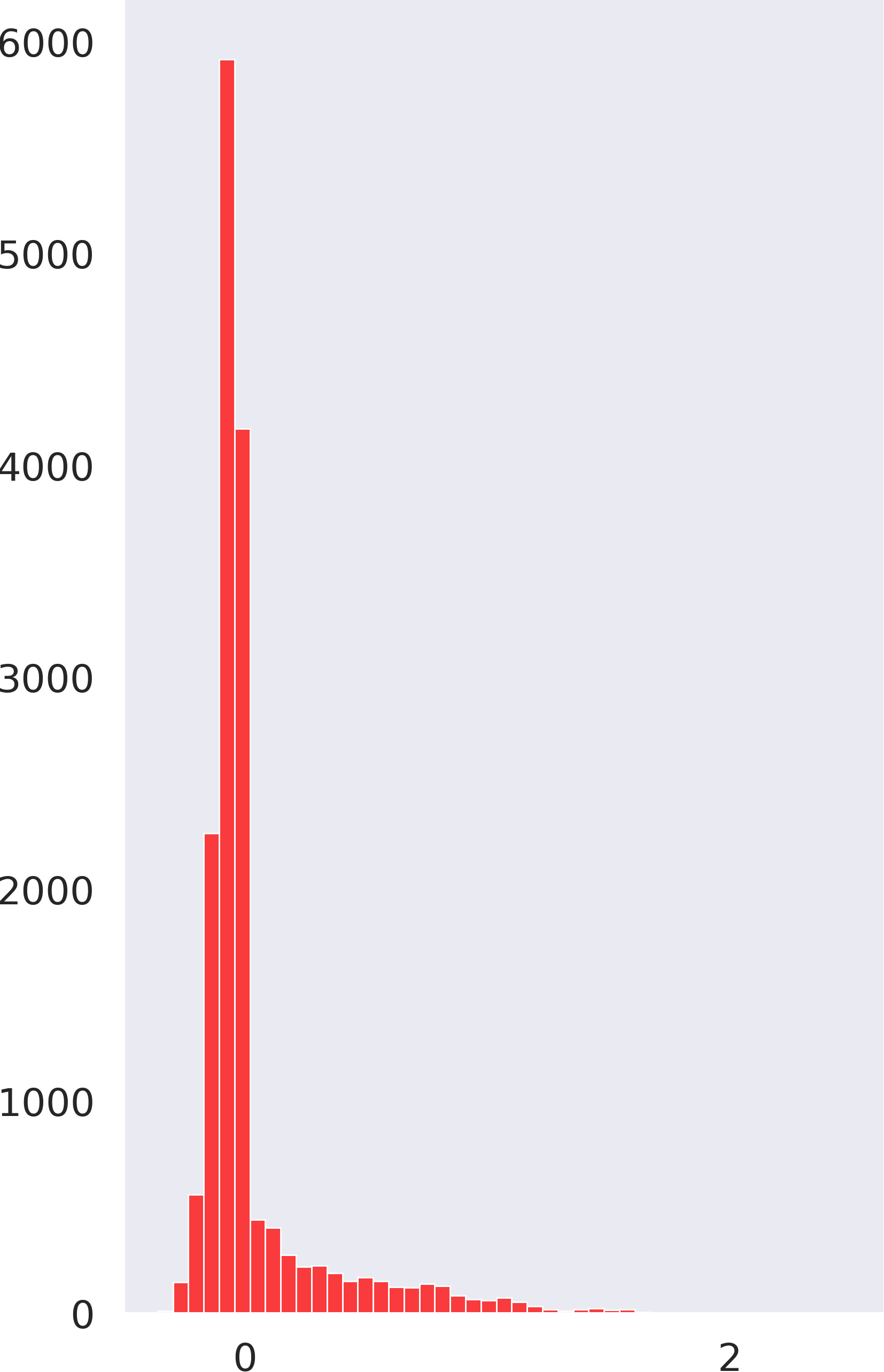}}
\label{dd:f} 
\subfloat[enhanced code]{
\includegraphics[width=0.85in]{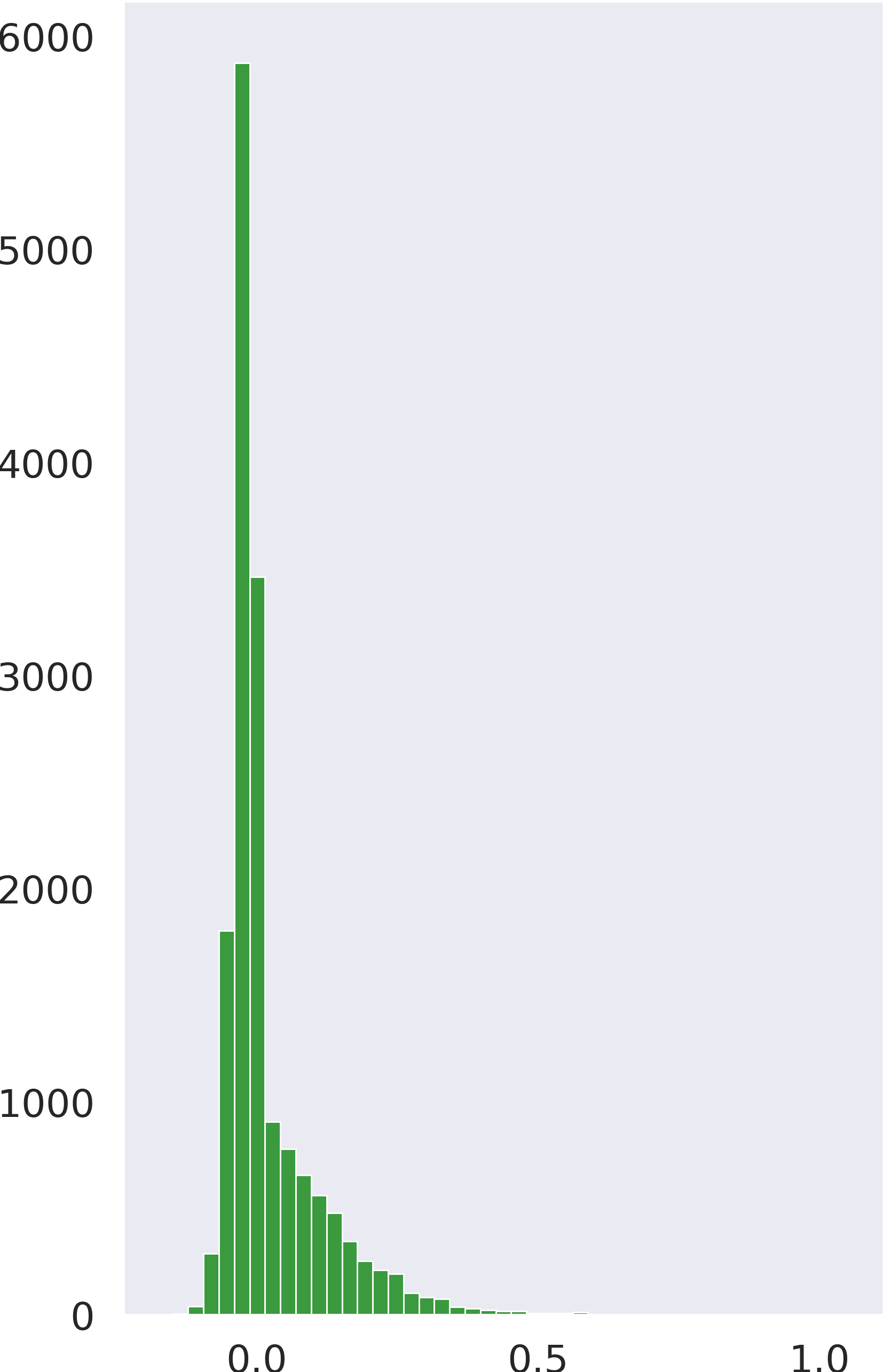}}
\label{dd:w} 
\vspace{-0.3cm}
\caption{Visual comparison of the  distribution of (a) the input latent code, (b) the latent semantic features of FoldingNet, and (c) the enhanced code of WarpingGAN ($M=1$).}
\label{cedistribution} 
\vspace{-0.3cm}
\end{figure}
\textbf{The effectiveness of the code enhancement} is demonstrated via the quantitatively and qualitatively in Table \ref{ablationquantitative} and Fig. \ref{ceglobal}, respectively. As listed in Exps. \#1 and \#2 of Table 3, it can be seen that this module can effectively improve the generated point cloud quality in terms of all metrics.  Fig. \ref{ceglobal} (a) visually demonstrates that the quality of the generated point cloud degrades dramatically without using this module. Besides, We also validated the advantage of integrating the global code $\widetilde{\mathbf{z}}$ into the local code $\mathbf{z}^j$ in Fig. \ref{ceglobal} (b). Moreover, we also investigated  the distribution of the enhanced latent code to understand this module better. As shown in Fig. \ref{cedistribution}, it can be seen that this module is able to transform the initial Gaussian distribution (Fig. \ref{cedistribution} (a))  to a distribution (Fig. \ref{cedistribution} (c)) which is very close to the distribution of semantic features extracted from real datasets by using FoldingNet (Fig. \ref{cedistribution} (b)). 

\begin{figure}[h]
\centering
\setlength{\abovecaptionskip}{0.1cm}
\setlength{\belowcaptionskip}{-0.4cm}
\includegraphics[width=3.0in]{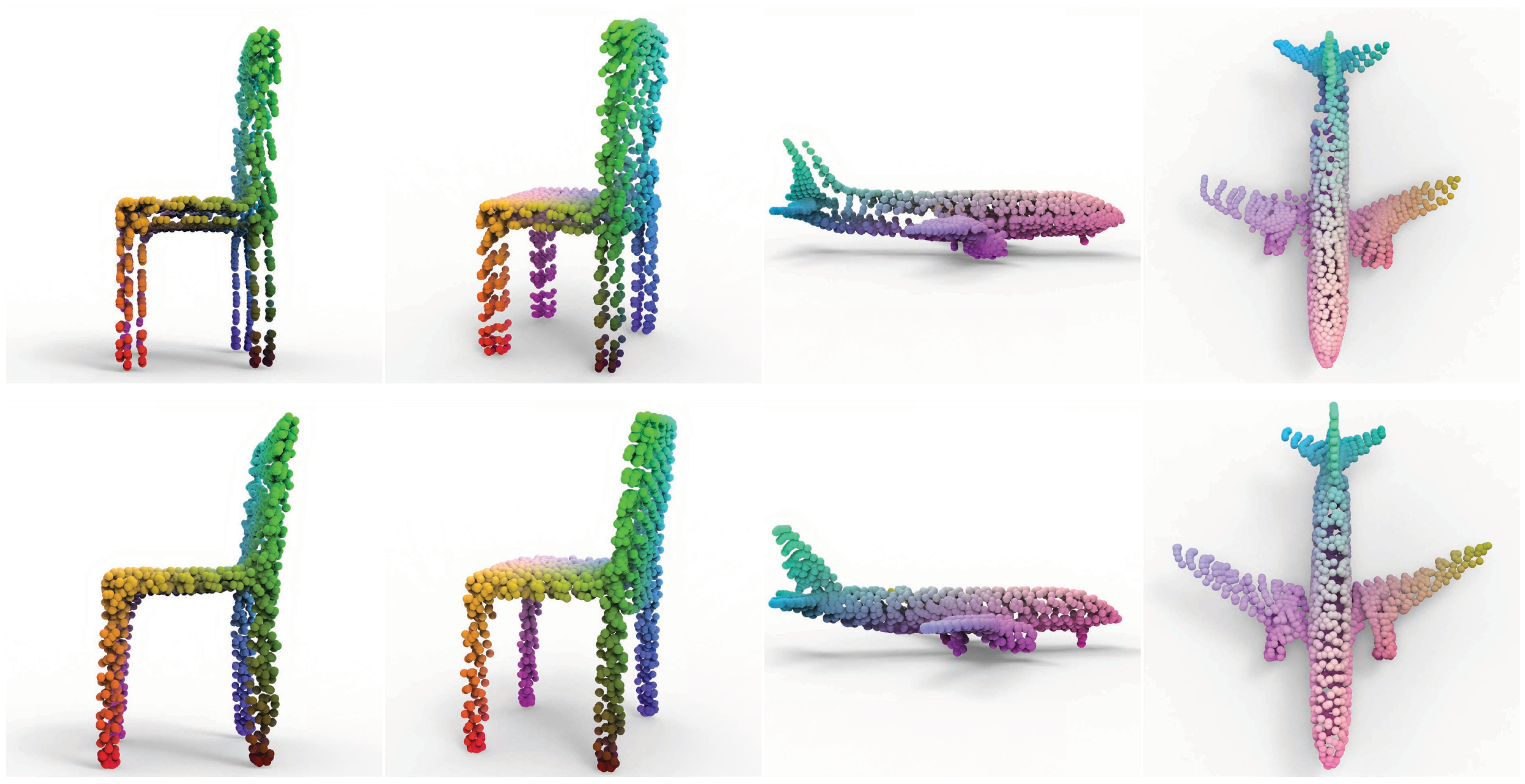}
\caption{Visual comparison of our WarpingGAN trained  (top row) without  and  (bottom row) with the stitching loss. 
}
\label{fig:Stitch}
\end{figure}
\textbf{The effectiveness of the stitching loss } is quantitatively validated by comparing the results of Exps. \#3 and \#7 listed in Table \ref{ablationquantitative}, where it can be seen that the results of all metrics improve when adopting the loss. Besides, as shown in Fig. \ref{fig:Stitch}, the point clouds generated by WarpingGAN trained without the stitching loss suffer from significant gaps between different partitions, while WarpingGAN trained with the stitching loss can greatly alleviate the gaps and increase the visual quality.  

\begin{figure}[h]
\centering
\setlength{\abovecaptionskip}{0.1cm}
\subfloat[2D vs. 3D priors]{
\includegraphics[width=1.45in]{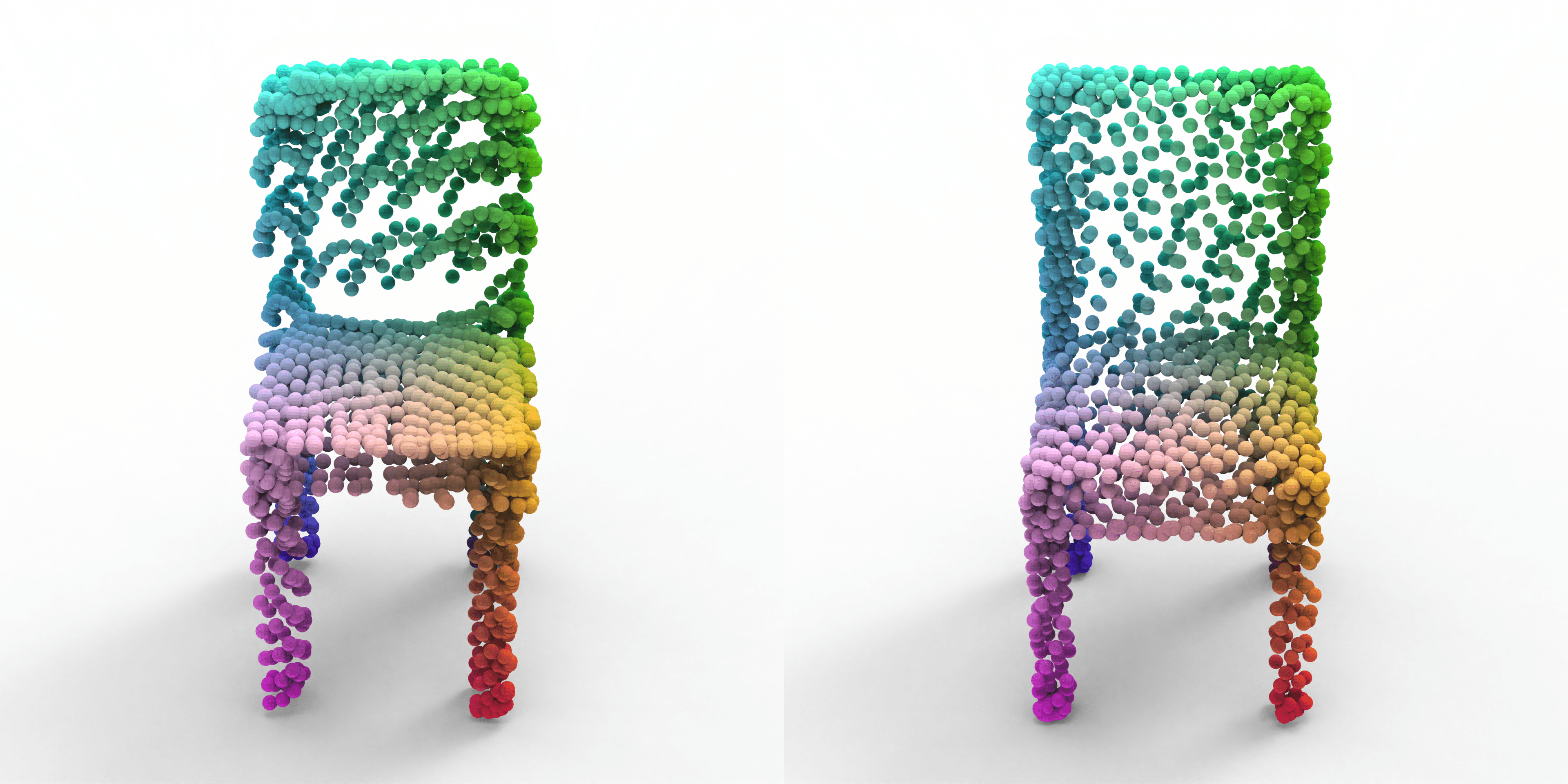}}
\label{prior:2d} 
\subfloat[non-uni vs. uni priors]{
\includegraphics[width=1.45in]{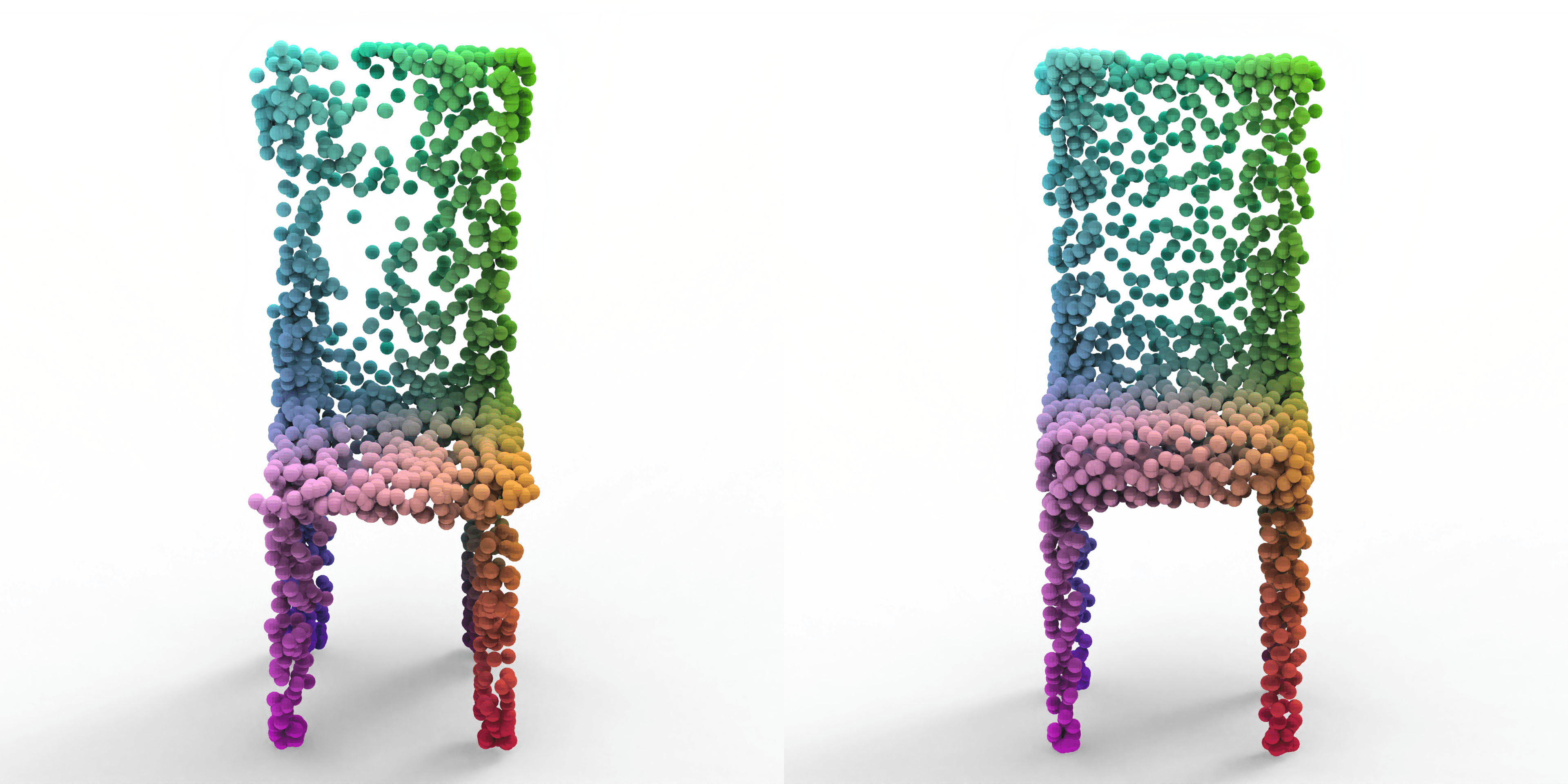}}
\label{prior:3d} 
\vspace{-0.1cm}
\caption{Visual comparison of  our WarpingGAN equipped with (a) 2D priors and 3D priors, and (b) 3D non-uniform priors and 3D uniform priors. 
}
\label{2dvs3d} 
\vspace{-0.2cm}
\end{figure}

\textbf{The effect of different prior settings.} First, we substituted the 3D uniform priors with 2D uniform priors, while keeping the remaining settings unchanged. By comparing the results of Exps. \#4 and \#7 in Table \ref{ablationquantitative}, we can conclude the advantage of 3D priors over 2D priors. From Fig. \ref{2dvs3d}(a), it can be seen that the generated point cloud by 3D uniform priors retain local details better. Besides, to demonstrate the necessity of the uniformity of the prior, we replaced the 3D uniform priors with 3D non-uniform priors and kept the remaining settings unchanged. As shown in Fig. \ref{2dvs3d} (b), the non-uniform priors cannot lead to uniformly distributed point clouds, which is consistent with the quantitative results in Exps. \#5 and \#7 of Table \ref{ablationquantitative}.

\begin{figure}[h]
\centering
\setlength{\abovecaptionskip}{0.1cm}
\subfloat[4 priors]{
\includegraphics[width=0.8in]{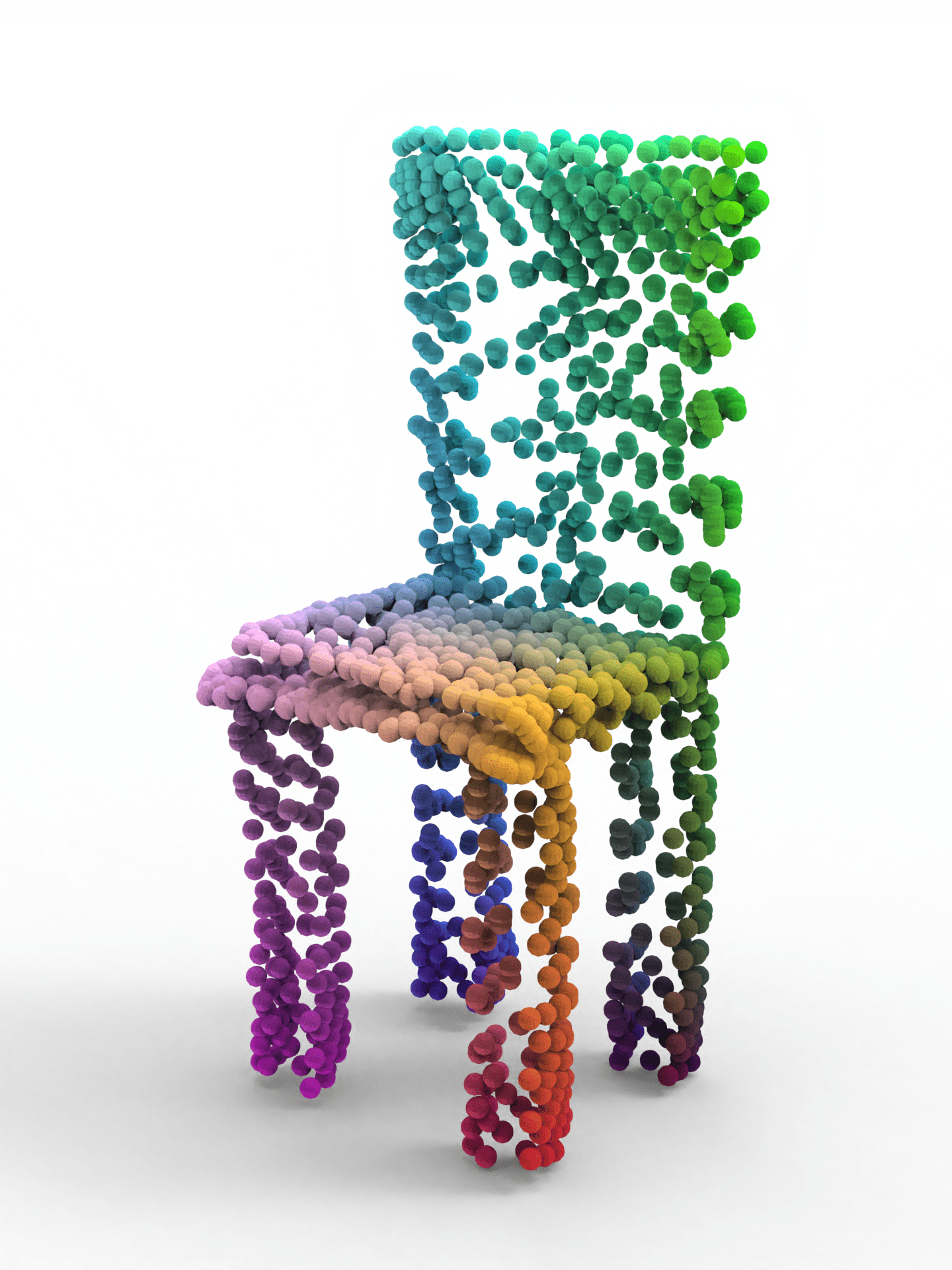}}
\label{diffpatch:4} 
\subfloat[16 priors]{
\includegraphics[width=0.8in]{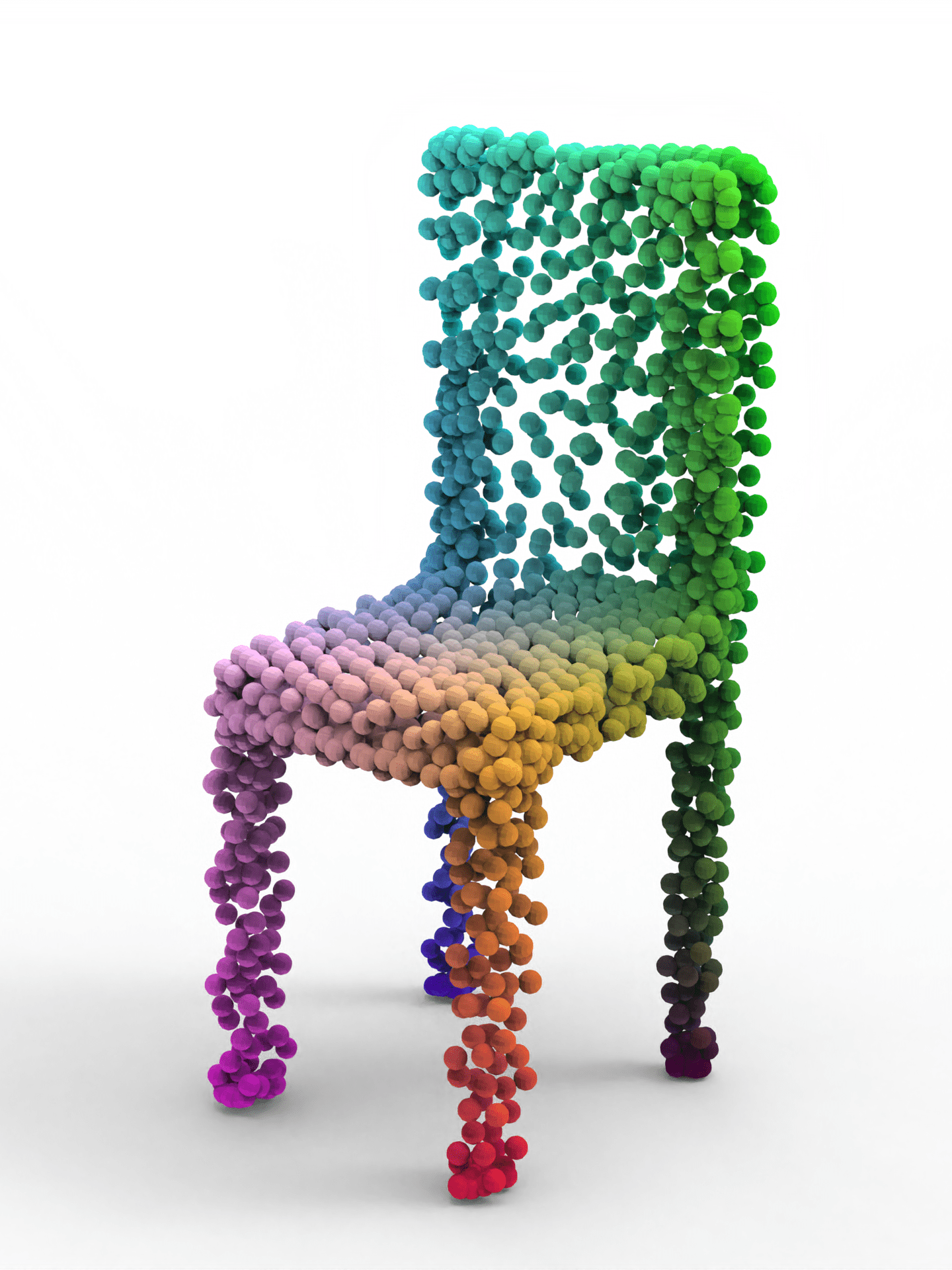}}
\label{diffpatch:16} 
\subfloat[64 priors]{
\includegraphics[width=0.8in]{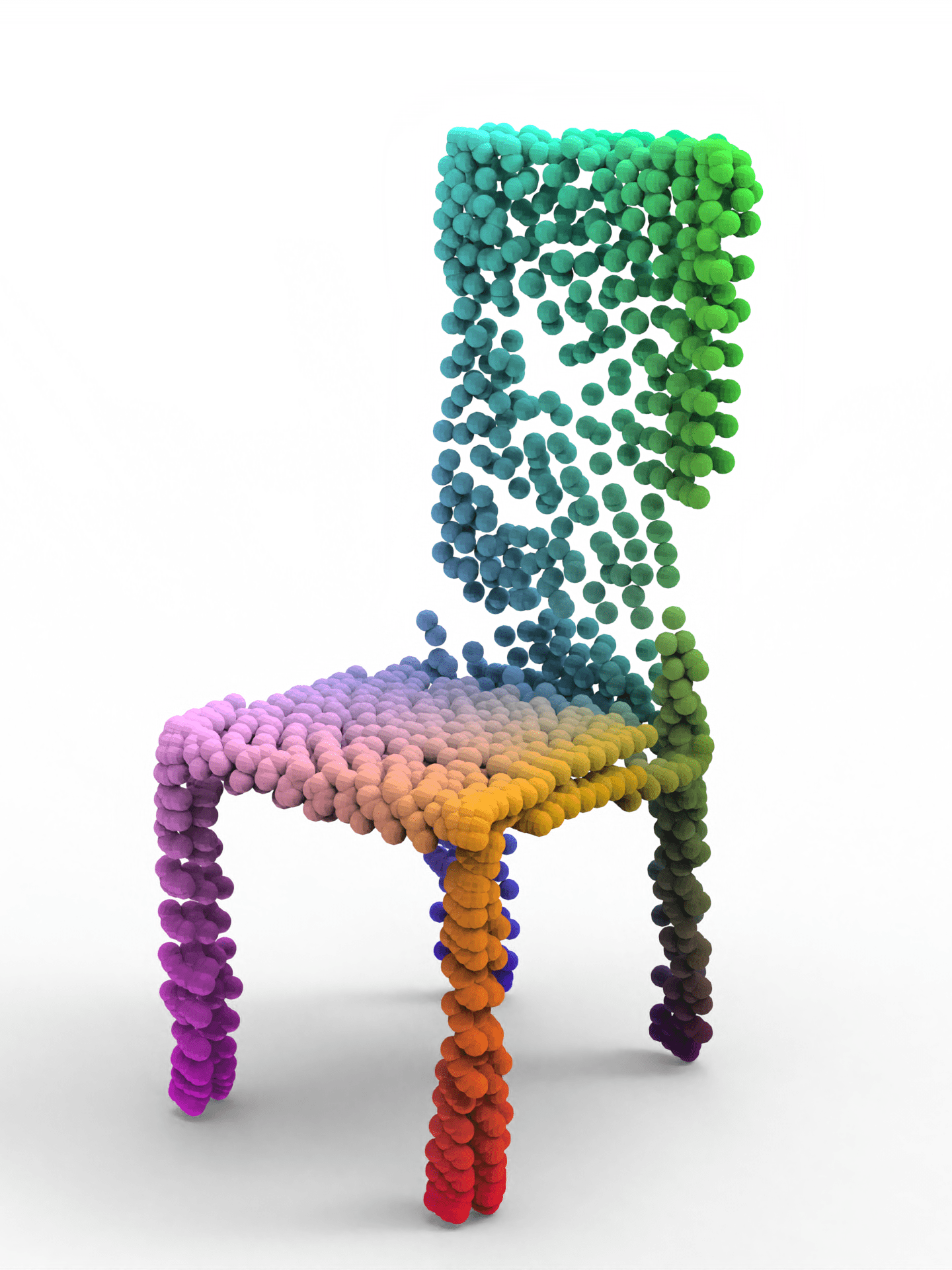}}
\label{diffpatch:64} 
\vspace{-0.1cm}
\caption{Visual comparison of our WarpingGAN equipped with different numbers of 3D priors. Note that the total size of priors under different settings are equal for generating point clouds each with 2,048 points.  
}
\label{differentpatches} 
\vspace{-0.4cm}
\end{figure}

 \textbf{The effect of the number of 3D priors.} 
We trained three WarpingGAN models with 4, 16 and 64 priors, respectively. The quantitative comparisons are listed in Exps. \#6, \#7, and \#8 of Table \ref{ablationquantitative}, where it can be seen that WarpingGAN with 16 priors produces the generally best performance, which is consistent with the visual comparison shown in Fig. \ref{differentpatches}. The reason is that a limited number of priors cannot fit 3D shapes well, while too many priors make it hard to optimize the stitching loss. In this paper, we set $M=16$. 

\begin{figure}[h]
\centering
\setlength{\abovecaptionskip}{0.1cm}
\subfloat[\scriptsize Folding-based vs. WarpingGAN]{
\includegraphics[width=1.45in]{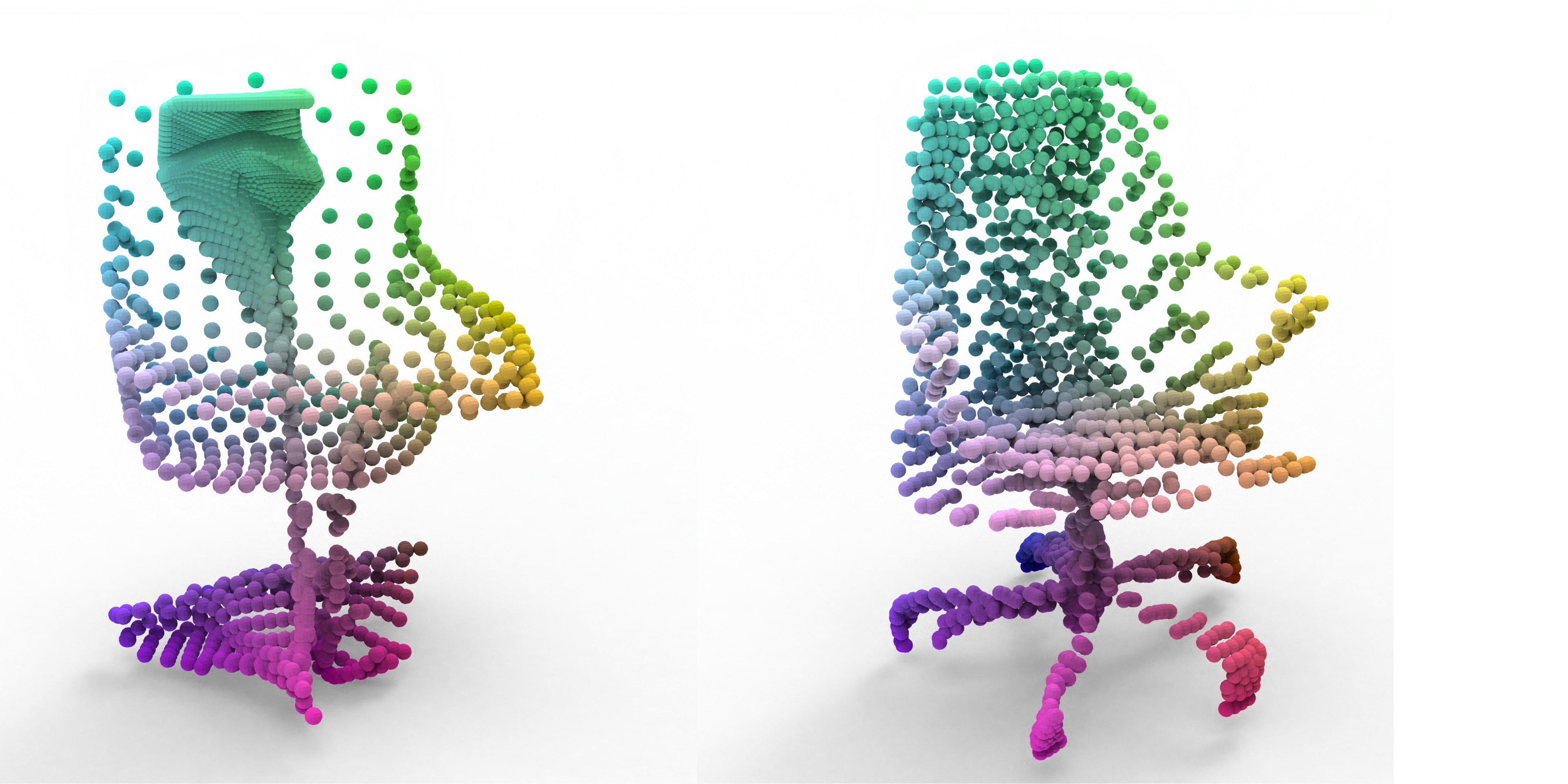}}
\label{cfa:f} 
\subfloat[\scriptsize Atlas-based vs. WarpingGAN]{
\includegraphics[width=1.45in]{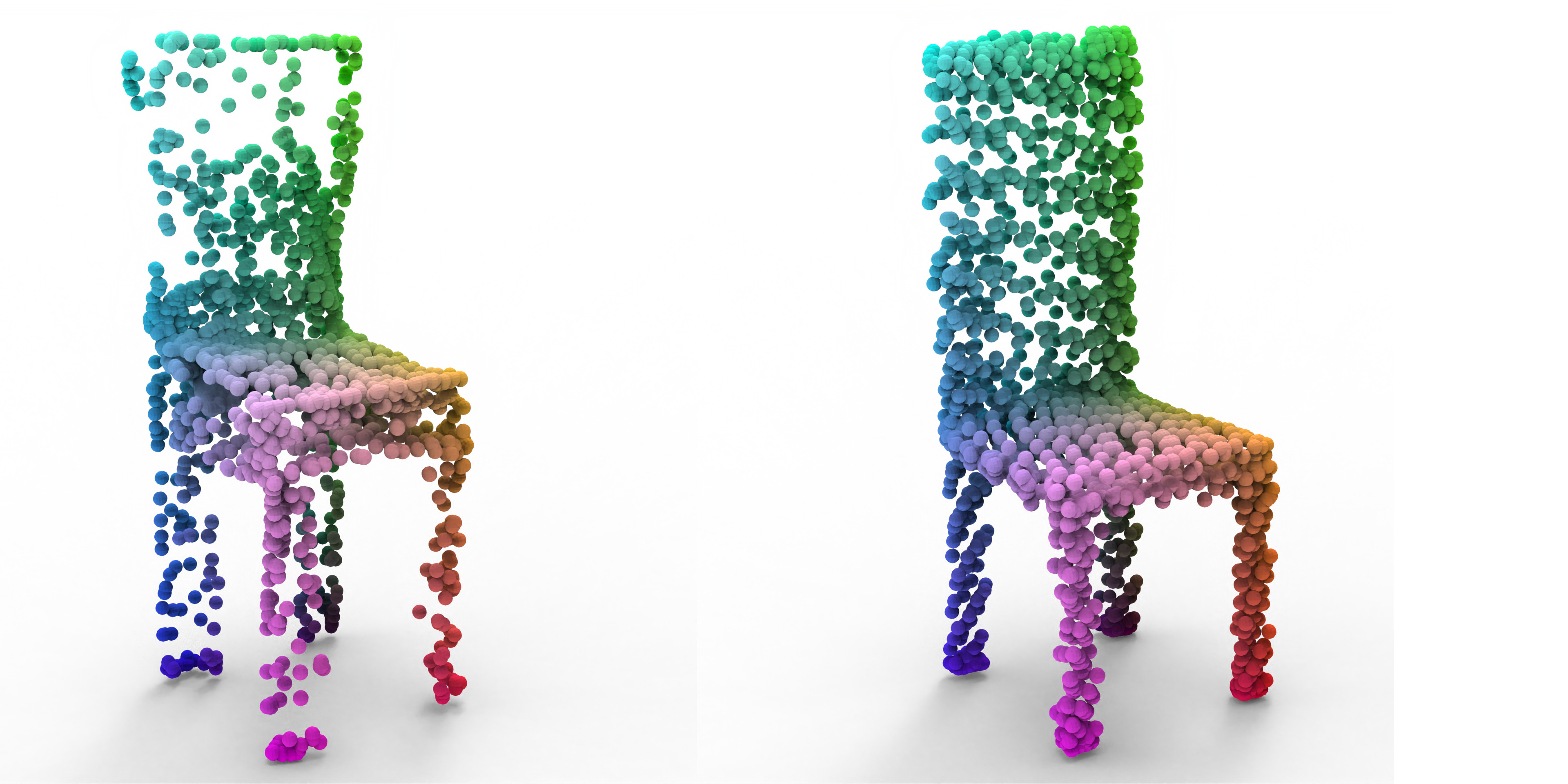}}
\label{cfa:a} 
\vspace{-0.1cm}
\caption{Visual comparison of WarpingGAN with (a) FoldingNet-based GAN and  (b) AtlasNet-based GAN.}
\label{comparefold} 
\vspace{-0.1cm}
\end{figure}

\textbf{Comparison with FoldingNet-based and AtlasNet-based GANs.} In this experiment, we retained the code enhancement module and substituted the prior warping module of the proposed generator with the decoders of FoldingNet and AtlasNet. Exps. \#9 and \#10 of Table \ref{ablationquantitative} provide the quantitative performance of these two baselines, which rev eal the limited performance of the direct extensions of these auto-encoder frameworks. Besides, as shown in Fig. \ref{comparefold}, FoldingNet-based GAN fails to generate a chair with a complex structure correctly and AtlasNet-based GAN loses much local details even when generating a chair with a simple structure, while our WarpingGAN can generate chairs with much better quality.

\section{Conclusion}
We presented WarpingGAN, a novel point cloud generation framework that is capable of generating high-quality point clouds in an effective and efficient manner. In contrast to existing approaches that usually produce point clouds by learning the direct mapping between the random latent codes and 3D shapes, we designed WarpingGAN by investigating a unified local-warping mechanism, in which multiple pre-defined 3D priors uniformly distributed in the 3D Euclidean space are \textit{conditionally} warped to various local regions of a shape. Meanwhile, by examining the principle of the discriminator, we customized a stitching loss to eliminate the gaps between different regions. Such a new mechanism makes WarpingGAN compact, efficient, and flexible. We conducted extensive experiments to demonstrate the significant advantages of  WarpingGAN over state-of-the-art methods in terms of quantitative metrics, visual quality, and efficiency.


{\small
\bibliographystyle{ieee_fullname}
\bibliography{egbib}
}

\end{document}


	\title{ WarpingGAN: Warping Multiple Uniform Priors for Adversarial 3D Point Cloud Generation \\ (Supplementary Materials)}
	
		\author{
		Yinzhi Tang$^1$\footnotemark[1] \quad Yue Qian$^1$\footnotemark[1] \quad Qijian Zhang$^1$\quad Yiming Zeng$^1$\quad Junhui Hou$^1$\quad Xuefei Zhe$^2$\\
$^1$
City University of Hong Kong~~ 
$^2$
Tencent AI lab~~ \\
{\tt\small \{yztang4-c, yueqian4-c, qijizhang3-c, ym.zeng\}@my.cityu.edu.hk, jh.hou@cityu.edu.hk}
}
 \maketitle

In this supplementary material, we provided the detailed network architecture of our WarpingGAN (Section \ref{sec:network archi}), the subjective evaluation of different methods (Section  \ref{sec:subjective}), and more visual results including results of ablation Studies on \textit{Airplane} (Section \ref{subsec:more ablation}), results of more shapes generated by our method (Section \ref{subsec:more shapes}), the video demo (Section  \ref{subsec:demo} and the \textcolor{red}{\href{https://youtu.be/GY1EeGI6jZ0}{\it demo https://youtu.be/GY1EeGI6jZ0}} video), and results of failure cases (Section \ref{subsec:failure}). 
 
 \section{Details of the Network Architecture}
 \label{sec:network archi}
 
Table~\ref{table:network} shows the  detailed  network  architecture of the proposed  WarpingGAN, including the code enhancement module and unified local-warping module in the generator and also the discriminator. In Table~\ref{table:network},  we presented the input and output dimensions of each layer, where \textit{Shared MLP} denotes one unified set of MLP parameters is applied to $N$ points in parallel. Note that we utilized LeakyReLU with the slope equal to 0.2 as the activation function.

\begin{table}[H]
  \centering
  \label{CorrNet3D_detail}
  \footnotesize
  \begin{tabular}{c|c}
    \toprule[1.5pt]
  Module & Architecture    \\ \hline\hline
  
           & MLP(128, 128)+LeakyReLU \\
           Code & MLP(128, 128)+LeakyReLU  \\
           Enhancement & MLP(128, 256)+LeakyReLU  \\
           & MLP(256, 256)+LeakyReLU  \\
           & MLP(256, 512)+LeakyReLU  \\\hline
         & Concat(512+512/16+3)$\rightarrow$547 \\
         $1^{st}$ Unified & Shared MLP(547,256)+LeakyReLU  \\
         Local-warping & Shared MLP(256,64)+LeakyReLU  \\
         & Shared MLP(64,3) \\\hline
                    & Concat(512+512/16+3)$\rightarrow$547 \\
         $2^{nd}$ Unified & Sharec MLP(547,256)+LeakyReLU  \\
         Local-warping & Shared MLP(256,64)+LeakyReLU  \\
         & Shared MLP(64,3) \\\hline
         &Shared MLP(3,64)+LeakyReLU \\
         &Shared MLP(64,128)+LeakyReLU \\
         &Shared MLP(128,256)+LeakyReLU \\
         Discriminator &Shared MLP(256,512)+LeakyReLU \\
         &Shared MLP(512,512)+LeakyReLU \\
         &MaxPooling$\rightarrow$ 512 \\
         &MLP(512,1) 
  \\\bottomrule[1.5pt]
  \end{tabular}
  \caption{Network architecture of WarpingGAN.}
  \label{table:network}
  \end{table}

\section{Subjective Evaluation}
\label{sec:subjective}
As argued in our manuscript, the MMD and COV-based quantitative evaluations may not  faithfully reflect the quality of generated data. Thus, we conducted subjective evaluation to compare different methods quantitatively. Specifically, We invited 50 volunteers covering undergraduate students,  postgraduate students with various research background, and researchers and engineers from industry to do the evaluation on \textit{Chair}, \textit{Airplane} and \textit{Car} three categories of ShapeNet. For each method, we displayed the video rendered from 20 randomly generated point clouds and asked the volunteers to rate the method with the score in the range of 1 and 5, based on the quality of the generated shapes, i.e., 1: bad, 2: poor, 3: fair, 4: good, 5:excellent. Besides, the point clouds randomly selected from the real dataset were displayed for reference. We refer the readers to the submitted video demo named \textit{demo.mp4} to examine the shapes in this subjective evaluation. Note that the names of methods are blind to volunteers, and we randomly displayed the videos of different methods for the three categories.

\begin{figure}[h]
\centering
\subfloat[TreeGAN]{
\includegraphics[width=2.2in]{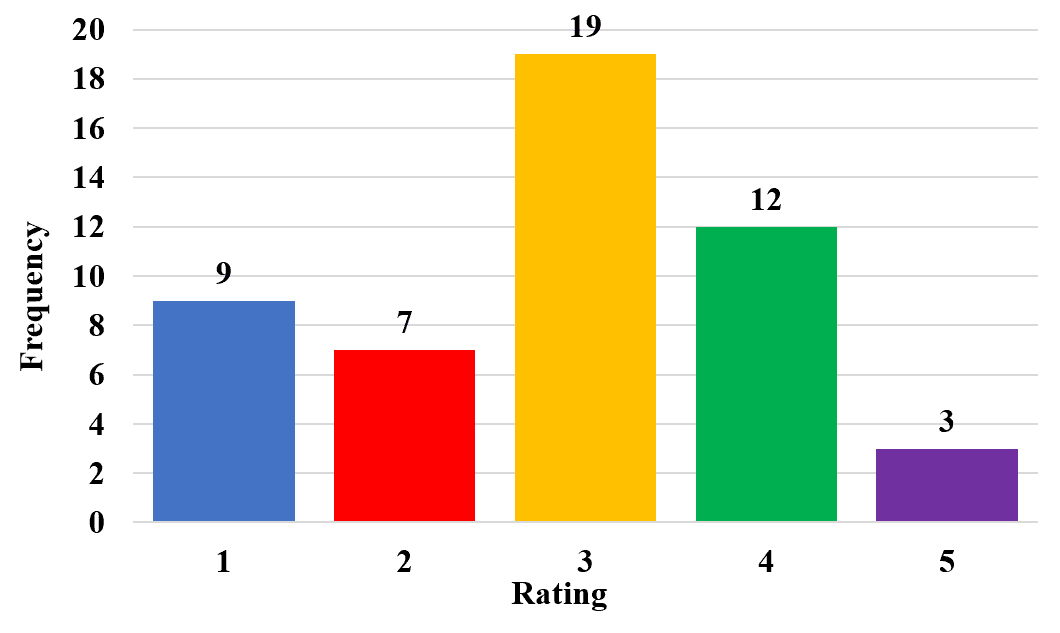}}
\label{subjective:treegan} 
\subfloat[PDGN]{
\includegraphics[width=2.2in]{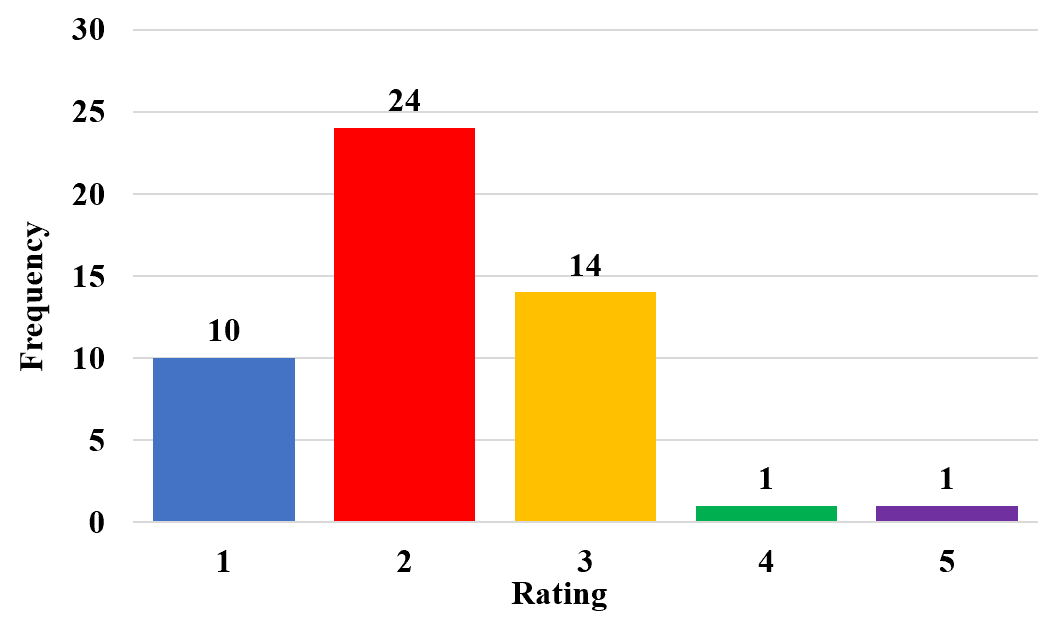}}
\label{subjective:pdgn} 
\subfloat[SP-GAN]{
\includegraphics[width=2.2in]{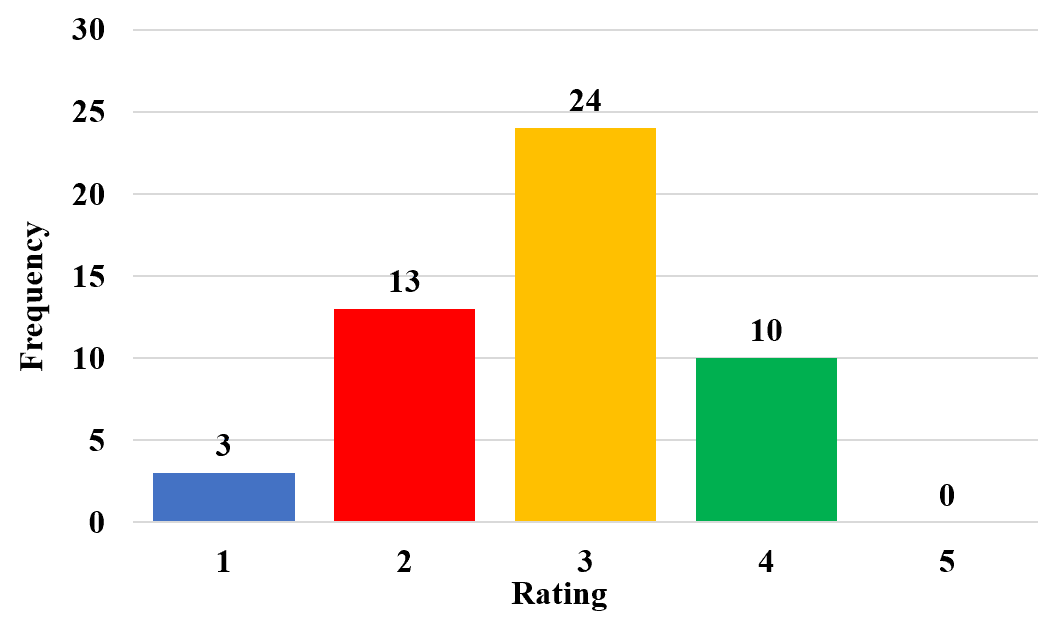}}
\label{subjective:spgan} 
\\
\subfloat[ShapeGF]{
\includegraphics[width=2.2in]{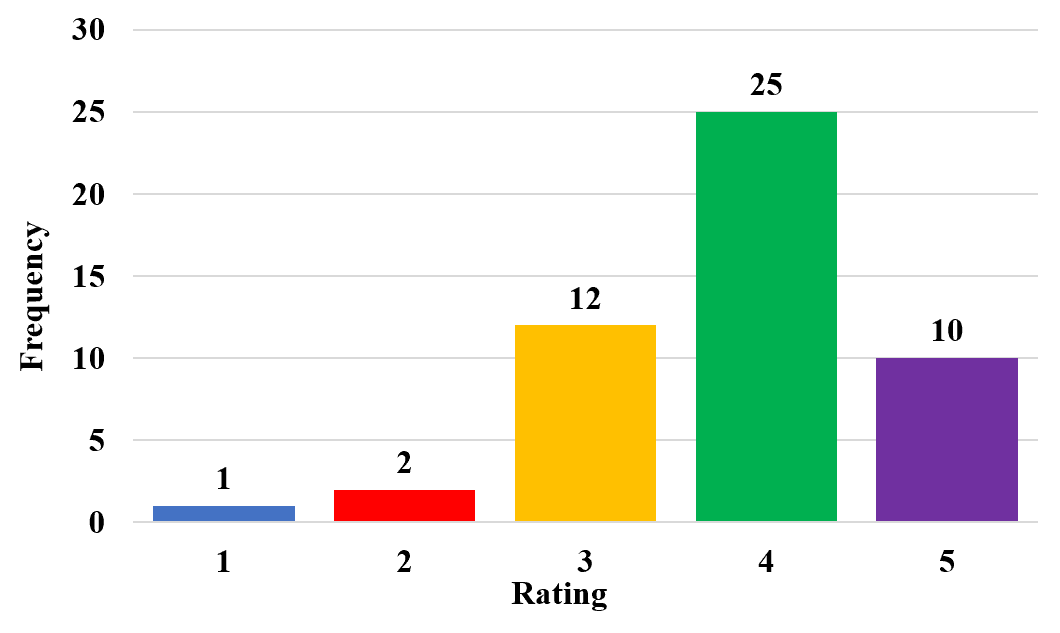}}
\label{subjective:shapegf} 
\subfloat[DPM]{
\includegraphics[width=2.2in]{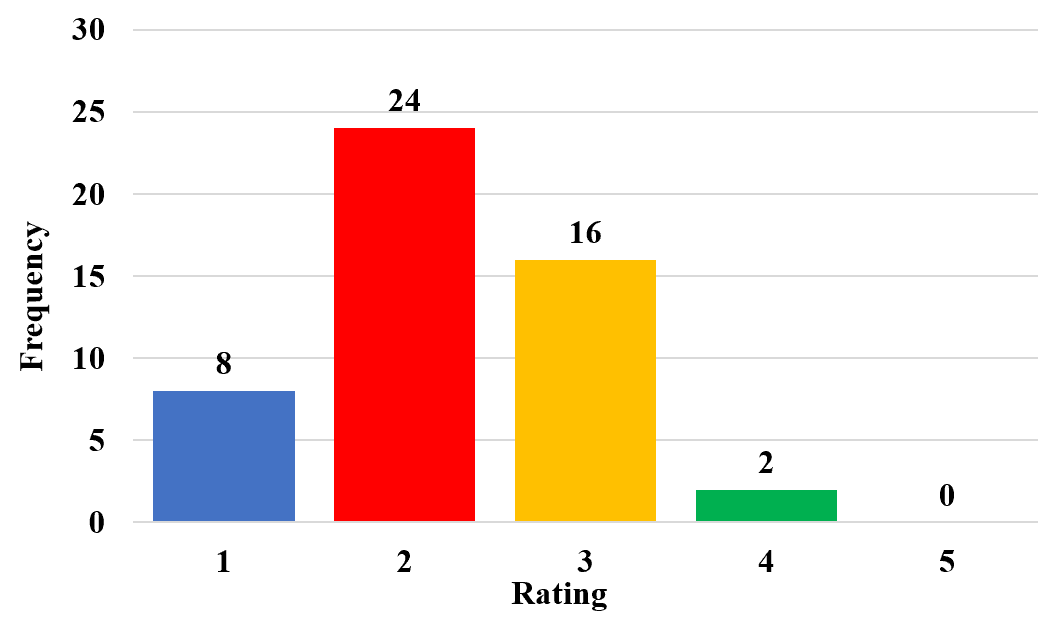}}
\label{subjective:dpm} 
\subfloat[WarpingGAN]{
\includegraphics[width=2.2in]{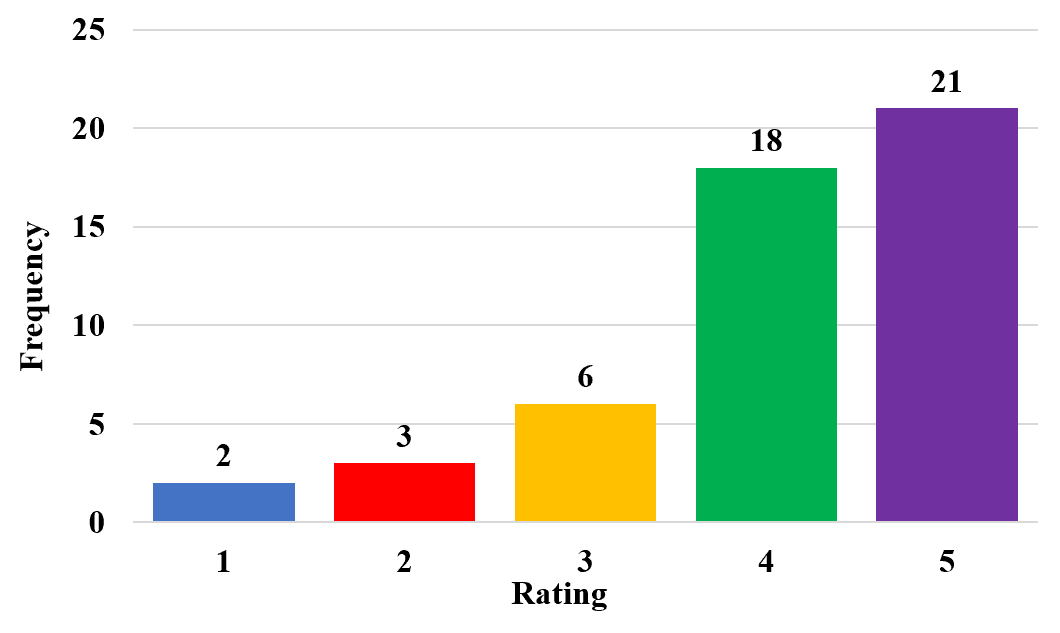}}
\label{subjective:warpinggan} 
\\
\subfloat[{Overall performance. The upper and bottom numbers are the std and mean values, respectively.}]{
\includegraphics[width=4.0in]{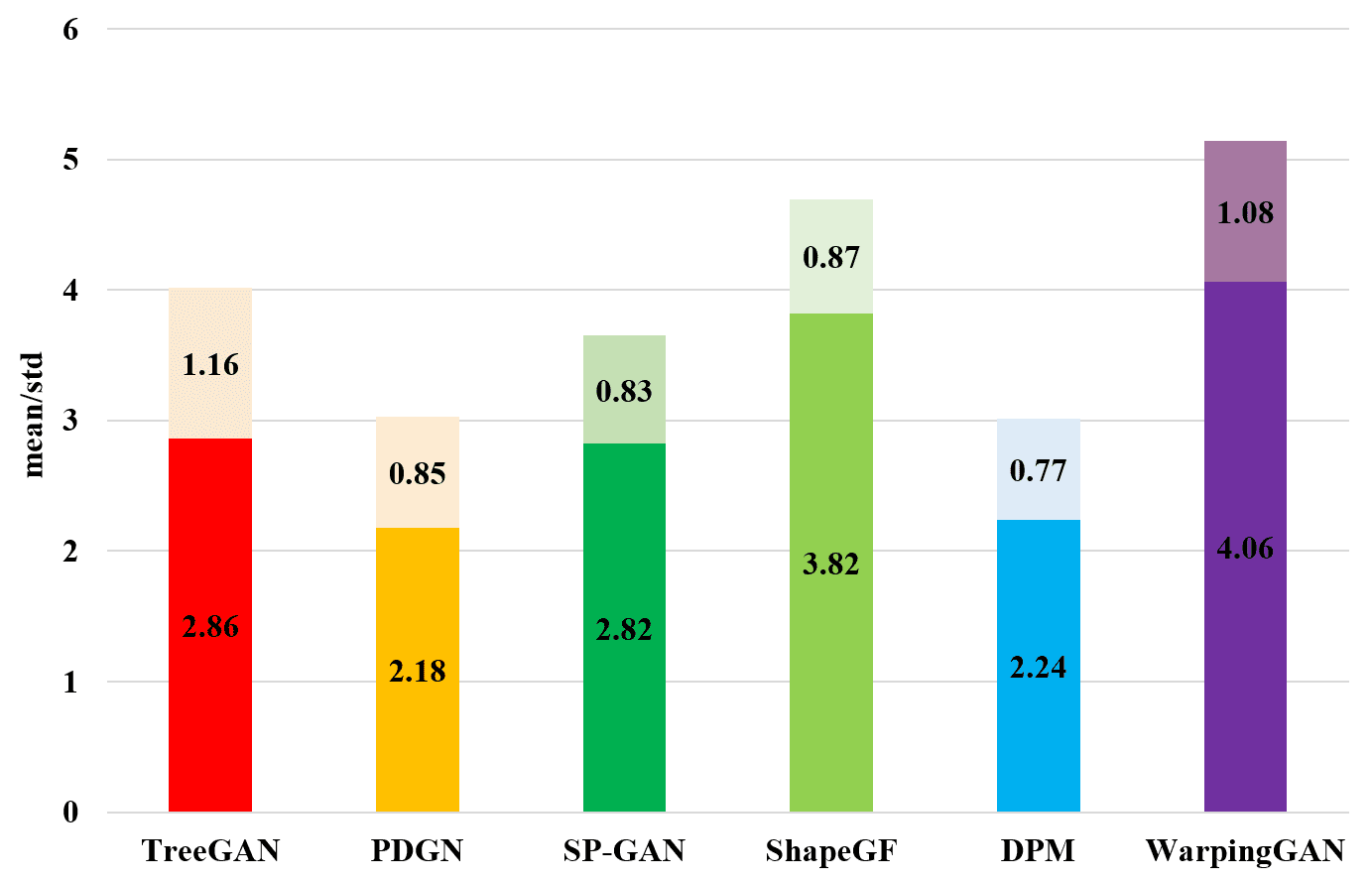}}
\label{subjective:meanstd} 
\caption{Results of the subjective evaluation on \textit{Chair}.}.
\label{subjective:chair} 
\end{figure}

\begin{figure}[h]
\centering
\subfloat[TreeGAN]{
\includegraphics[width=2.2in]{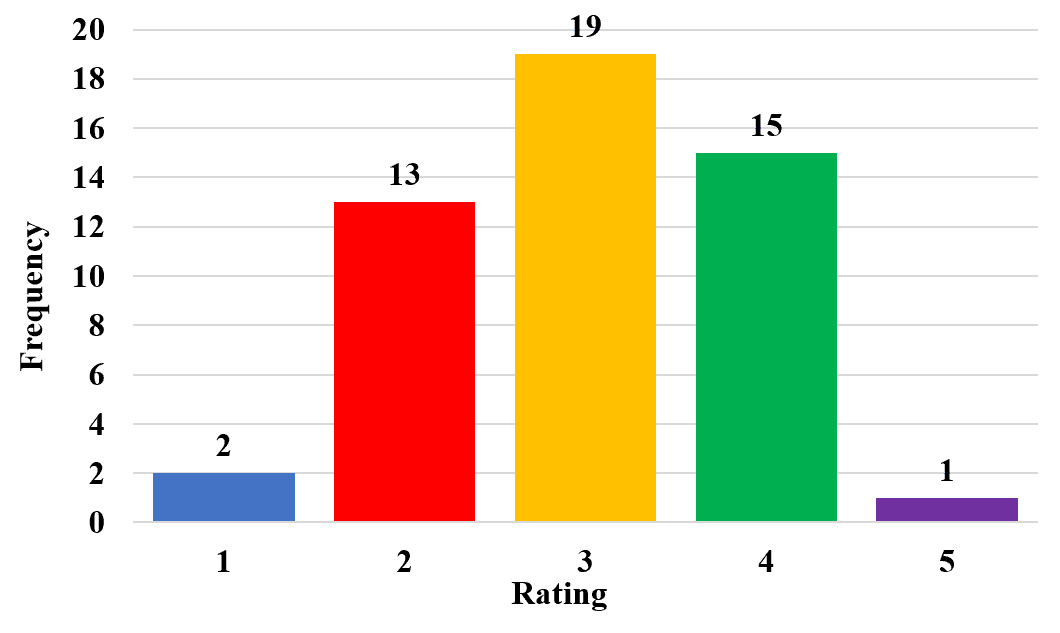}}
\label{subjectiveairplane:treegan} 
\subfloat[PDGN]{
\includegraphics[width=2.2in]{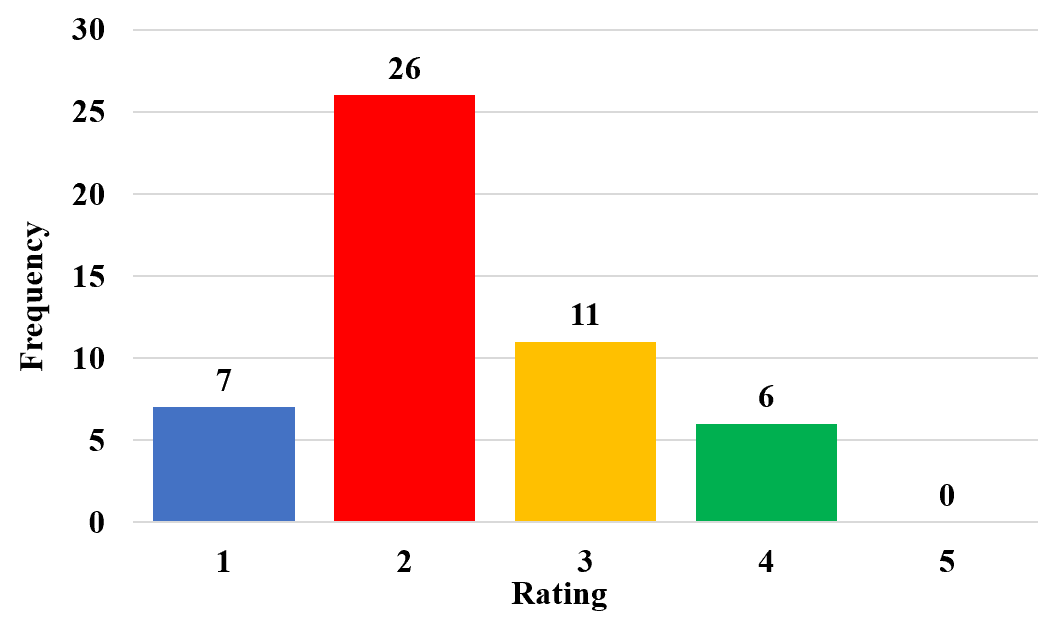}}
\label{subjectiveairplane:pdgn} 
\subfloat[SP-GAN]{
\includegraphics[width=2.2in]{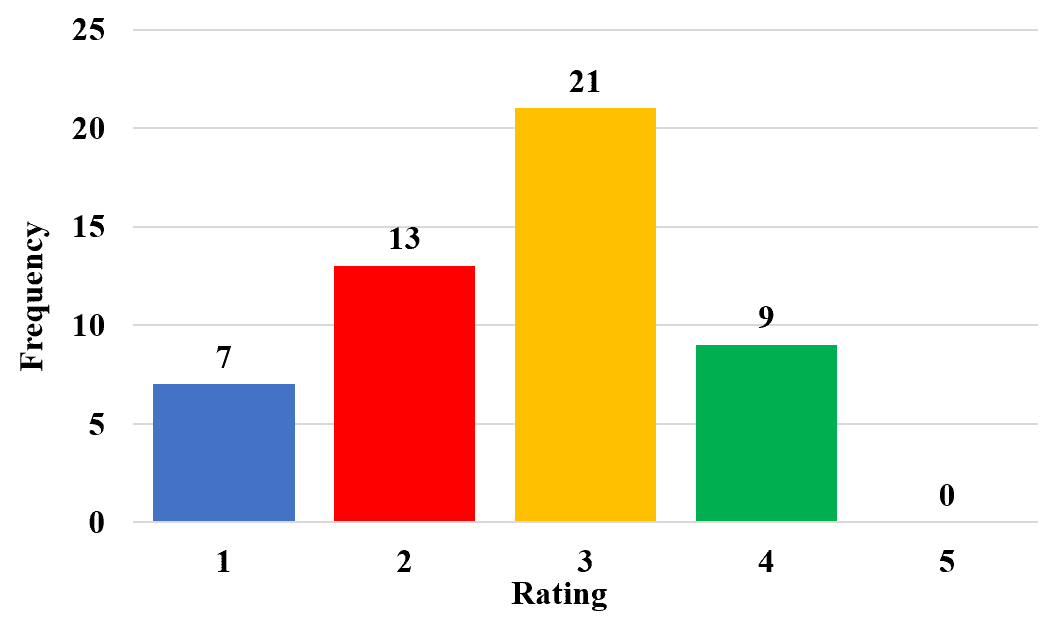}}
\label{subjectiveairplane:spgan} 
\\
\subfloat[ShapeGF]{
\includegraphics[width=2.2in]{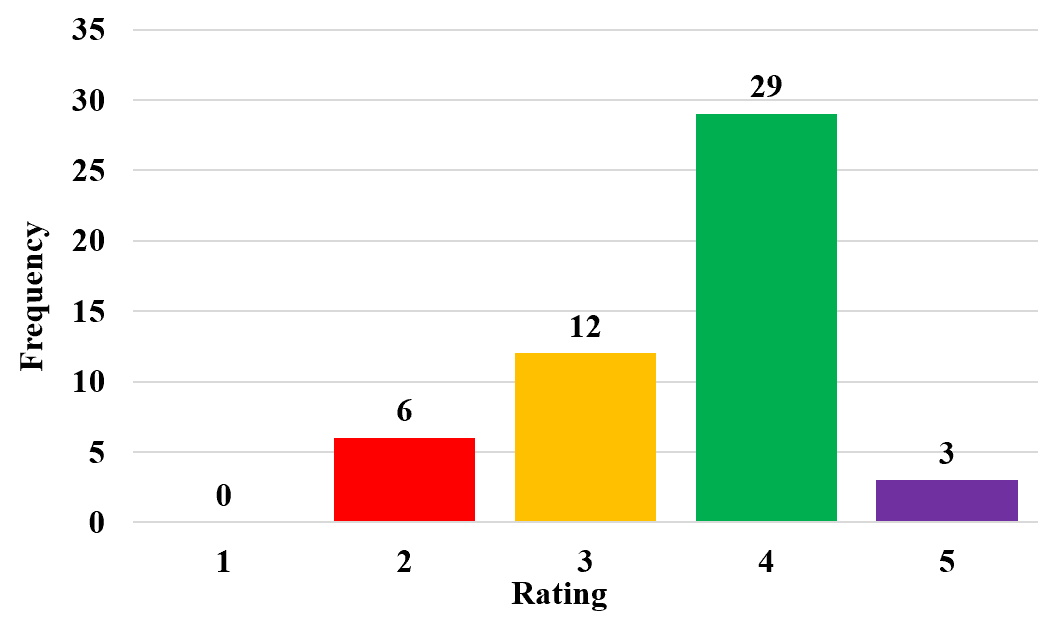}}
\label{subjectiveairplane:shapegf} 
\subfloat[DPM]{
\includegraphics[width=2.2in]{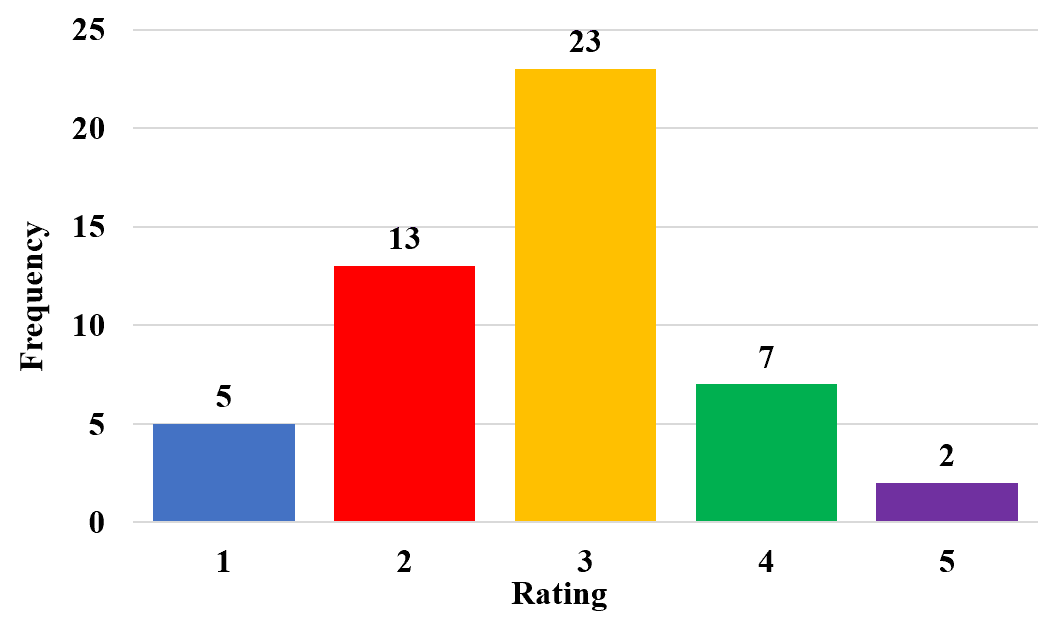}}
\label{subjectiveairplane:dpm} 
\subfloat[WarpingGAN]{
\includegraphics[width=2.2in]{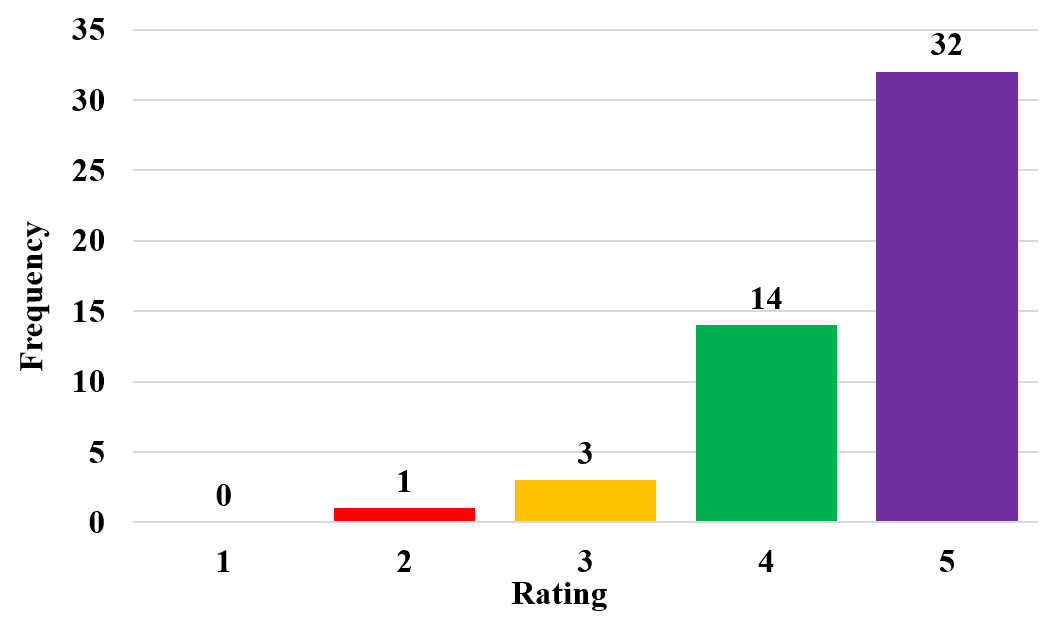}}
\label{subjectiveairplane:warpinggan} 
\\
\subfloat[{Overall performance. The upper and bottom numbers are the std and mean values, respectively.}]{
\includegraphics[width=4.0in]{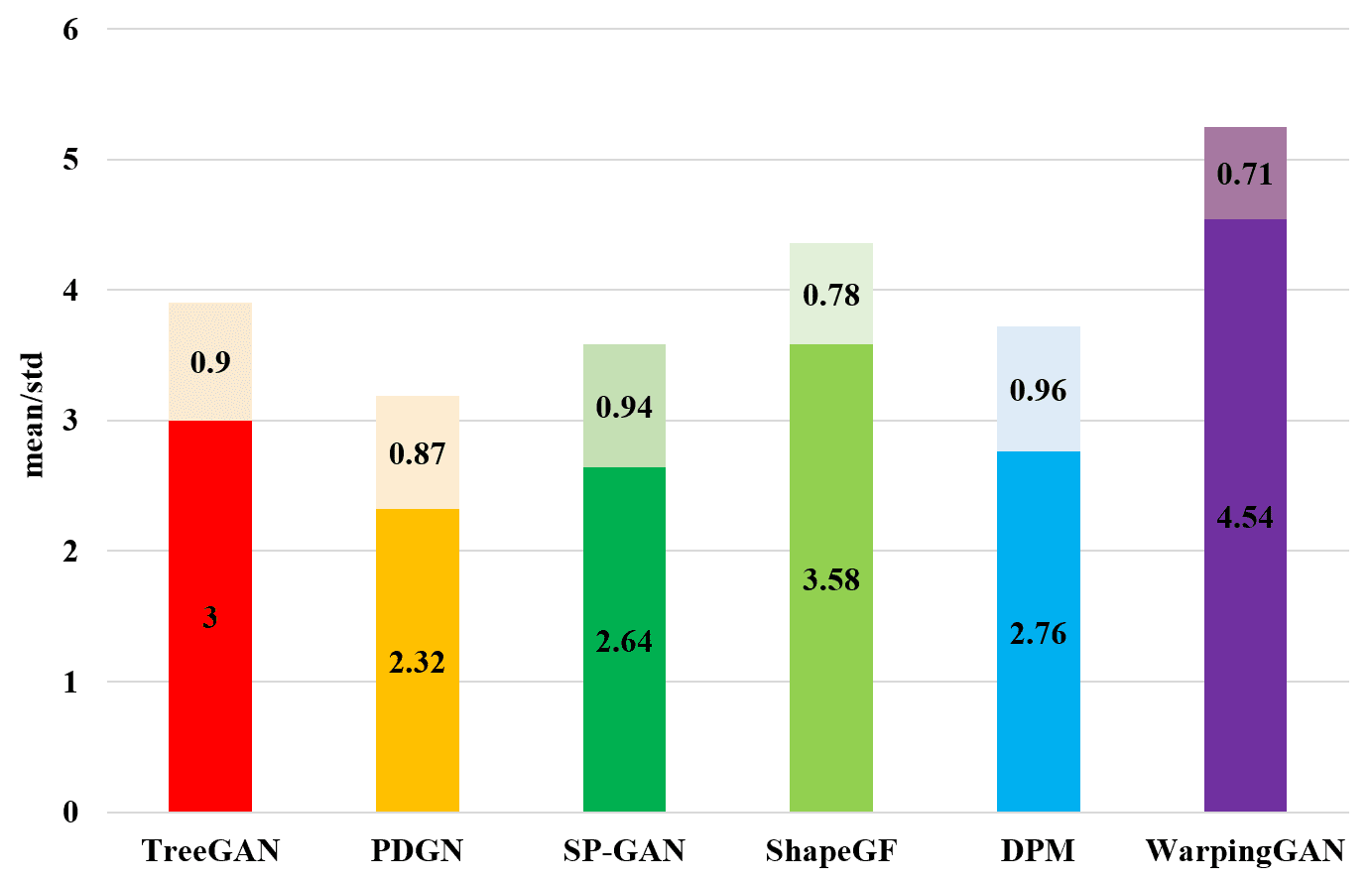}}
\label{subjectiveairplane:meanstd} 
\caption{Results of the subjective evaluation on \textit{Airplane}. }
\label{subjective:airplane} 
\end{figure}

\begin{figure}[h]
\centering
\subfloat[TreeGAN]{
\includegraphics[width=2.2in]{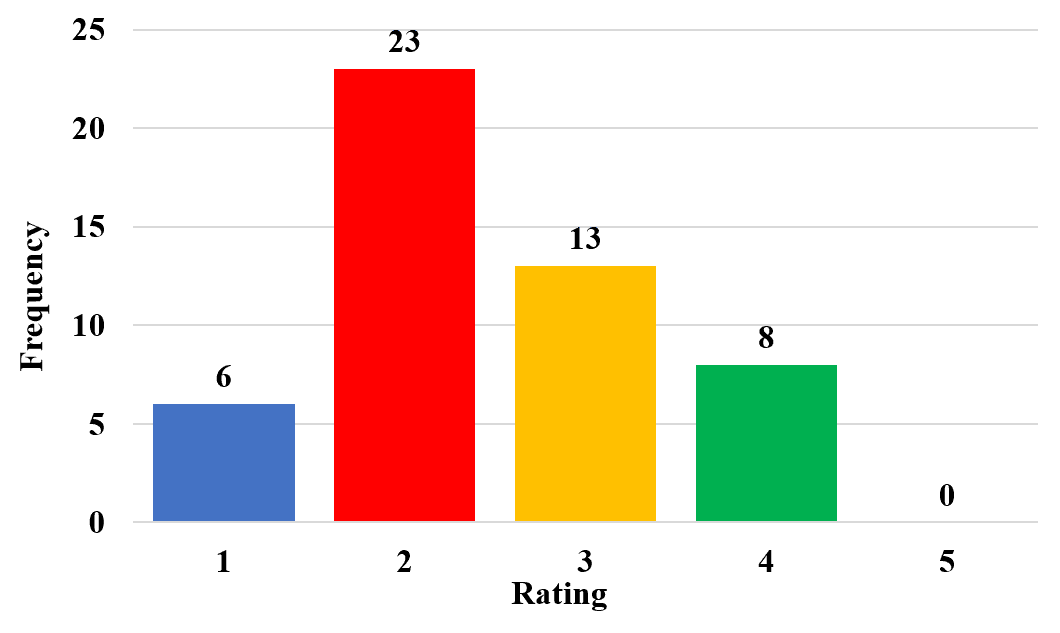}}
\label{subjectivecar:treegan} 
\subfloat[PDGN]{
\includegraphics[width=2.2in]{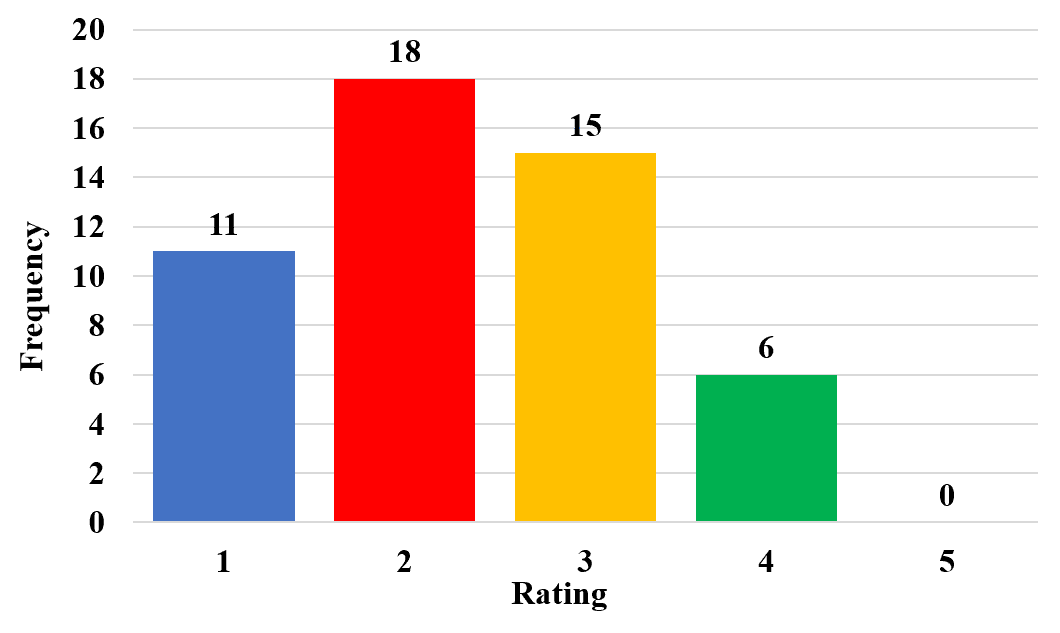}}
\label{subjectivecar:pdgn} 
\subfloat[SP-GAN]{
\includegraphics[width=2.2in]{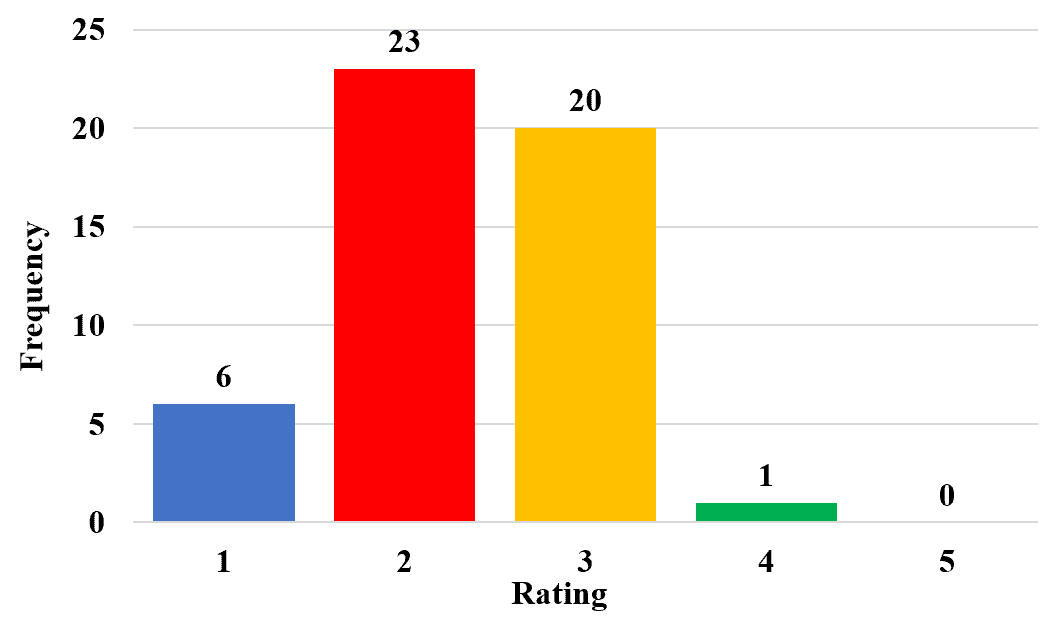}}
\label{subjectivecar:spgan} 
\\
\subfloat[ShapeGF]{
\includegraphics[width=2.2in]{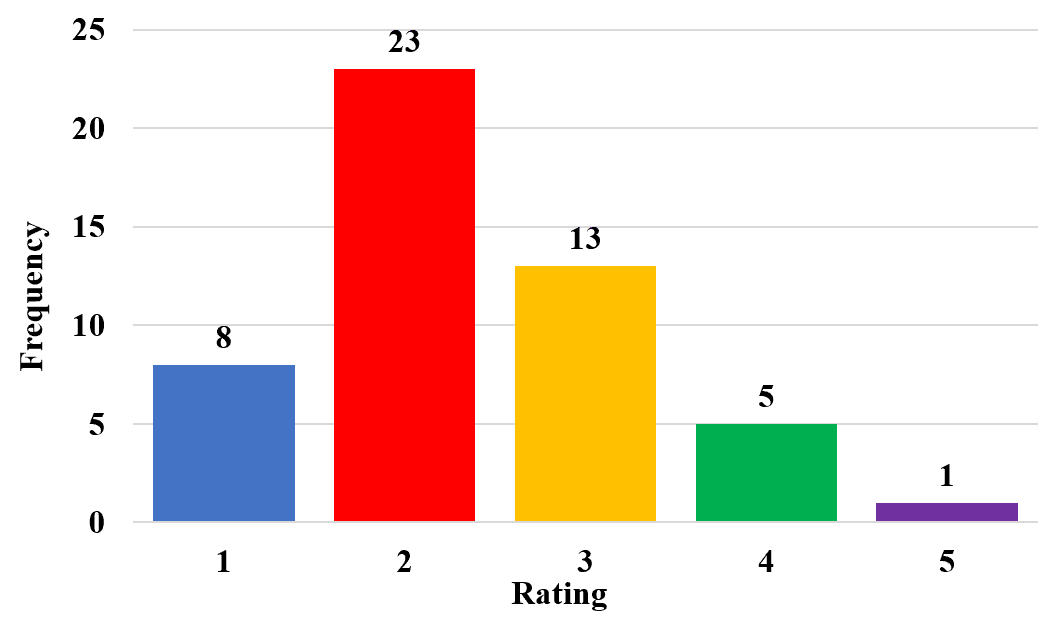}}
\label{subjectivecar:shapegf} 
\subfloat[DPM]{
\includegraphics[width=2.2in]{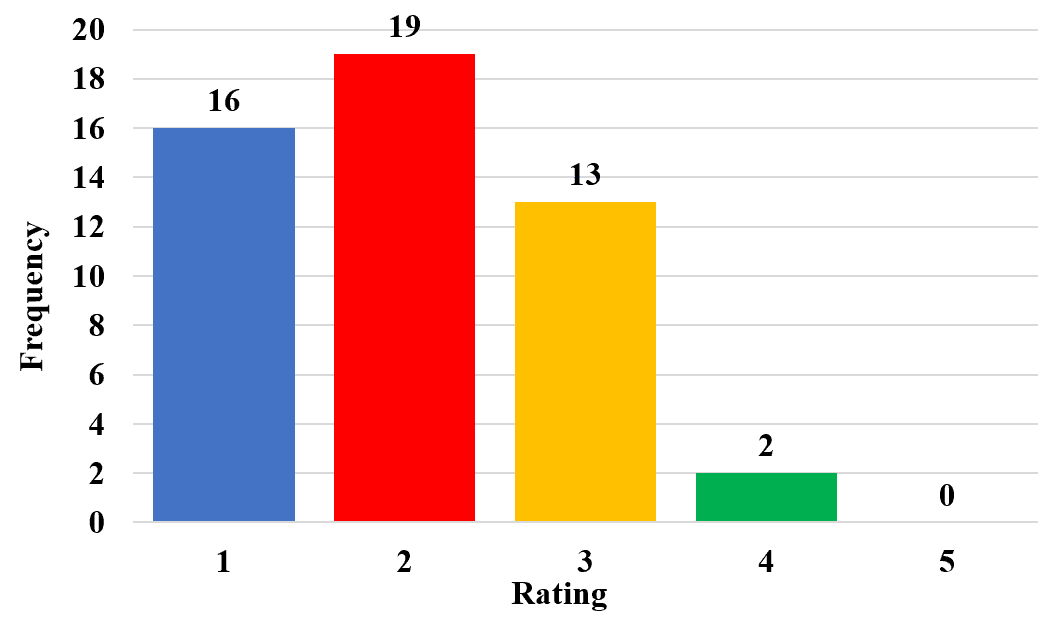}}
\label{subjectivecar:dpm} 
\subfloat[WarpingGAN]{
\includegraphics[width=2.2in]{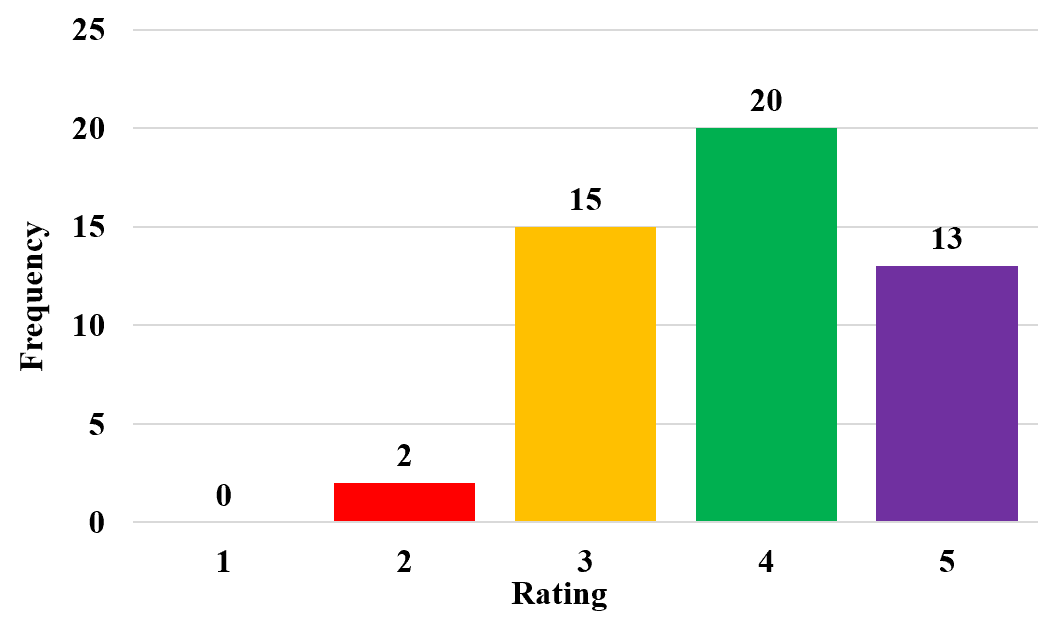}}
\label{subjectivecar:warpinggan} 
\\
\subfloat[Overall performance. The upper and bottom numbers are the std and mean values, respectively.]{
\includegraphics[width=4.0in]{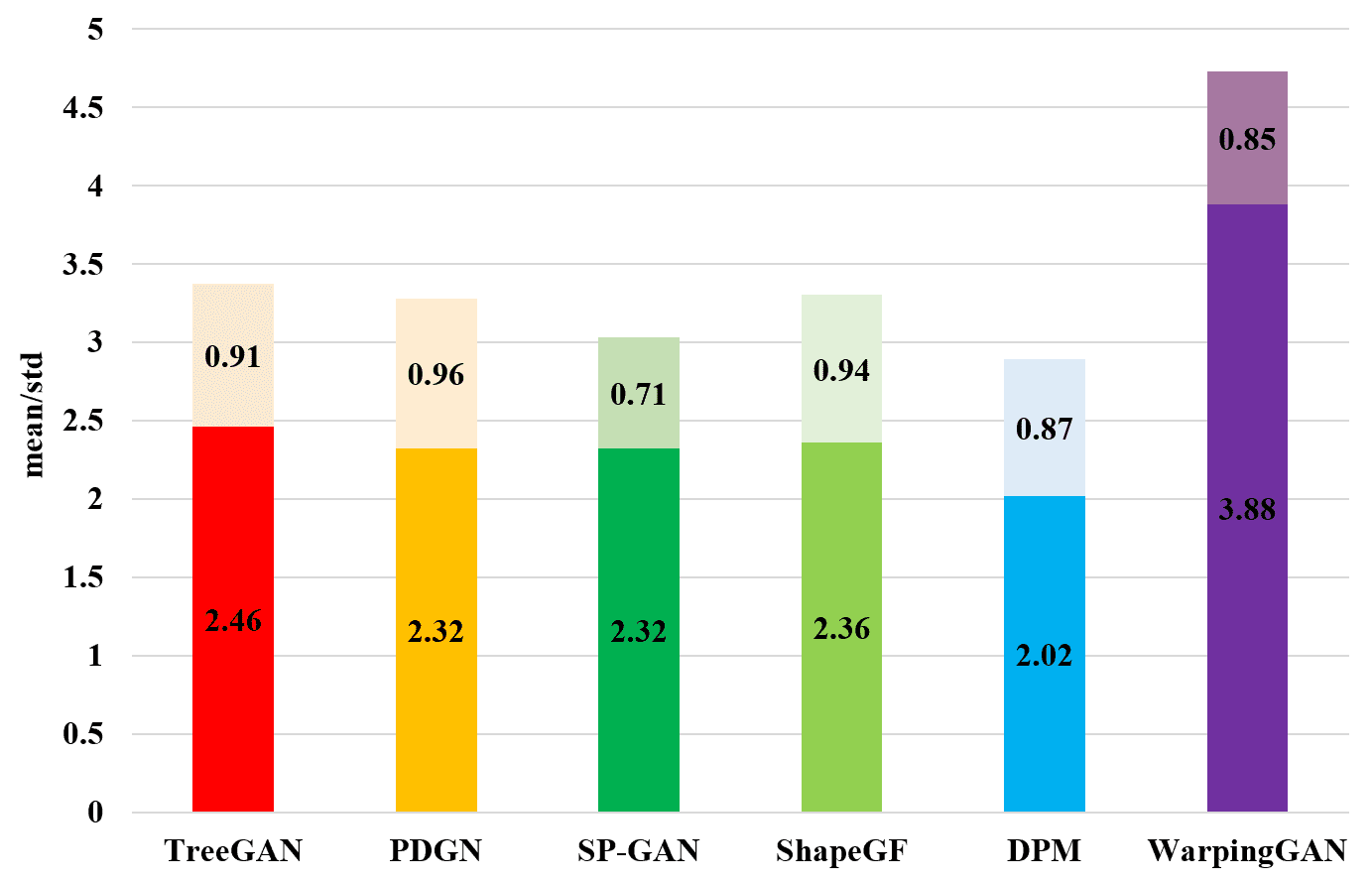}}
\label{subjectivecar:meanstd} 
\caption{Results of the subjective evaluation on  \textit{Car}. }
\label{subjective:car} 
\end{figure}

Figs.~\ref{subjective:chair},~\ref{subjective:airplane}, and \ref{subjective:car} show the results of the subjective evaluation, where we provided the score distribution of each method and  the mean value and the standard deviation (std) of the scores. It can be seen that our WarpingGAN consistently obtains the highest mean scores 
for all three categories. Particularly, for \textit{Chair} and \textit{Airplane}, most volunteers rated our WarpingGAN with 5, while rating the other methods with 2 $\sim$ 4. 
For \textit{Car}, the compared methods only obtain the mean scores around 2, while the  mean score of our WarpingGAN is about 4. This subjective evaluation convincingly demonstrate the advantage of our WarpingGAN over state-of-the-art methods in terms of the quality of generated data.

\section{More Visual Results}

\subsection{Visual Results of Ablation Studies on Airplane}
\label{subsec:more ablation}
\begin{figure}
\centering
\subfloat[without code enhancement]{
\includegraphics[width=3.3in]{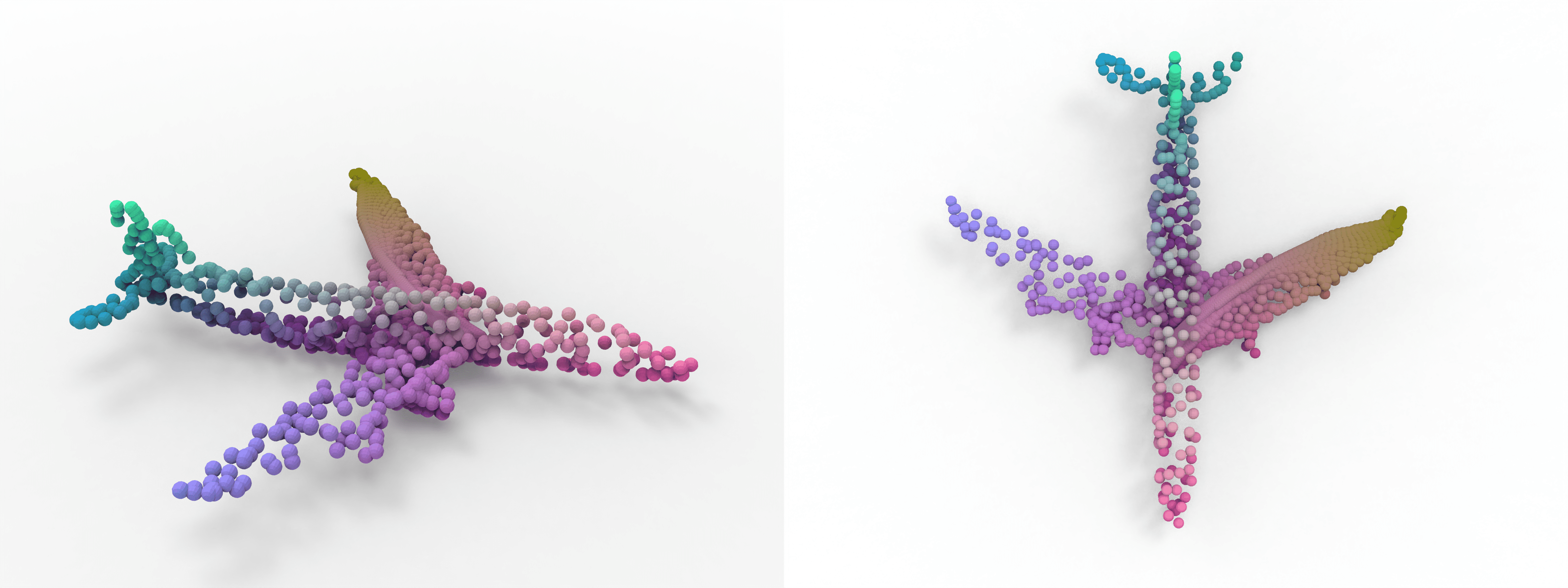}}
\label{a1:wow} 
\subfloat[with code enhancement]{
\includegraphics[width=3.3in]{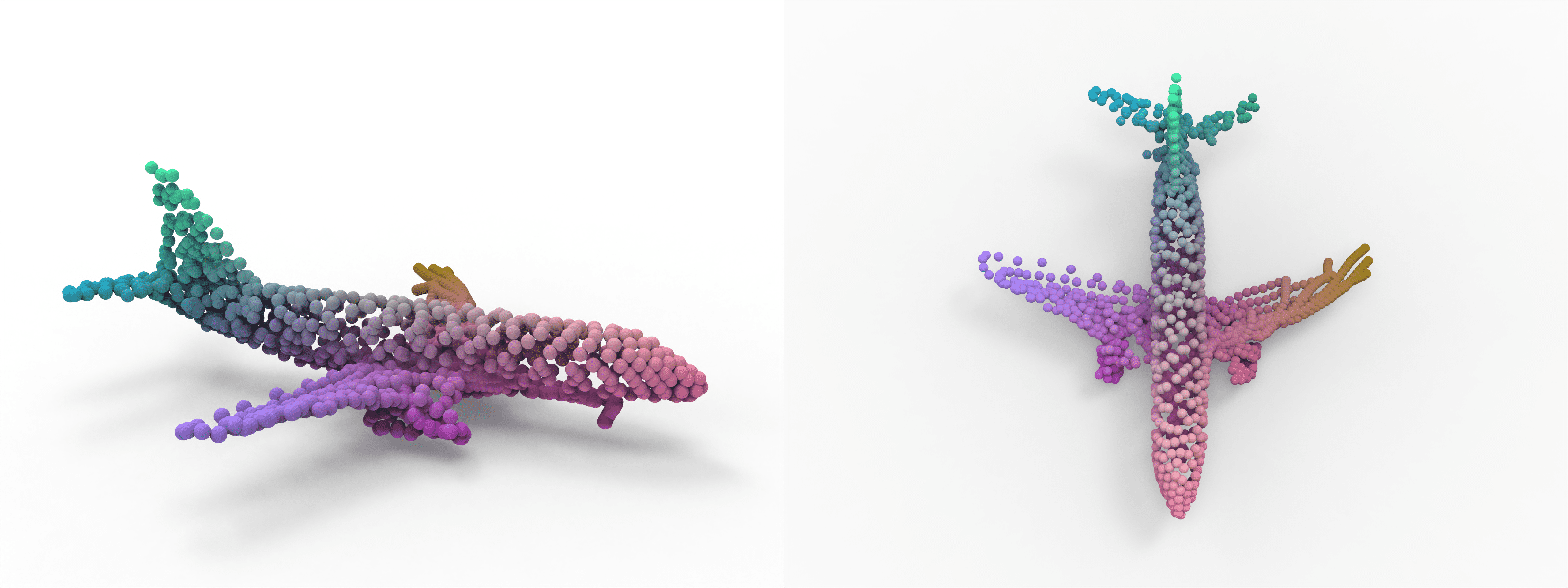}}
\label{a1:wog} 
\caption{Visual comparison of our WarpingGAN (a) without  and (b) with code enhancement.}
\label{ab:code}
\end{figure}

\begin{figure}
\centering
\subfloat[without global shape code]{
\includegraphics[width=3.3in]{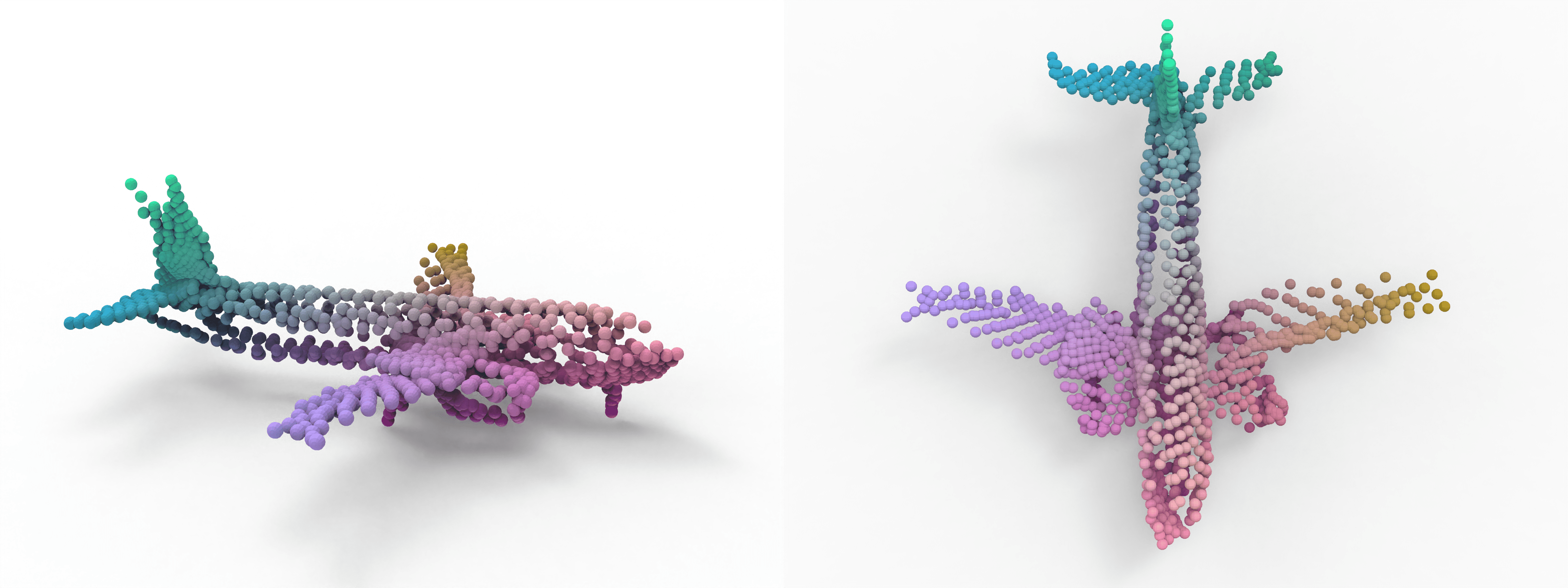}}
\label{a1:wow} 
\subfloat[with global shape code]{
\includegraphics[width=3.3in]{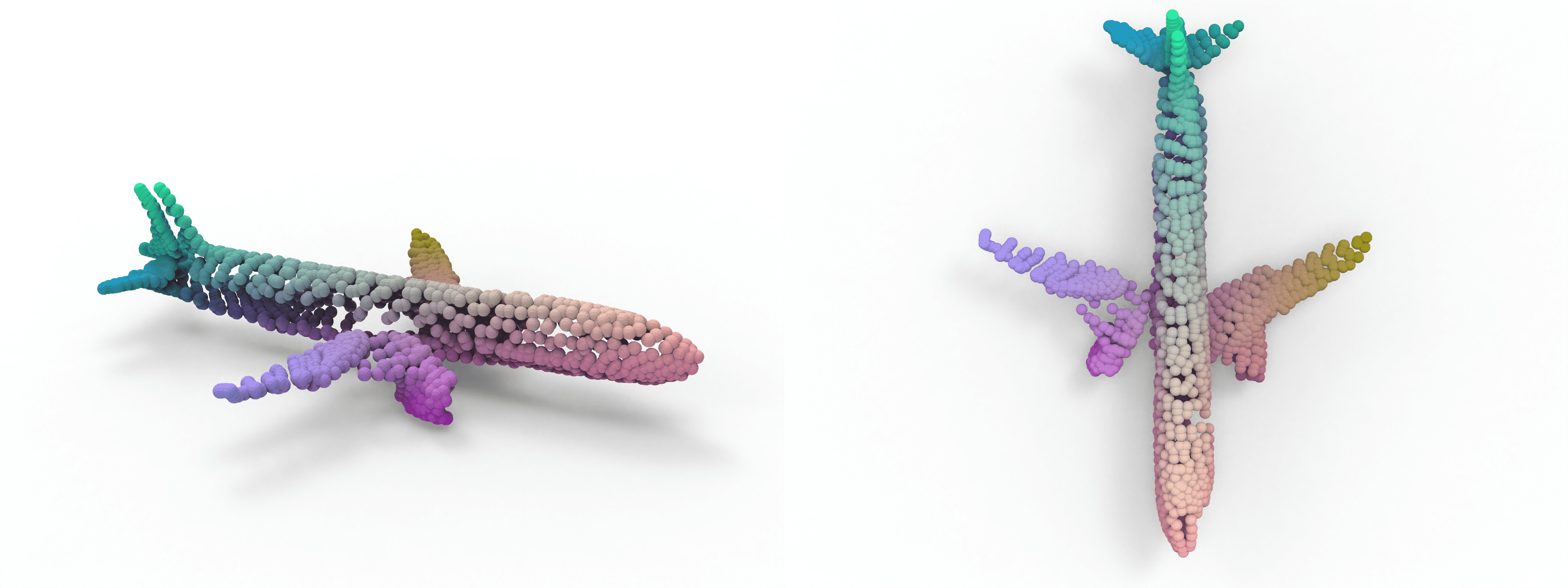}}
\label{a1:wog} 
\caption{Visual comparison of our WarpingGAN (a) without and (b) with global shape code.}
\label{ab:global}
\end{figure}

\begin{figure}
\centering
\subfloat[2D prior]{
\includegraphics[width=3.3in]{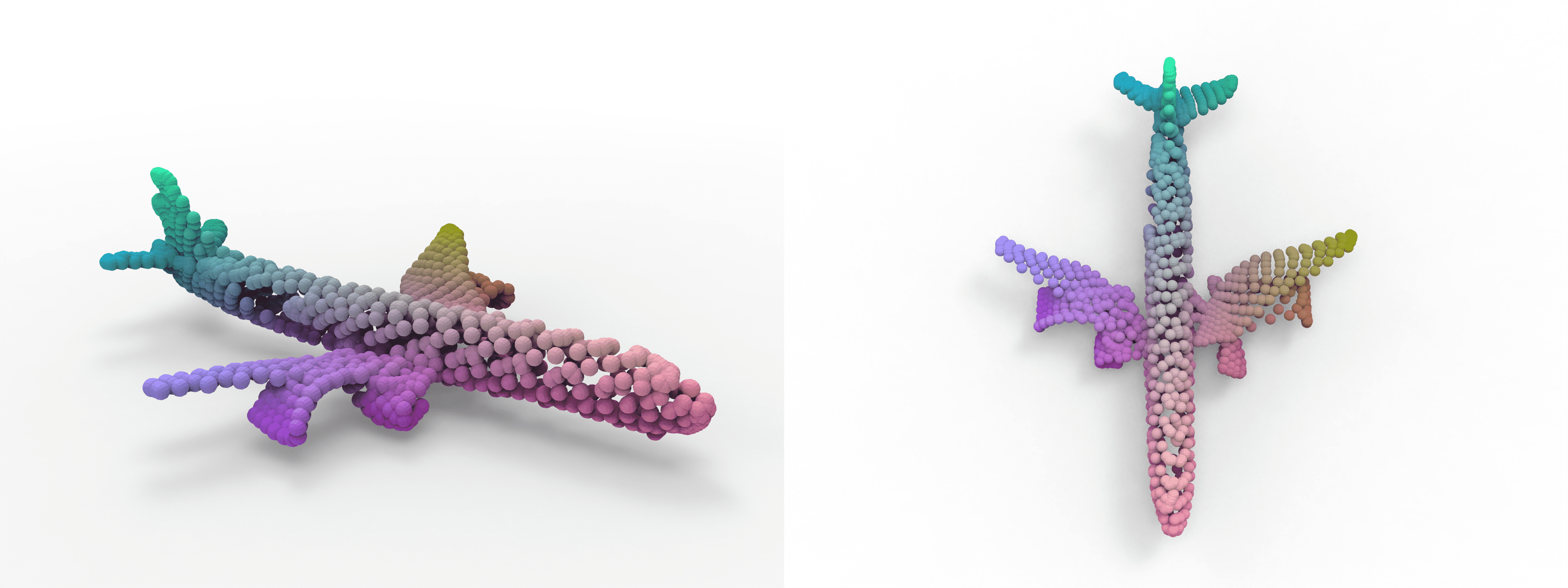}}
\label{a1:woe} 
\subfloat[3D prior]{
\includegraphics[width=3.3in]{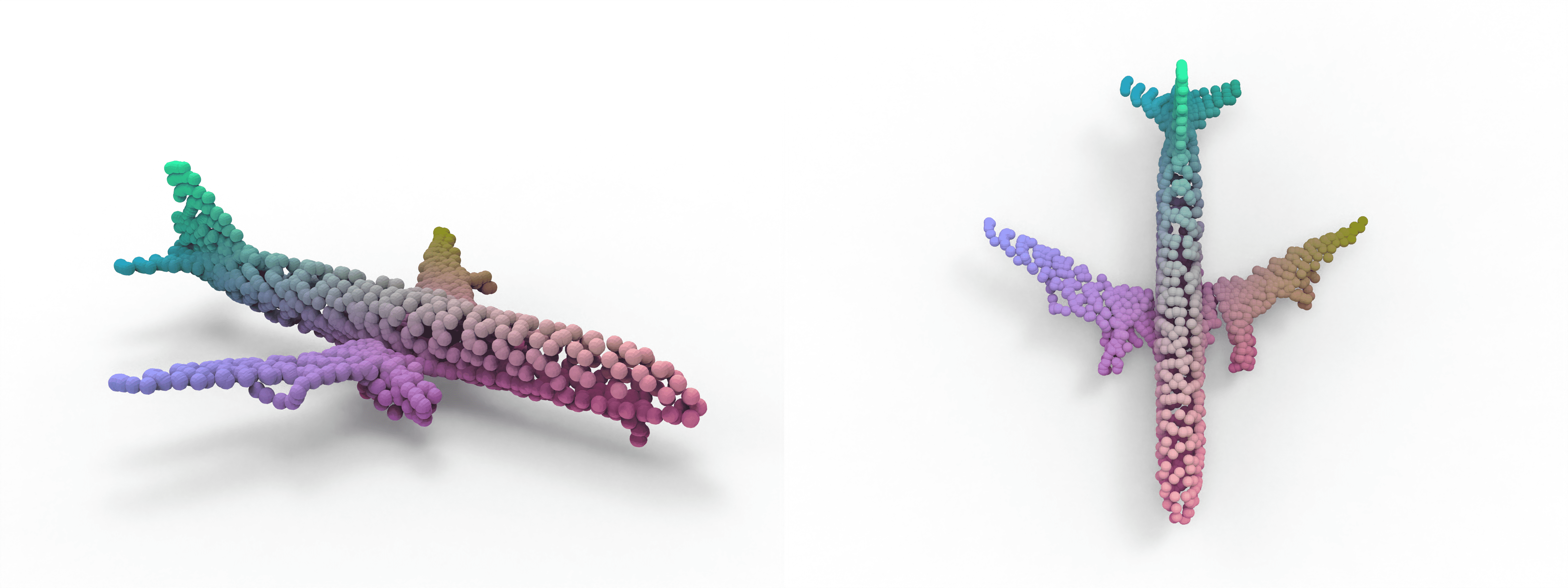}}
\label{a1:we} 
\caption{Visual comparison of our WarpingGAN equipped with (a) 2D and (b) 3D priors.}
\label{ab:3d}
\end{figure}

\begin{figure}
\centering
\subfloat[non-uniform priors]{
\includegraphics[width=3.3in]{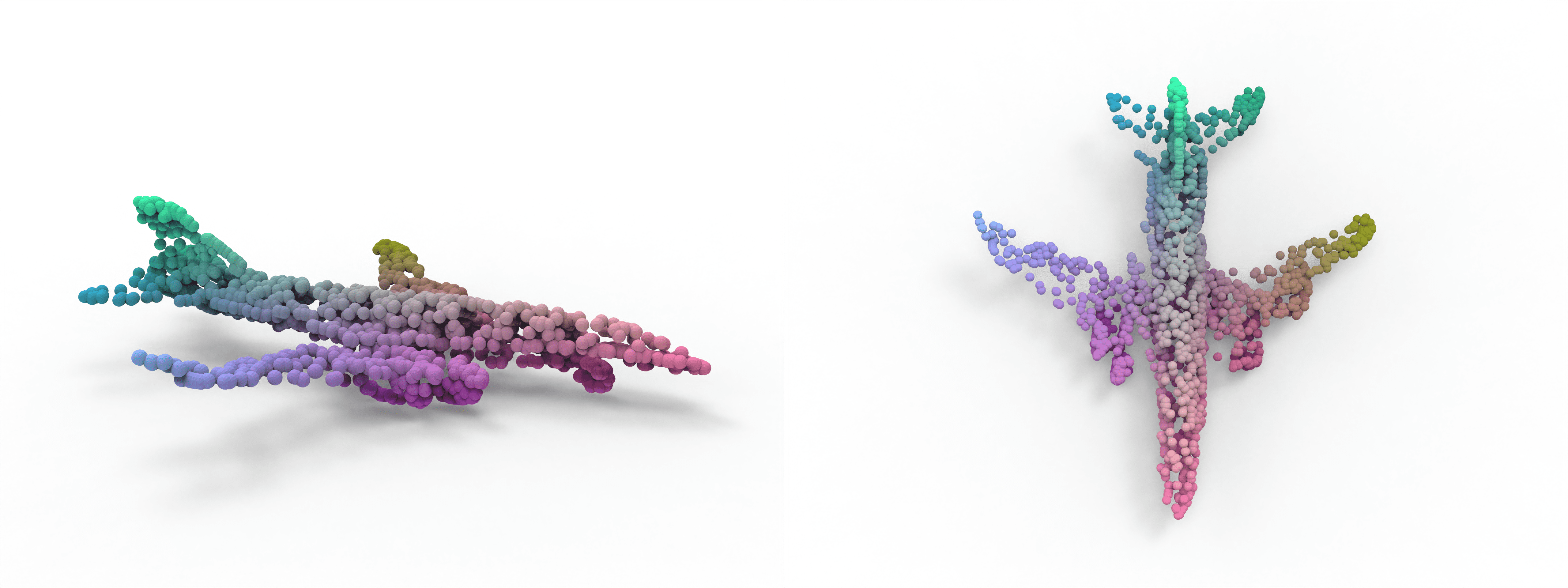}}
\label{a1:wow} 
\subfloat[uniform priors]{
\includegraphics[width=3.3in]{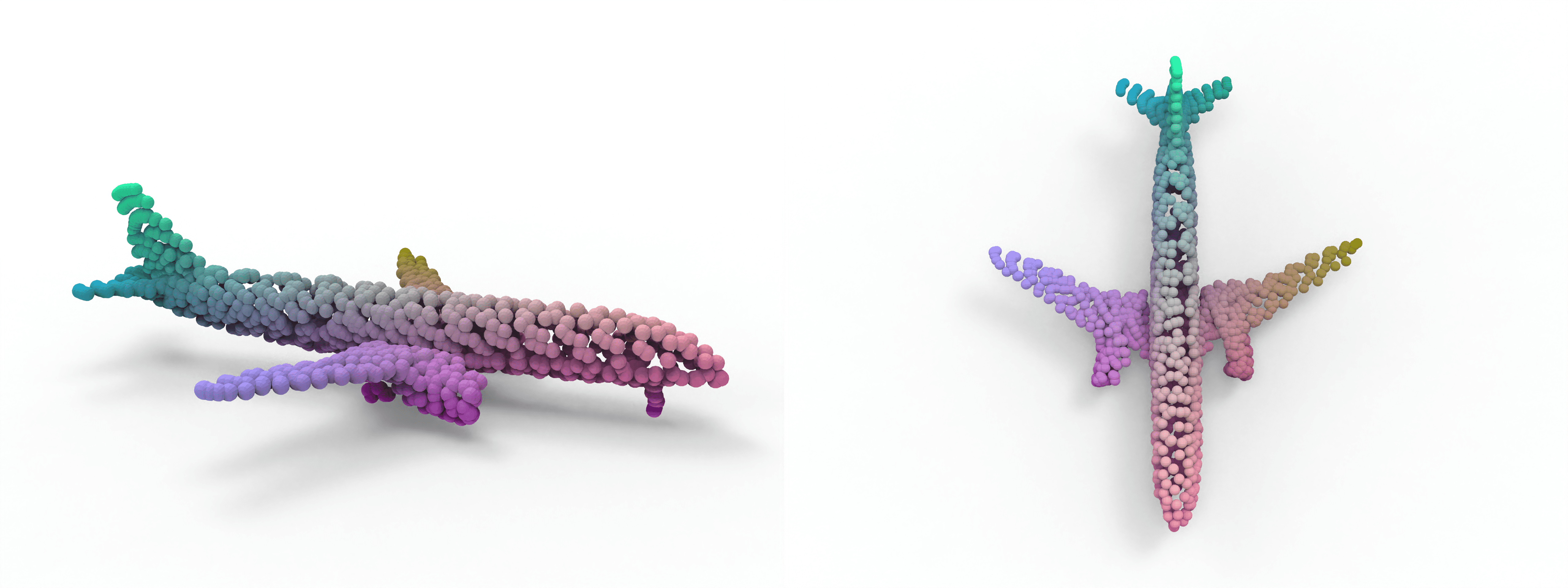}}
\label{a1:wog} 
\caption{Visual comparison of our WarpingGAN equipped with (a) non-uniform and (b) uniform 3D priors.}
\label{ab:uniform}
\end{figure}

In Section \textcolor{red}{4.3} of the manuscript, we conducted the ablation studies of the proposed WarpingGAN quantitatively and visually over the \textit{Chair} category. Here, we also provided the visual results of the ablation studies on the \textit{Airplane} category. We followed the settings of Section \textcolor{red}{4.3}, including the ablation studies towards the code enhancement module (Fig.~\ref{ab:code}), the global shape code (Fig.~\ref{ab:global}), the 2D vs. 3D priors (Fig.~\ref{ab:3d}), and the non-uniform vs. uniform priors (Fig.~\ref{ab:uniform}). Here we omitted the ablation study towards the stitching loss on \textit{Airplane}, since it has already been demonstrated in the manuscript (i.e., Section \textcolor{red}{4.3} \textbf{The effectiveness of the stitching loss} of the manuscript). These results further demonstrate the effectiveness of each module of our WarpingGAN.

\subsection{Visual Illustration of More Shapes}
\label{subsec:more shapes}
\begin{figure}
\centering
	\includegraphics[width=6in]{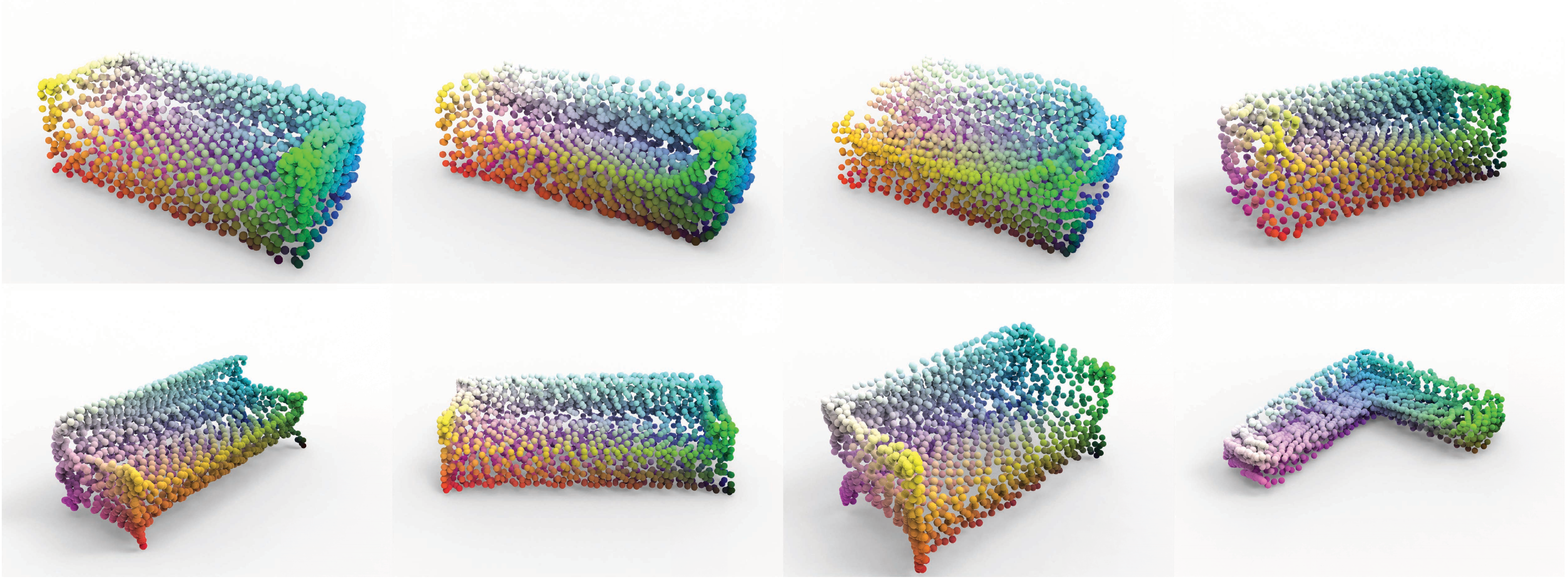}
	\caption{\textit{Sofa} generated by WarpingGAN.}
	\label{sofa}
\end{figure}

\begin{figure}
\centering
	\includegraphics[width=6in]{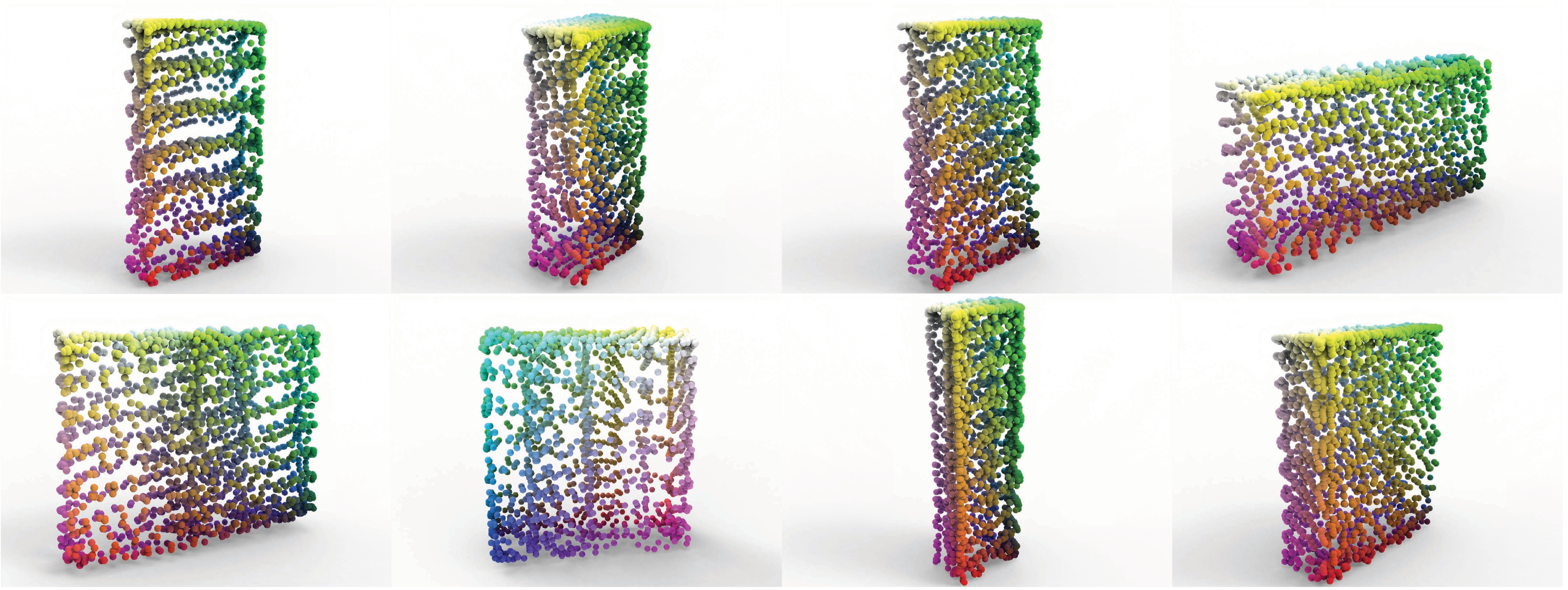}
	\caption{\textit{Cabinet} generated by WarpingGAN.}
	\label{cabinet}
\end{figure}

\begin{figure}
\centering
	\includegraphics[width=6in]{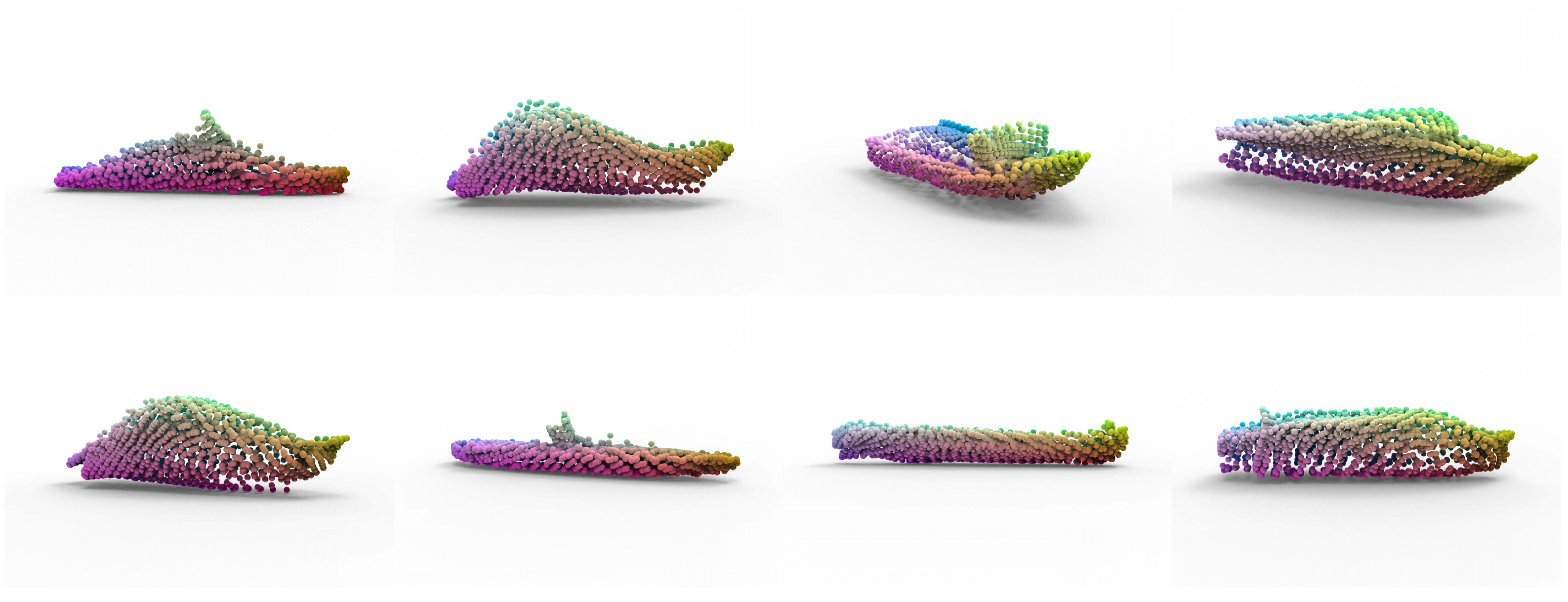}
	\caption{\textit{Vessel} generated by WarpingGAN.}
	\label{vessel}
\end{figure}

\begin{figure}
\centering
	\includegraphics[width=6in]{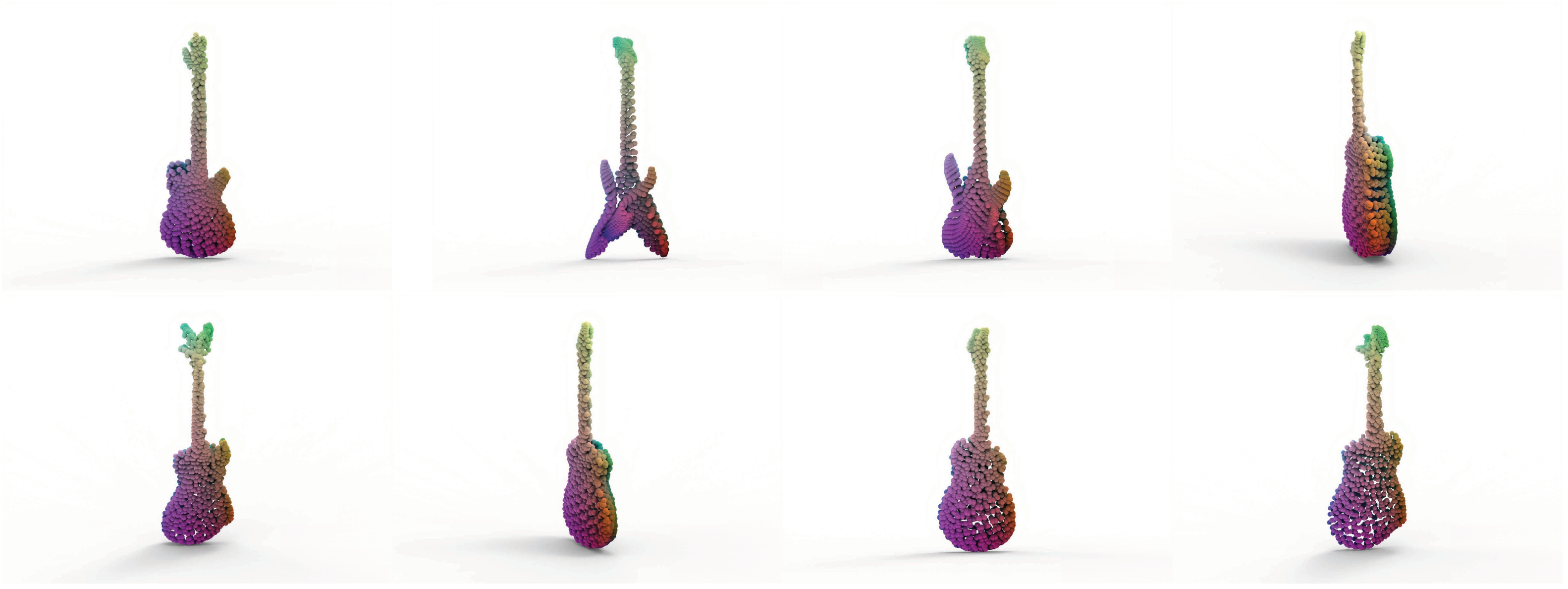}
	\caption{\textit{Guitar} generated by WarpingGAN.}
	\label{guitar}
\end{figure}

\begin{figure}
\centering
	\includegraphics[width=6in]{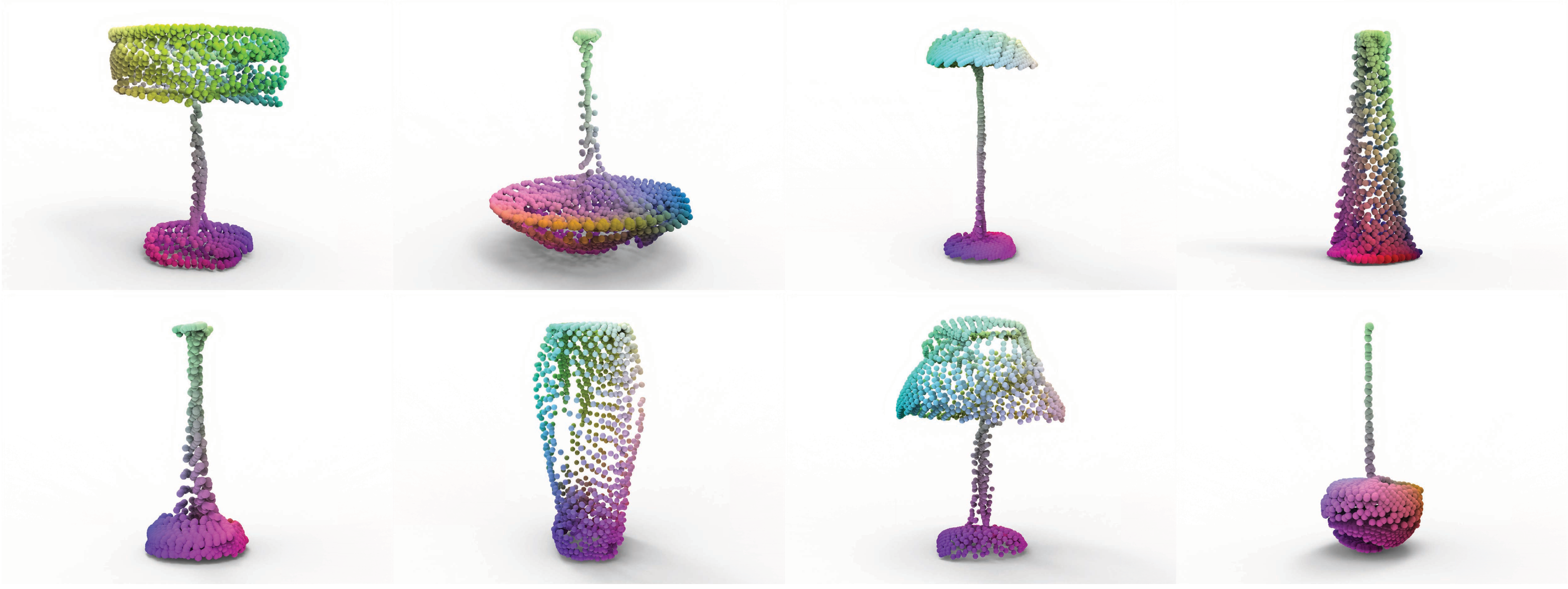}
	\caption{\textit{Lamp} generated by WarpingGAN.}
	\label{lamp}
\end{figure}

\begin{figure}
\centering
	\includegraphics[width=6in]{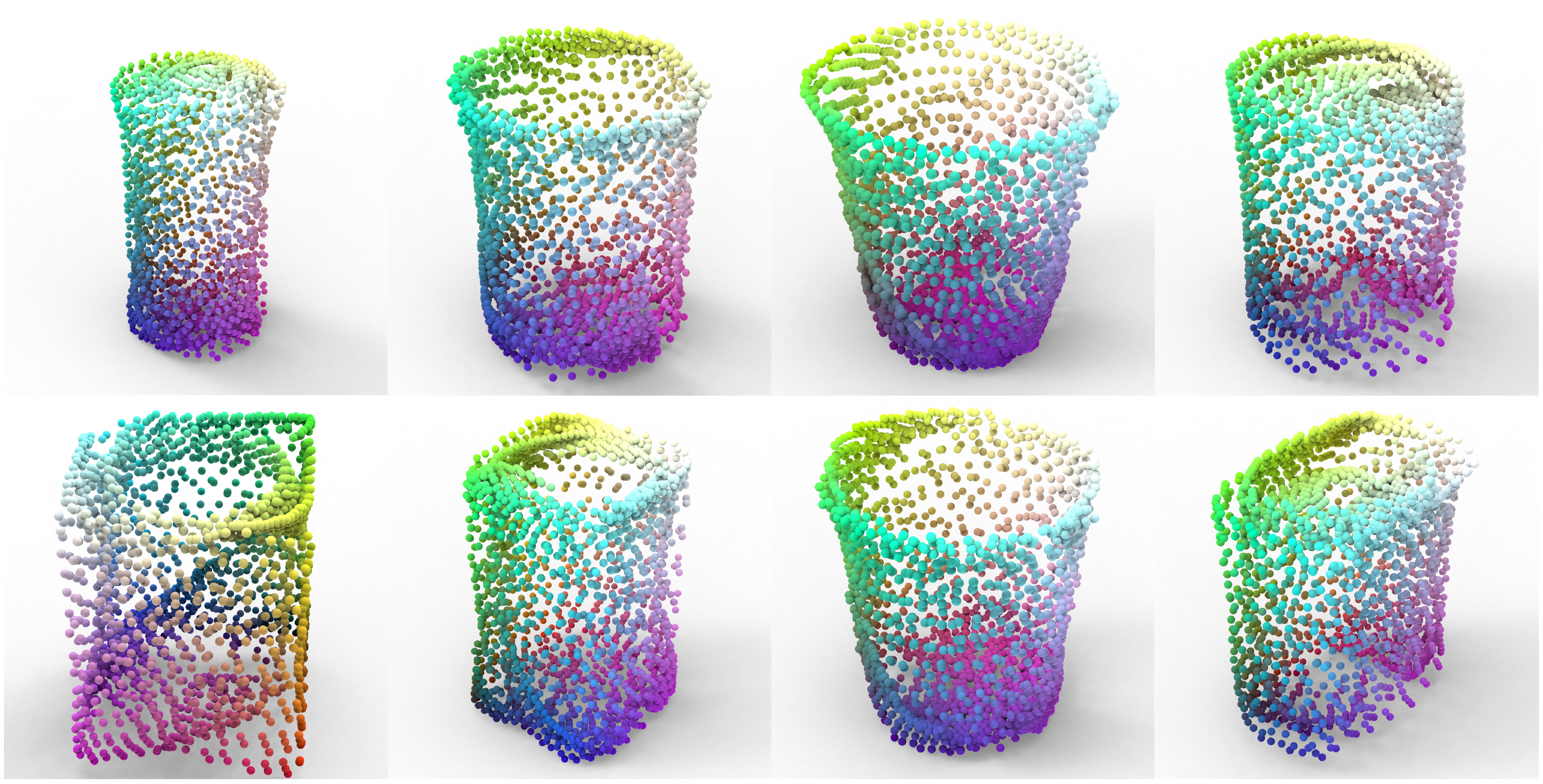}
	\caption{\textit{Can} generated by WarpingGAN.}
	\label{can}
\end{figure}

\begin{figure}
\centering
	\includegraphics[width=6in]{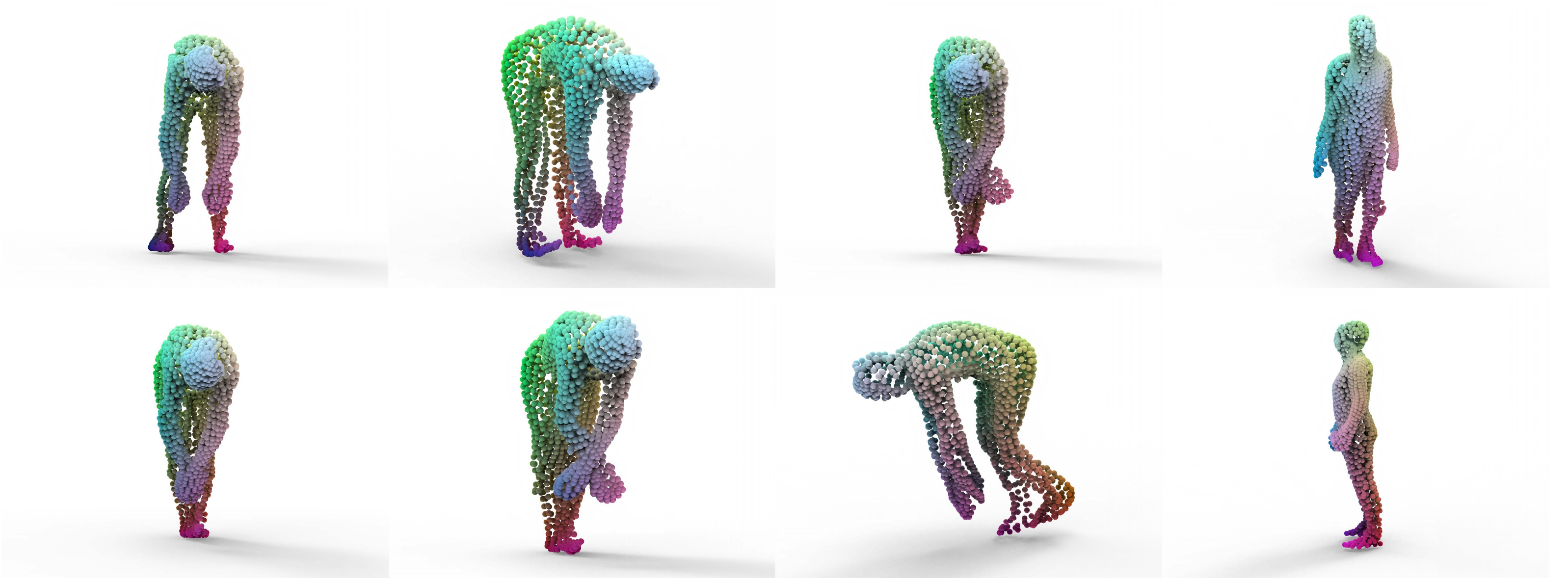}
	\caption{\textit{Human Body} generated by WarpingGAN.}
	\label{human}
\end{figure}

We provided visual results of generated 3D point clouds by our WarpingGAN for more categories, including \textit{Sofa} (Fig.~\ref{sofa}), \textit{Cabinet} (Fig.~\ref{cabinet}), \textit{Vessel} (Fig.~\ref{vessel}), \textit{Guitar} (Fig.~\ref{guitar}), \textit{Lamp} (Fig.~\ref{lamp}), \textit{Can} (Fig.~\ref{can}) and \textit{Human} (Fig.~\ref{human}).


\subsection{Video Demo}
\label{subsec:demo}
We provided a video demo (\textit{demo.mp4}) to compare the quality of generated shapes by different methods. Note that we also utilized the shapes in this video for the subjective evaluation.

\subsection{Failure and Low-Quality Cases}
\label{subsec:failure}
\begin{figure}
\centering
\subfloat[Chair]{
\includegraphics[width=1.2in]{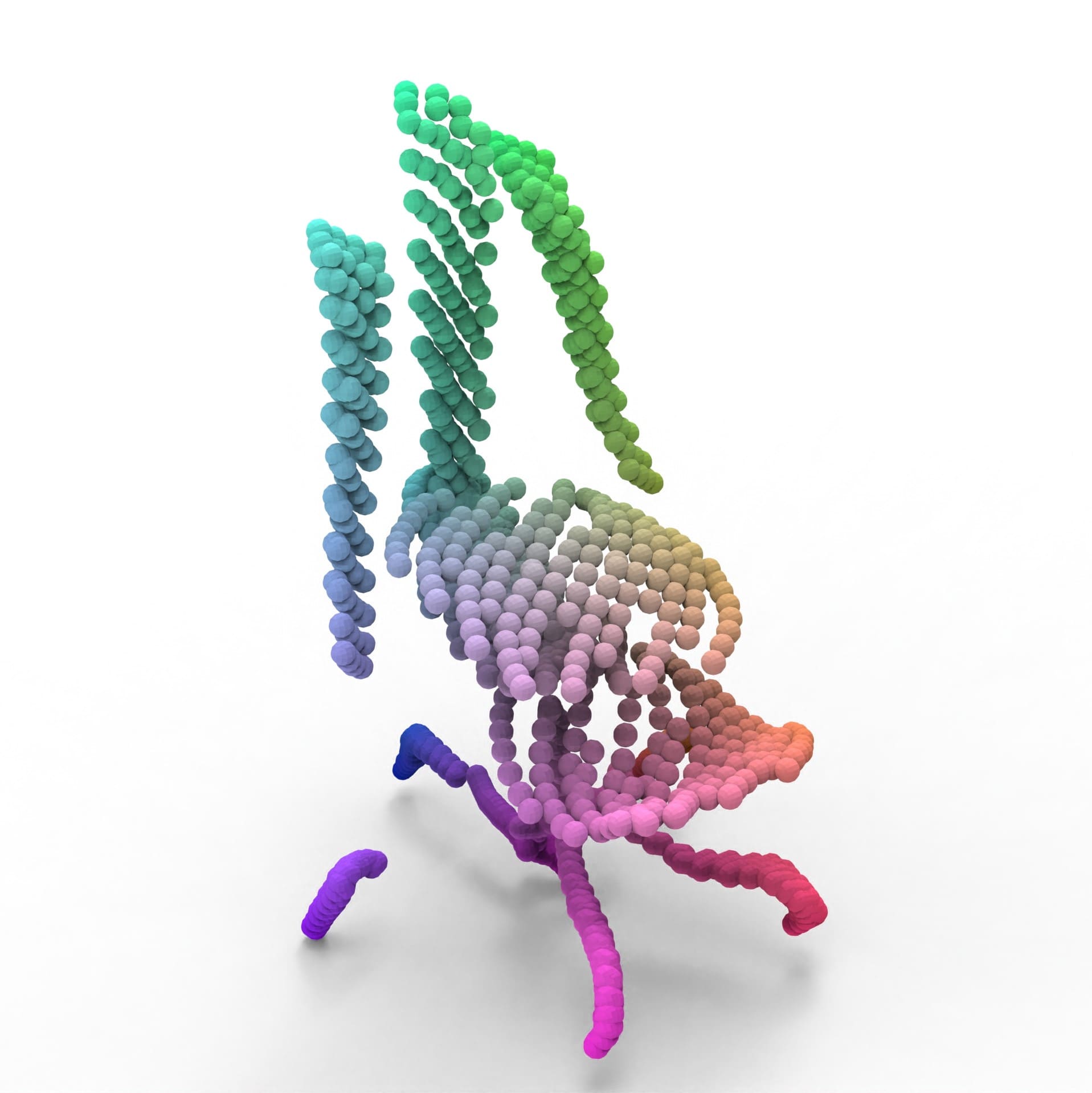}}
\label{failure:chair} 
\subfloat[Airplane]{
\includegraphics[width=1.2in]{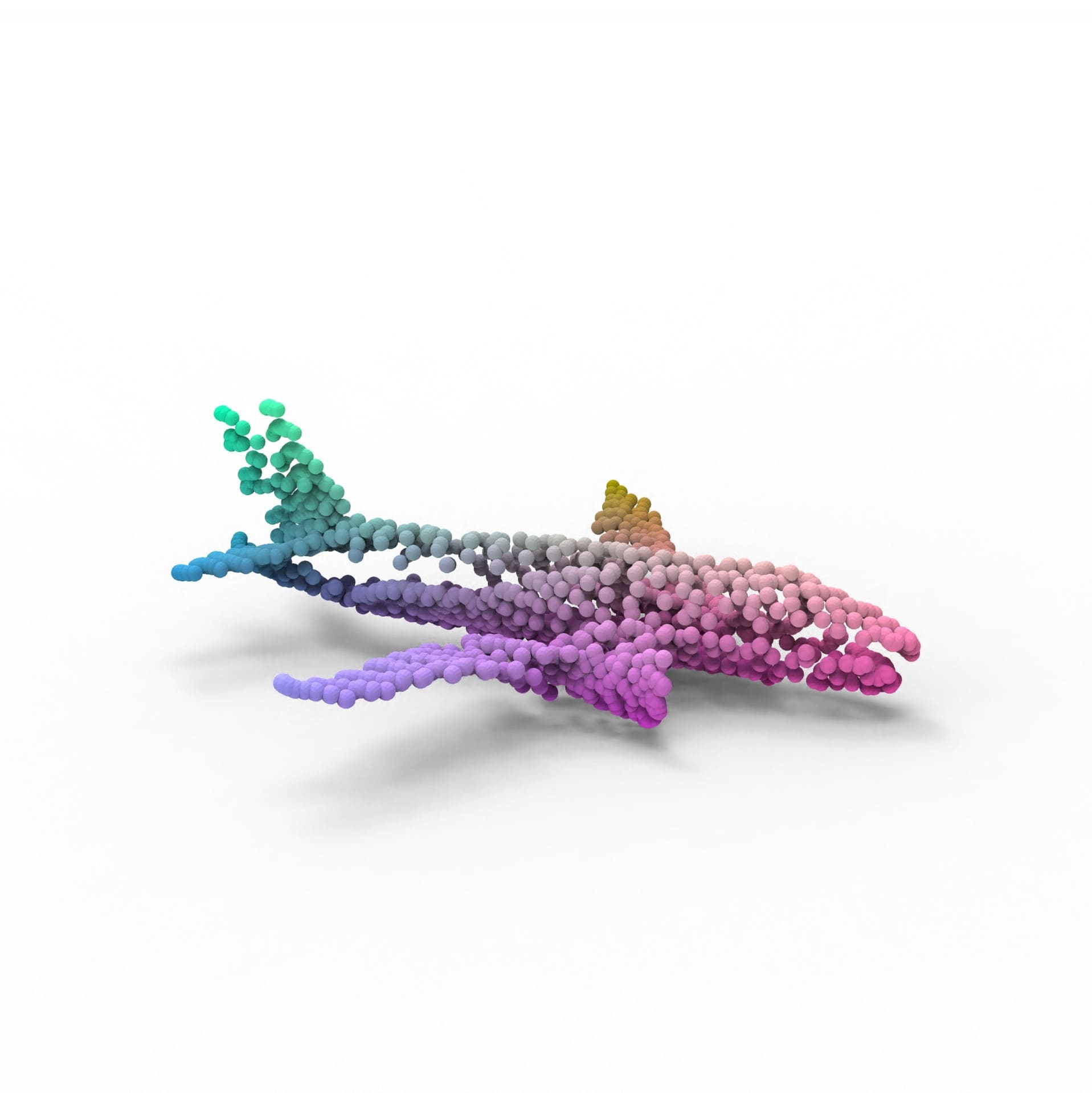}}
\label{failure:airplane} 
\subfloat[Lamp]{
\includegraphics[width=1.2in]{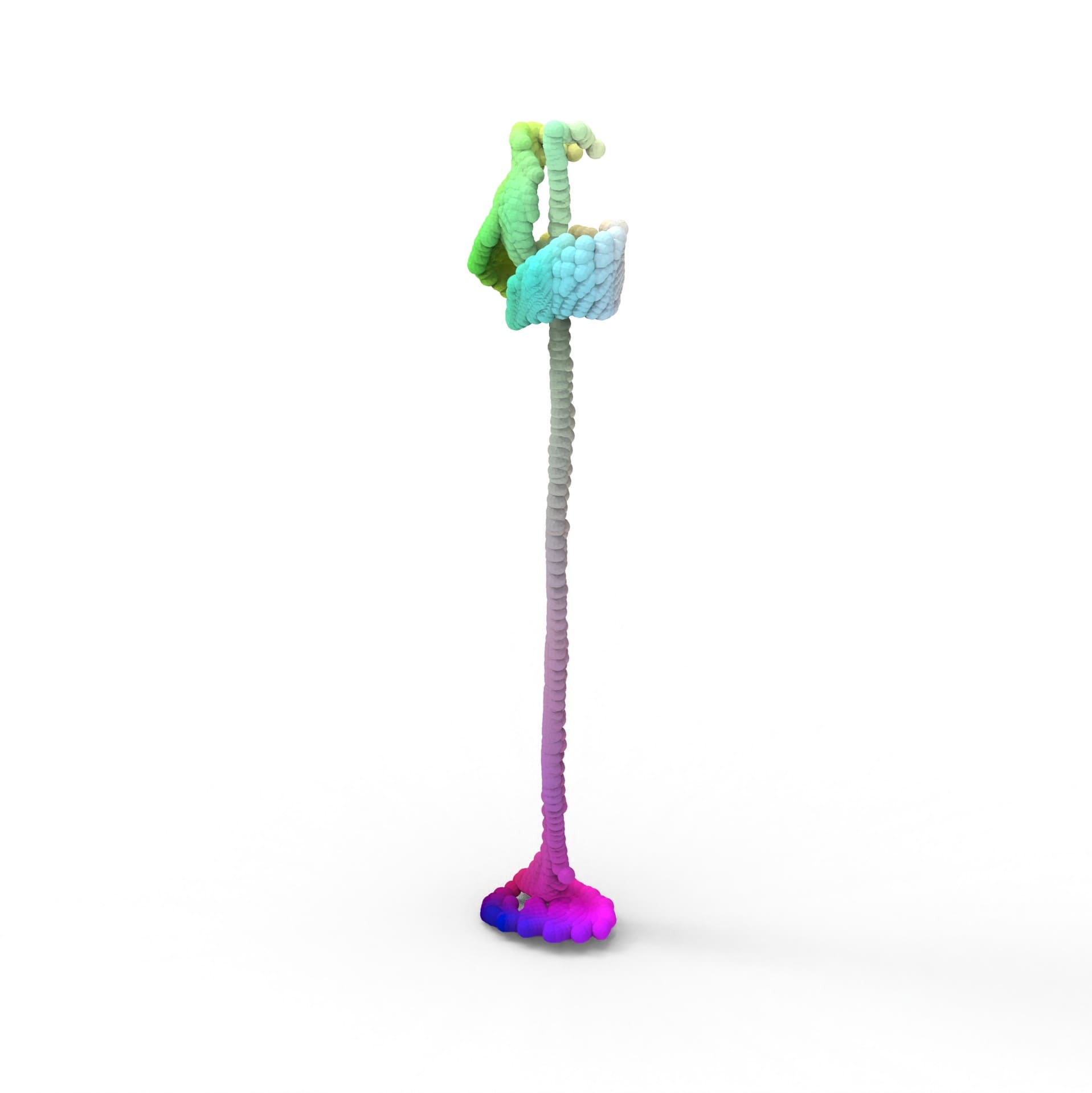}}
\label{failure:lamp} 
\subfloat[Vessel]{
\includegraphics[width=1.2in]{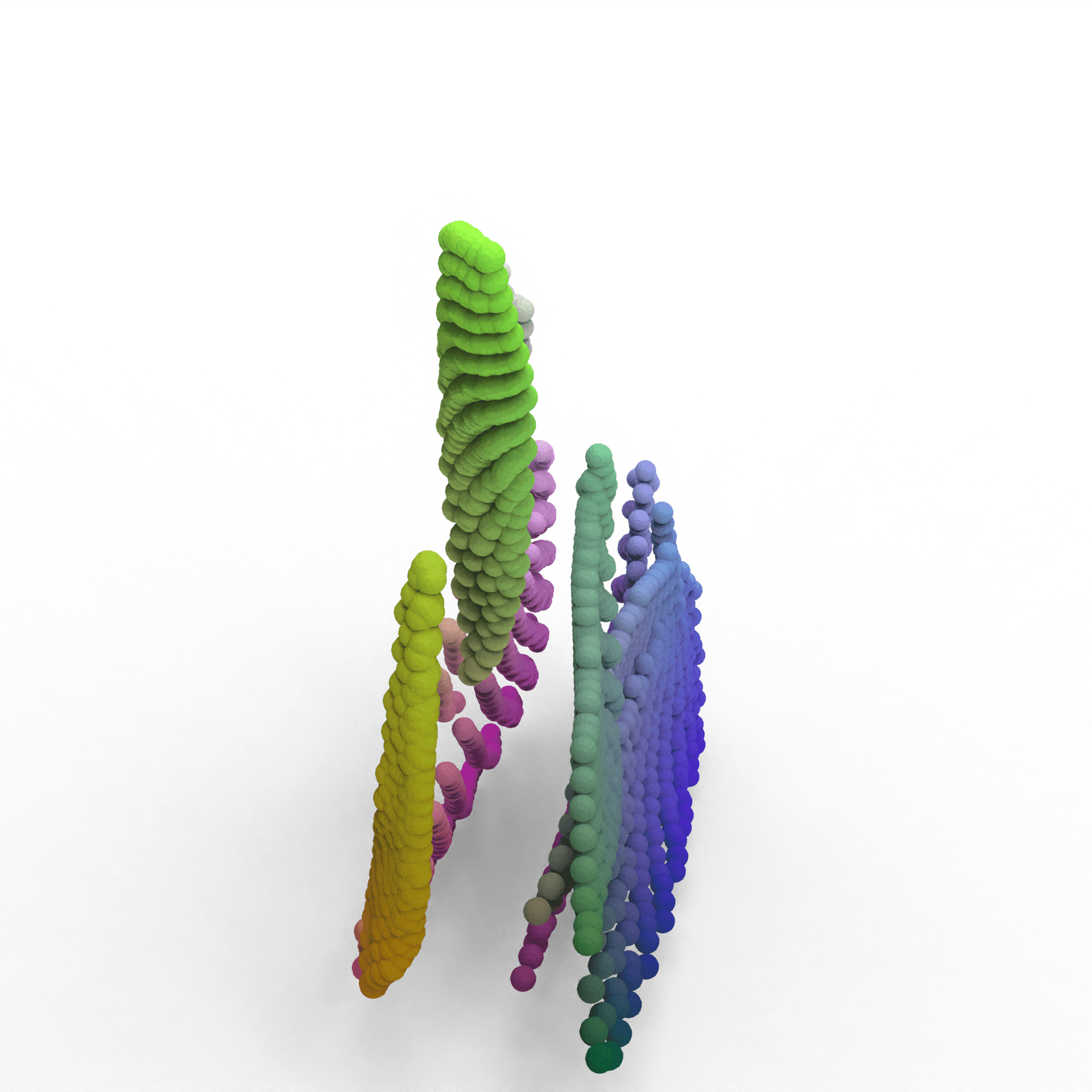}}
\label{failure:vessel} 
\subfloat[Human]{
\includegraphics[width=1.2in]{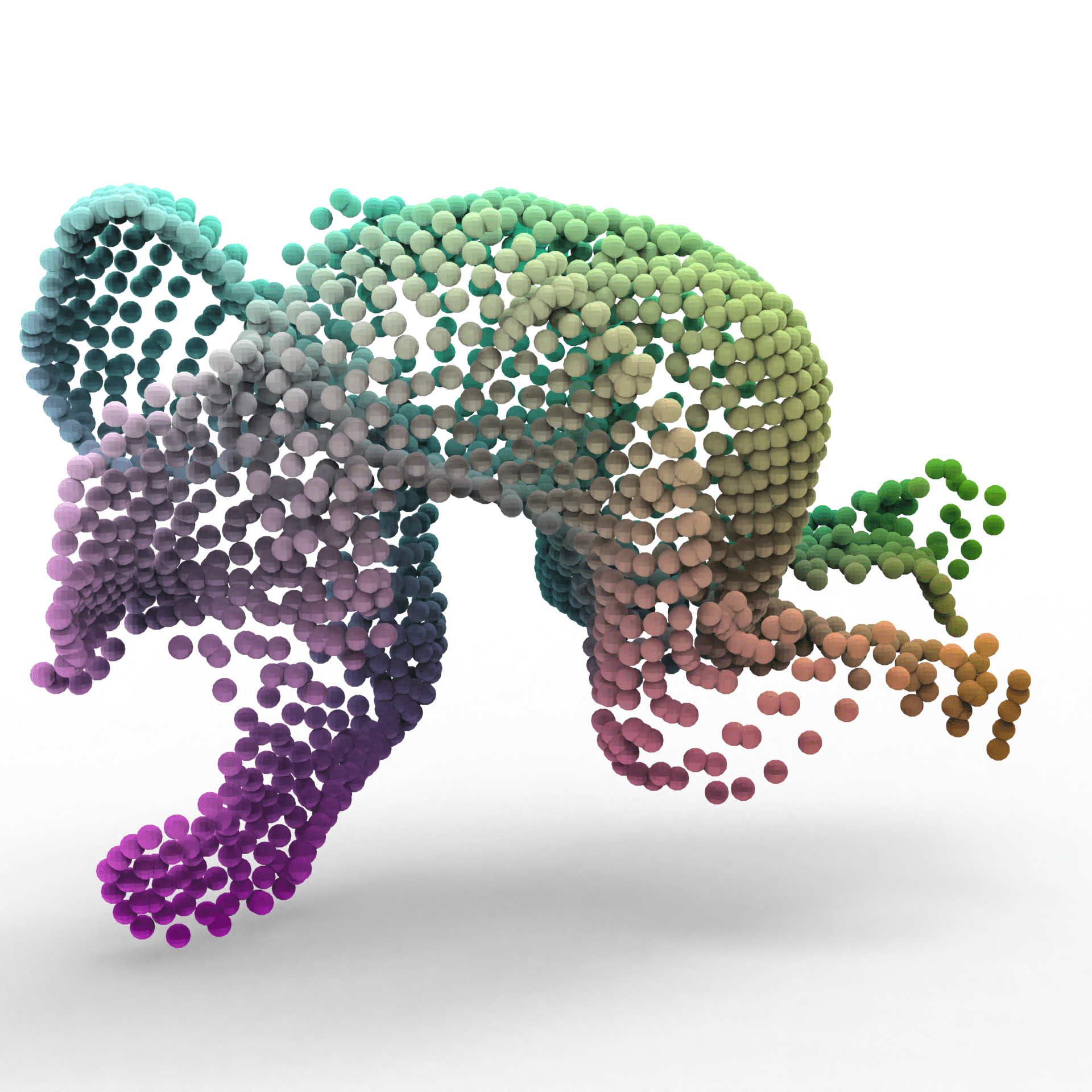}}
\label{failure:human} 
\caption{Failure and low-quality cases of (a) \textit{chair}, (b) \textit{airplane}, (c) \textit{lamp}, (d) \textit{vessel} and (e) \textit{human}.}
\label{failure}
\end{figure}

Although our WarpingGAN achieves better performance than state-of-the-art methods, failure and low-quality cases still occur, as GAN-based 3D point cloud generation is a pretty challenging problem and is difficult to train, especially equipped with the weak discriminator PointNet. Thus, WarpingGAN sporadically fails to generate shapes and cannot learn local details well. In Fig. \ref{failure}, we presented several failure and low-quality cases of our WarpingGAN, which may bring motivations to the subsequent studies along this stream. 

